\documentclass[twoside,11pt]{article}

\usepackage{blindtext}
\usepackage{jmlr2e}

\usepackage{algorithm}
\usepackage{tikz}
\usepackage{booktabs}
\usepackage{subfigure}
\usepackage{algorithmicx}
\usepackage[noend]{algpseudocode}
\usepackage{multirow}
\usepackage{makecell}
\usepackage{adjustbox}
\usepackage{amsmath}
\usepackage{bm}
\usepackage{tabularx}
\usepackage{enumitem,comment}

\usepackage{lastpage}
\ShortHeadings{SA-based Distortion Risk Measure Optimization}{Jiang, Heidergott, Hu, and Peng}
\firstpageno{1}

\begin{document}

\title{Stochastic Approximation Methods\\ for Distortion Risk Measure Optimization\vspace{1.5em}}

\author{\name Jinyang Jiang \email jinyang.jiang@stu.pku.edu.cn \\
       \addr Guanghua School of Management\\
       Peking University,
       Beijing 100871, China
       \AND
       \name Bernd Heidergott \email b.f.heidergott@vu.nl \\
       \addr School of Business and Economics\\
       Vrije Universiteit Amsterdam,
       Amsterdam 1081 HV, Netherlands
       \AND
       \name Jiaqiao Hu \email jqhu@ams.stonybrook.edu \\
       \addr Department of Applied Mathematics and Statistics\\
       State University of New York at Stony Brook,
       Stony Brook, NY 11794, USA
       \AND
       \name Yijie Peng \email pengyijie@pku.edu.cn \\
       \addr Guanghua School of Management \\
       School of Artificial Intelligence for Science\\
       Peking University,
       Beijing 100871, China\\
       Xiangjiang Laboratory, 
       Changsha 410200, China
       }

\editor{}

\maketitle

\begin{abstract}
Distortion Risk Measures (DRMs) capture risk preferences in decision-making and serve as general criteria for managing uncertainty. This paper proposes gradient descent algorithms for DRM optimization based on two dual representations: the Distortion-Measure (DM) form and Quantile-Function (QF) form. The DM-form employs a three-timescale algorithm to track quantiles, compute their gradients, and update decision variables, utilizing the Generalized Likelihood Ratio and kernel-based density estimation. The QF-form provides a simpler two-timescale approach that avoids the need for complex quantile gradient estimation. A hybrid form integrates both approaches, applying the DM-form for robust performance around distortion function jumps and the QF-form for efficiency in smooth regions. Proofs of strong convergence and convergence rates for the proposed algorithms are provided. In particular, the DM-form achieves an optimal rate of $O(k^{-4/7})$, while the QF-form attains a faster rate of $O(k^{-2/3})$. Numerical experiments confirm their effectiveness and demonstrate substantial improvements over baselines in robust portfolio selection tasks. The method’s scalability is further illustrated through integration into deep reinforcement learning. Specifically, a DRM-based Proximal Policy Optimization algorithm is developed and applied to multi-echelon dynamic inventory management, showcasing its practical applicability.
\end{abstract}


\section{Introduction}
\label{sec:INTRODUCTION}
Risk measures play a central role in decision-making under uncertainty across diverse domains. Among them, the Distortion Risk Measure (DRM) provides a universal framework for representing risk preferences, particularly when dealing with asymmetric loss distributions. By distorting the underlying probability distribution, DRM emphasizes the risks of extreme events and offers decision-makers more accurate and context-specific risk assessments, making it a versatile criterion for managing uncertainty.
Due to its flexibility and adaptability, DRM is of significant importance in areas such as machine learning, finance, insurance, and other domains where risk control and optimization are essential. Maximizing the DRM objective function not only enhances the accuracy of risk measurement but also optimizes decision-making processes in dynamically changing environments, offering decision-makers more reasonable risk control strategies.

\begin{figure}[!b]
\vspace{-0.5cm}
\centering
\hspace{2cm}\includegraphics[trim=0cm 0.5cm 0cm 0cm, width=0.85\linewidth]{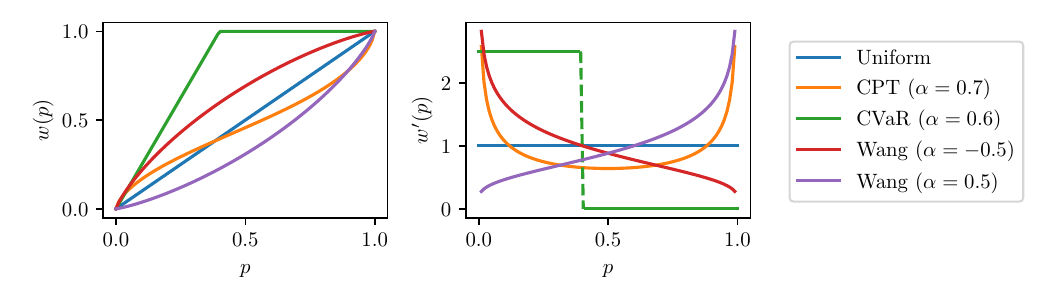}
\caption{Visualization of example distortion functions and their derivatives.}
\label{fig: distortion functions}
\end{figure}

As a weighted integral of the cumulative distribution function (CDF), DRM provides a unified formulation for a wide range of risk measures by distorting the probability with different distortion functions as shown in Figure \ref{fig: distortion functions}. 
The weight coefficients are associated with the derivatives of distortion functions. The derivative corresponding to Cumulative Prospect Theory \cite[CPT,][]{tversky1992advances} implies that two-sided tail behavior is given increased consideration, while other listed risk measures pay attention to one-sided tail.
With careful configuration, DRM has a better capacity to reflect human risk perception.

It is worth noting that DRMs are typically applied in constrained optimization. For example, in the field of investment or reinforcement learning (RL), one may wish to minimize the Conditional Value-at-Risk (CVaR) of a portfolio or decision policy 
subject to a minimum mean return, or conversely, maximize the mean return subject to a CVaR constraint. For ease of exposition, this paper focuses on the unconstrained setting where the DRM serves as the objective. The extension to the more realistic constrained case can be handled using penalty functions \citep{2times} or the Arrow–Hurwicz algorithm \citep{arrow1958studies}.

Due to our focus on efficiently optimizing larger problems, gradient-based methods have almost become the only viable solution.
However, unlike mean-based objectives, even basic risk measures may impose difficulties in estimating their gradients, which motivates researchers to develop various techniques, such as kernel estimation \citep{liu2009kernel,hong2009simulating}, infinite perturbation analysis \citep{hong2009estimating,jiang2015estimating}, and measure-valued differentiation \citep{heidergott2016measure}.
\cite{glynn2021computing} further utilize the Generalized Likelihood Ratio (GLR) method \citep{peng2020maximum} to derive DRM estimators.
Existing literature primarily focuses on applying estimation toolkits to risk measures directly, while neglecting the coordination between the optimization framework and gradient estimation, where significant potential exists for designing algorithms that achieve both correctness and efficiency.

We derive two dual representations of the DRM gradient.
When using the distortion function as the measure, the resulting numerical integration format takes the form of a weighted sum of quantile gradients. This approach, 
called {\em Distortion-Measure form} (DM-form), only requires the integrability of the distortion function, without assuming smoothness. However, accurately estimating the quantile gradient in this format is a challenging task, as the quantile gradient involves quantiles that change as the optimization progresses. On the other hand, when using the quantile function as the measure, the resulting expression is a weighted sum of distribution gradients, with weights determined by the derivative of the distortion function, which therefore must be smooth.
This representation is named as {\em Quantile-Function form} (QF-form) in the following.
We explore both choices and propose two corresponding multiscale algorithms for obtaining DRM gradients that can later be applied in optimization.
Both algorithms are based on the discretization of numerical integration. This offers the opportunity of combining the DM-form with the QF-form, and we will do so by introducing the {\em Hybrid form}, which will be the basis for establishing a gradient estimator achieving superior overall performance than the DM- and QF-form.

The paper makes the following main contributions:
\begin{itemize}
    \item We propose two stochastic approximation (SA) algorithms for DRM optimization, based on the DM- and QF-form representations of the DRM gradient.
    For both algorithms, we establish strong convergence and establish convergence rates.
    \item We introduce a new hybrid algorithm by integrating the two principal ones and later extend it to deep RL by developing a DRM-based policy optimization algorithm that demonstrates the scalability of our methodology.
    \item Numerical experiments are conducted to illustrate the performance of our algorithms against existing baselines on realistic business applications, including robust portfolio selection and complicated inventory management.
\end{itemize}

The rest of the paper is organized as follows:
Section \ref{section: Problem Formulation & Preliminaries} introduces the DRM optimization problem and discusses the underlying challenges.
In Section \ref{section: Three-timescale DRM Optimization}, we present QF-form leading to a three-timescale algorithm for DRM optimization, and in Section \ref{section: Two-timescale DRM Optimization}, we introduce the more efficient two-timescale optimization via the DM-form.
Finally, in Section~\ref{sec:hybrid}, the hybrid estimator is introduced.
Convergence analysis, including strong convergence and convergence rates, is provided in appendices.
Section~\ref{section: Numerical Results} presents simulation experiments that evaluate and illustrate the performance of the proposed algorithms.

\section{Problem Formulation \& Preliminaries}\label{section: Problem Formulation & Preliminaries}
Suppose that $\theta\in\Theta\subset\mathbb{R}^d$, for $ d \geq 1 $, is a decision variable on a compact and convex parameter space, affecting the distribution of $Y$.
We refer to $ Y$ as the {\em observable} and denote its CDF by $ F ( \cdot ;\theta )$ and its Probability Density Function (PDF) by $f ( \cdot ; \theta )$.
A distortion function $ w ( \cdot ) $ is a mapping $w: [0,1] \mapsto [ 0,1] $ with $w(0)=0$ and $w(1)=1$.
For given $ w (\cdot )$, the associated DRM of $ Y $ at $ \theta $ is defined as
\begin{align}
    \mathcal{J}(\theta) = \int_{0}^{+\infty}w(1-F(y;\theta))dy + \int_{-\infty}^{0} (w(1-F(y;\theta))-1) dy. \label{eq:drm_def_1}
\end{align}
Note that the convexity of $w(\cdot)$ models a risk-averse preference, and the concavity a risk-seeking preference.
Examples of DRMs are Value-at-Risk (VaR), 
where $w(z)=\mathbf{1}\{z>1-\alpha\}$ with $\mathbf{1}\{\cdot\}$ denoting the indicator function, and
CVaR, where $w(z)=\min\{\frac{z}{1-\alpha},1\}$, respectively. 
If $ w (\cdot)$ is concave, then the associated DRM is coherent, which is an essential property of a DRM for being able to reflect risk in a consistent way.

Any gradient search algorithm applied to solve a complex optimization problem containing a DRM requires an efficient method to estimate the gradient $ \nabla_\theta {\mathcal{J}} ( \theta ) $.
With proper integrability conditions for the limit interchange, e.g., $\int_{\mathbb{R}}\sup_{\theta\in\Theta}\Vert w'(1-F(y;\theta))\nabla_{\theta}F(y;\theta)\Vert dy<\infty$,
the direct gradient of the objective (\ref{eq:drm_def_1}) can be given by 
\begin{align}
    \nabla_{\theta}\mathcal{J}(\theta)= \int_{\mathbb{R}}-w'(1-F(y;\theta))\nabla_{\theta}F(y;\theta)dy. \label{eq:drm_grad_0}
\end{align}
It should be noted that $\nabla_{\theta}F(y;\theta) $ in (\ref{eq:drm_grad_0}) can be translated into an unbiased estimator by representing it via an appropriate measure-valued derivative, see \cite{FH}.
Unfortunately, evaluating (\ref{eq:drm_grad_0}) then requires carrying out two experiments in parallel, and it thus does not lend itself to be applied to streaming data.

If $w(\cdot)$ is non-decreasing and continuous from the left, then the equality (\ref{eq:drm_def_1}) can be transformed into a Lebesgue-Stieltjes integral over a bounded interval \citep{dhaene2012remarks}, that is,
\begin{align}
    \mathcal{J}(\theta) = \int_{0}^{1} F^{-1}(1-z;\theta)dw(z). \label{eq:drm_def_2}
\end{align}
The gradient of (\ref{eq:drm_def_2}) is given by the weighted integral of VaR gradients, also known as quantile gradients. If $F^{-1}(z;\cdot)$ is sufficiently smooth and the density $f(\cdot;\theta)$ is non-zero and continuous, then the gradient of (\ref{eq:drm_def_2}) can be obtained by applying the implicit function theorem \citep{glynn2021computing}. 
This gives the following DM-form of the DRM gradient,
\begin{align}
\text{DM-form:} \quad 
    \nabla_{\theta}\mathcal{J}(\theta) = -\int_{0}^{1} \nabla_{\theta}F^{-1}(z;\theta)d\tilde{w}(z)
    = -\int_{0}^{1} -\frac{\nabla_{\theta}F(q;\theta)}{f(q;\theta)}\bigg|_{q=F^{-1}(z;\theta)}d\tilde{w}(z), \label{eq:drm_grad}
\end{align}
where $\tilde{w}(z) = w(1-z)$.
Unfortunately, the DRM gradient in (\ref{eq:drm_grad}) cannot be calculated analytically and has to be estimated via simulation.
In doing so, one further faces the notoriously difficult problem of estimating the density $f(\cdot;\theta)$.

In the following, we introduce an alternative representation for the DRM gradient.
Provided that the distortion function is sufficiently smooth, we change the integration measure from $ d\tilde{w}(\cdot)$ to $dF^{-1}(\cdot;\theta) $, and obtain the 
QF-form of the DRM gradient as 
\begin{align}
\text{QF-form:} \quad 
    \nabla_{\theta}\mathcal{J}(\theta)
    = \int_{0}^{1} \nabla_{\theta} F(q;\theta)\big|_{q=F^{-1}(z;\theta)} \tilde{w}'(z) d F^{-1}(z;\theta). \qquad \qquad \qquad \qquad \ \ \    \label{eq:QF}
\end{align}
The QF-form of the DRM gradient requires approximating the quantile function $ F^{-1} (\cdot;\theta)$. 
Even more challenging is that, in an optimization context, $ \theta $ changes over the iterations, and the estimation of $ F^{-1}( \cdot ; \theta )$ must be dynamically updated at each step.

\section{Related Literature}

In machine learning, model performance is typically studied through its expected behavior, and its improvement relies on standard gradient-based descent methods. Consequently, efficient gradient computation and associated direct optimization emerge as two central challenges. More broadly, gradient estimation is a form of sensitivity analysis, which has been extensively studied since the development of sample-path methods such as infinitesimal perturbation analysis and the score function method. For performance criteria that can be expressed in expectation forms, related methodologies are well established in applied probability and operations research; see \cite{kroese2013handbook,FH}. However, DRMs do not fall into this category, for which both sensitivity analysis and optimization remain challenging. This paper advances the study of DRM optimization via multi-timescale SA methods and connects to the following two streams of the literature.

\textbf{Sensitivity Analysis of Risk Measures.}
Although the literature on risk measures and their sensitivity estimation is vast, as discussed in Section \ref{section: Problem Formulation & Preliminaries}, many of them can be unified through transformations of quantiles, which underscores the central role of quantiles in representing risk. Thus, we first review the quantile literature, which is most relevant to our study and has long been recognized as a challenge in the field.
An early reference of quantile estimation for stochastic models is \cite{heidelberger1984quantile},
where an importance sampling (IS) scheme is developed to transform a given quantile into a median.
The breakthrough papers \citep{hong2009simulating,hong2009estimating} provided the first results on quantile sensitivity analysis, establishing sufficient conditions for the asymptotic unbiasedness of the estimator. 
The main difficulty of this approach is its requirement for a.s. Lipschitz continuity of the underlying random variables, a condition often hard to verify \citep{jiang2015estimating}. To relax it, \cite{Fu:09} propose a conditioning method in a portfolio scenario. While handling non-Lipschitz mappings $Y = h(X(\theta))$, with a loss function $h$ of the $\theta$-dependent random variable $X(\theta)$, it requires detailed analysis at each discontinuity and the invertibility of $h$, which can be infeasible for complex models \citep{Hong:09OR}.
The quantile sensitivity was further advanced by \cite{liu2009kernel} through the introduction of kernel estimators. Subsequent work by \cite{LiuHong2} showed that kernel estimators not only exhibit negligible bias in practice but can also be employed to smooth discontinuities of $h$. For an alternative to deal with discontinuities for sample path derivatives, we refer to \cite{LT}.

Extensions of sensitivity analysis to CVaR can be found in \cite{hong2009simulating,2}. \cite{lei2018applications} study distribution sensitivity estimation for both finite-horizon and steady-state simulations. Regarding DRMs, \cite{gourieroux2006sensitivity} analyze sensitivity with respect to distortion function parameters and establish a central limit theorem. \cite{cao2017sensitivity} propose a DRM sensitivity estimator as an integral of the quantile-based estimators in \cite{hong2009estimating}. More recently, \cite{glynn2021computing} provided results based on both the vanilla form (\ref{eq:drm_grad_0}) and the DM-form (\ref{eq:drm_grad}), together with their asymptotic properties.
\cite{lei2022estimating} further derive confidence region results for DRMs and their sensitivity estimators.
For a broader overview of developments in this area, we refer to \cite{fu2015handbook}.

\textbf{Optimization of Risk Measures and Applications in RL.}
Studies reviewed in the first part have provided gradient or sensitivity estimators for static models. Through batching, that is, applying single-timescale SA based on multiple samples, these estimators can be directly applied to risk optimization, which is one of the ultimate purposes of this line of research. However, such estimators remain biased for any finite batch size, thereby hindering the efficiency and accuracy of optimization. A more refined integration of sensitivity estimation with the optimization recursion structure could enable asymptotically unbiased and more efficient procedures. \cite{hu2022stochastic} developed a three-timescale stochastic approximation method for quantile objectives in systems with closed-form structures. Extending this line of work, \cite{hu2025quantile} proposed finite-difference and stochastic perturbation-based three-timescale algorithms that are compatible with black-box optimization, albeit under stronger smoothness assumptions. Further generalizations to CVaR-based black-box optimization were provided in \cite{hu2025simulation}.

Risk measures are introduced into RL to produce safe and robust decision-making. Early work includes studies on the expected exponential utility \citep{borkar2001sensitivity} and CVaR \citep{petrik2012approximate,tamar2014policy} as objectives. \cite{prashanth2016cumulative} employ a stochastic perturbation-based approach for optimizing CPT. \cite{jiang2022quantile} develop a two-timescale iterative algorithm to optimize quantiles. More recently, risk-based objectives are investigated in multi-agent deep RL combined with game-theoretic formulations \citep{slumbers2023game}.
Risk measures are also employed as constraints, where sensitivity analysis still plays an important role. For instance, \cite{bertsekas1997nonlinear} applies a Lagrangian approach to address risk-constrained RL, and both quantile and CVaR are considered as risk constraints in \cite{borkar2014risk} and \cite{chow2017risk}. Moreover, several risk-oriented studies propose value-based algorithms that manage uncertainty by combining value distribution approximation with risk-sensitive greedy selection, thereby enhancing the robustness of decision-making \citep{bellemare2017distributional,dabney2018distributional,rowland2024analysis}.

Compared with the existing literature, in addition to the DM-form, this paper introduces a QF-form that complements the DM-form in various aspects, and integrates the two to form a family of flexible DRM sensitivity estimation approaches. While prior studies mainly focus on constructing estimators for DRM or its sensitivity, our work further contributes by carefully designing three- and two-timescale SA algorithms for solving optimization problems involving DRM. These algorithms allow complex risk measure optimization to perform effectively under flexible batch size selection, without requiring a trade-off between iteration efficiency and estimation unbiasedness, which will be further discussed in the next section. Moreover, this paper extends the application of DRM optimization to the setting of deep RL, representing another addition to the literature.

\section{The DM-Form: Three-Timescale DRM Optimization}\label{section: Three-timescale DRM Optimization}

In this section, we develop the first algorithm based on the DM-form  
and provide theoretical guarantees by establishing strong convergence and characterizing the convergence rate.

For the DM-form of the DRM gradient, we discretize the integration interval $[0,1]$ using a grid $0< z_0< z_1<\cdots<z_N <1$, with $ z_0\to 0 $, 
$ z_N\to1$, and $\sup_i\vert z_i-z_{i-1}\vert\to0$ as $N\to\infty$. A standard choice is the uniform partition $z_i=(i+1)/(N+2)$ for $i=0,\cdots,N$.
Then we take 
$$
\mathcal{J}_N ( \theta ) = -\sum_{i=1}^N F^{-1} ( \tilde{z}_i ;\theta) (\tilde{w}(z_i) - \tilde{w}(z_{i-1}) )
$$
as the approximate DRM.
The DM-form gradient for proxy $ \mathcal{J}_N ( \theta )$ is given by
\begin{align}
    {\nabla}_{\theta}\mathcal{J}(\theta) \approx {\nabla}_\theta \mathcal{J}_N(\theta) = - \sum_{i=1}^N d_i(\theta) (\tilde{w}(z_i) - \tilde{w}(z_{i-1}) ) , \label{eq:drm_grad_numerical}
\end{align}
for $ N$ sufficiently large,
where
\begin{equation}\label{eq:true}
d_i(\theta) = -{\nabla_{\theta}F(q;\theta)|_{q=F^{-1}(\tilde{z}_i;\theta)}}/{f(F^{-1}(\tilde{z}_i;\theta);\theta)} 
\end{equation}
is the true gradient of the $\tilde{z}_i$-quantile evaluated at $ \theta$; 
and $\tilde{z}_i $ is an arbitrary point in $[z_{i-1} , z_{i}]$.
To simplify the notation, we will henceforth set
$\tilde z_i = z_i$, for $ i=1, \cdots, N$. 
The numerical integral substitute enables us to focus on tracking a series of quantiles and their gradients.

Instead of using a large batch of samples for plug-in estimation of the quantile and density at each iteration of the optimization, we propose a method to estimate different components of the DRM gradient and update the parameters with flexible batch sizes simultaneously. Our approach involves three recursions, each running with its individual timescale.
The input for our algorithm is the simulation output $ Y_k$  of the observable following $\theta_k$.
The first two recursions trace the quantiles in (\ref{eq:true}) and their gradients, while the last updates the decision variable $\theta$. 

Given a prefixed decision variable $\theta$, the quantile estimates can be found using the following recursive procedure:
\begin{align}
    q_{k+1,i} = q_{k,i} + \gamma^{q}_k({z}_i-\mathbf{1}\{Y_{k}\leq q_{k,i}\}),\quad \text{for}\ i=1,\cdots,N, \label{eq:iter_q_single}
\end{align}
where $\gamma^{q}_k$ is the step-size, and $q_{k,i}$ is an estimate of $F^{-1}({z}_i;\theta)$ at the $k$-th iteration. 
Note that $F(y;\theta)=\mathbb{E}_\theta[\mathbf{1}\{Y\leq y\}]$, recursion (\ref{eq:iter_q_single}) is an SA that solves $F(q;\theta)={z}_i$ in the $L^2$-norm sense, where we make use of the fact that the density is nonnegative; see, e.g., \cite{Spa:2003, hu2022stochastic, jiang2023quantile}.

Since the quantile gradient $d_i(\theta)$ takes a complicated ratio form, it does not have an analytical formula for general estimation and faces the ratio bias issue when the numerator and denominator are estimated separately with finite samples. 
However, the ratio-form representation of $d_i(\theta)$ can be reformulated as the solution to a linear equation, thereby avoiding this bias issue \citep{hu2022stochastic}.
Alternatively, by leveraging the definition of $d_i(\theta)$, we can approximate it  by directly solving
$$
D^*(\theta) = \arg\min_{D\in\mathbb{R}^{d}} \big \Vert D+ \nabla_{\theta}F(q;\theta)|_{q=F^{-1}({z}_i;\theta)} /{f(F^{-1}({z}_i;\theta);\theta)}\big \Vert^2, 
$$
where
$
-\nabla_{\theta}F(q;\theta)|_{q=F^{-1}({z}_i;\theta)} -  f(F^{-1}({z}_i;\theta);\theta) D
$
can serve as a descent direction, which is justified by the non-negativity of the density function. The solution $D^*(\theta) = d_i(\theta) $ is globally unique.
Thus, we construct the following recursions to track the quantile gradients when $\theta$ is prefixed, i.e., for $i=1,\cdots,N$,
\begin{align}
D_{k+1,i} = D_{k,i}+\gamma^{D}_{k}(G_1(Y_{k};\theta,F^{-1}({z}_i;\theta)) - G_2(Y_{k};\theta,F^{-1}({z}_i;\theta))  D_{k,i}), \label{eq:iter_d_single}
\end{align}
where $\gamma^{D}_{k}$ is the step-size, $G_1(Y;\theta,q)$ and $G_2(Y;\theta,q)$ are estimators of $-\nabla_{\theta}F(q;\theta)$ and $f(q;\theta)$, respectively, and $D_{k,i}$ is an estimate of $d_i(\theta)$ at the $k$-th iteration.

In applications, the CDF~$F(\cdot;\theta)$ of the observable $ Y $ is typically not available in closed form; therefore, we have to resort to simulation.
Assume that 
\begin{align}
    Y(\theta) = \mathcal{L}(X(\theta);\theta), \label{eq:model}
\end{align}
where $X(\theta)\in\mathcal{X}\subset\mathbb{R}^{\hat d}$,
for $ \hat d \geq 1 $, is generated from some input distribution with PDF $f_X(\cdot;\theta)$, and let $\mathcal{L}(\cdot;\theta)$ represent the input evaluation procedure. 
Using the GLR method \citep{peng2020maximum}, 
instances for $G_1(Y;\theta,q)$ and $G_2(Y;\theta,q)$ in (\ref{eq:iter_d_single}) are available; see Appendix~\ref{sec:yije} for further details.
Although these are unbiased estimators, they require full knowledge of $\mathcal{L}(\cdot;\theta)$, which is often too restrictive in many artificial intelligence applications, such as deep RL. In such settings, the system dynamics are typically unknown or treated as a black box.
Therefore, in the remaining part, we focus on the case where
$\mathcal{L}(\cdot;\theta)$ does not depend on $ \theta $, i.e., $\nabla_{\theta}\mathcal{L}(x;\theta)\equiv 0$. Under this assumption, the GLR estimator (\ref{eq:glr_1}) then simplifies to 
\begin{align}
    G_1(x;\theta,y) = -\mathbf{1}\{\mathcal{L}(x;\theta)\leq y\} \nabla_{\theta}\ln f_X(x;\theta). \label{eq:lr_111}
\end{align}
The minimum knowledge required for (\ref{eq:lr_111}) is the logarithmic density gradient $\nabla_{\theta}\ln f_X(\cdot;\theta)$, called the score function, which is well studied in the literature on stochastic optimization.
For an early reference, we refer to
\cite{reiman1986sensitivity}
and to \cite{kroese2013handbook,FH} for an overview of the current state of the literature.
We also substitute $G_2(y;\theta,q)$ with the kernel-based method \citep{wand1994kernel} for better flexibility, i.e., 
$G_3(y;\theta,q,h) = \frac{1}{h}\mathcal{K}(\frac{y-q}{h})$,
where the additional parameter $h$ is the bandwidth, $\mathcal{K}(\cdot)$ denotes a kernel function that satisfies $\int_{\mathbb{R}} \mathcal{K}(z)dz=1$, $\int_{\mathbb{R}} z\mathcal{K}(z)dz=0$ and $\int_{\mathbb{R}} z^2\mathcal{K}(z)dz<\infty$, for example, the density function of a standard normal distribution $\mathcal{K}(z) = \frac{1}{\sqrt{2\pi}} \exp{\{-\frac{z^2}{2}\}}$. To control the estimation variance, it is also assumed that the kernel function is square integrable, i.e., $\int_{\mathbb{R}}\mathcal{K}^2(z)dz<\infty$, which is a common setting in kernel-based methods. Although $G_3(Y;\theta,y,h)$ is not an unbiased estimator of $f(y;\theta)$, the bias can be controlled by varying $h$. 
In the remainder of this paper, we work under this assumption and write the evaluation function as $\mathcal{L}(\cdot)$ for simplicity. However, all results can be easily extended to the general case (\ref{eq:model}) when the GLR estimators are available.

In the last recursion, we substitute $d_i(\theta_k)$ with the estimated quantile gradients instead to compute the numerical integral (\ref{eq:drm_grad_numerical}) and perform the gradient ascent method.
This leads to our proposed three-timescale DRM optimization algorithm as follows:
\begin{align}
    D_{k+1,i} &= D_{k,i}+\gamma^{D}_{k}(G_1(X_{k};\theta_k, q_{k,i}) - G_3(Y_{k};\theta_k, q_{k,i}, h_k)  D_{k,i}),\ &\text{for}\ i=1,\cdots,N, \label{eq:iter_d}\\
    q_{k+1,i} &= q_{k,i} + \gamma^{q}_k({z}_i-\mathbf{1}\{Y_{k}\leq q_{k,i}\}),\ &\text{for}\ i=1,\cdots,N, \label{eq:iter_q}\\
    \theta_{k+1} &= \varphi_{\Theta}\big(\theta_k + \gamma^{\theta}_{k} \sum_{i=1}^N -D_{k,i}(\tilde{w}(z_i) - \tilde{w}(z_{i-1}) )\big), \label{eq:iter_theta}
\end{align}
where $\gamma^{\theta}_k$ is the size of the gradient search step, $\varphi_{\Theta}(\cdot)$ represents a projection operation that brings an iterate $\theta_{k+1}$ back to the parameter space $\Theta$ whenever it becomes infeasible, and $h_k$ is a decreasing bandwidth of kernel smoothing. Iterative updates can be evaluated through the observed realization of $ Y_k $, which enables us to use past observations in optimization.

\begin{remark}\label{re:proj1}
We have $ \theta_{k+1} = \theta_k $ in (\ref{eq:iter_theta}) if either the update 
$
\gamma^{\theta}_{k} \sum_{i=1}^N -D_{k,i}(\tilde{w}(z_i) - \tilde{w}(z_{i-1}) )
$
is zero (i.e., $ \theta_k $ is a stationary point), or $ \theta_k $ plus the update
is projected back to $ \theta_k$.
When $ \Theta $ has smooth boundaries, the latter implies that the above update is perpendicular to the boundary of $ \Theta$, and in the general case that the above update vector is in the convex cone to $ \Theta $ at $ \theta_k $.
For details we refer to 
\cite{kushner2003stochastic,FH}.
\end{remark}

Next, we analyze the convergence properties of our algorithm (\ref{eq:iter_d})-(\ref{eq:iter_theta}).
For ease of reference, we now introduce the key assumptions for analysis, where we denote the Euclidean norm by $\Vert \cdot \Vert$. 

\begin{assumption} \label{aspt:diff_quantile}
$\nabla_\theta F(q;\theta)$ and $f(q;\theta)$ are continuous in $q\in\mathbb{R}$, and $\theta\in\Theta$; 
$\nabla_\theta F^{-1} (z;\theta)$ and $F^{-1} (z;\theta)$ are continuous in $z\in[0,1]$, and $\theta\in\Theta$.
\end{assumption}

\begin{assumption} \label{aspt:step_size}
The step-size sequences $\{\gamma^D_k\}$, $\{\gamma^{q}_k\}$, $\{\gamma^{\theta}_k\}$, and bandwidth sequence $\{h_k\}$ satisfy

\noindent(a) $\gamma^D_k>0$, $\sum_{k=0}^{\infty}\gamma^D_k=\infty$, $\sum_{k=0}^{\infty}(\gamma^D_k)^2<\infty$; 
(b) $\gamma^{q}_k>0$, $\sum_{k=0}^{\infty}\gamma^{q}_k=\infty$, $\sum_{k=0}^{\infty}(\gamma^{q}_k)^2<\infty$;\\
\noindent(c) $\gamma^{\theta}_k>0$, $\sum_{k=0}^{\infty}\gamma^{\theta}_k=\infty$, $\sum_{k=0}^{\infty}(\gamma^{\theta}_k)^2<\infty$;\hspace{0.8em}
(d) $\gamma^{\theta}_k=o(\gamma^{q}_k)$, $\gamma^{q}_k=o(\gamma^D_k)$;\\
\noindent(e) $h_k>0$, $h_k=o(1)$, $\sum_{k=0}^{\infty} (\gamma^{D}_k)^2 h_k^{-1}< \infty$.
\end{assumption}

\begin{assumption} \label{aspt:dist_grad}
The log-density gradient of simulation inputs w.r.t. $\theta$ is bounded, i.e., there exists a constant $C_{g}>0$ such that  $\sup_{X\in\mathcal{X}, \theta\in\Theta}\Vert \nabla_{\theta}\ln f_X(X;\theta) \Vert< C_{g}.$
\end{assumption}

Let $Q_i$ represent the set that contains the intervals between $q_{k,i}$ and $F^{-1}({z}_i;\theta_k)$ for $ k \in \mathbb{N}$. 

\begin{assumption} \label{aspt:density}
For each $ N \geq 1$, it holds for $\{q_{k,i}\}_{k \in \mathbb{N}} $, $ i=1 , \ldots , N$, given in (\ref{eq:iter_q}) and 
for $ \{ \theta_k \}_{k \in \mathbb{N}} $ in (\ref{eq:iter_theta})
that there exist constants $\varepsilon_f^N \in(0,1)$ and $C_f>0$ such that w.p.1, 

\noindent(a) $\inf_{(k,q)\in \mathbb{N}\times\bigcup_{i} Q_i}f(q;\theta_k)>\varepsilon_f^N$,  $\sup_{(k,q)\in\mathbb{N}\times\mathbb{R}} f(q;\theta_k)<C_f $; \\
\noindent(b) $\sup_{(k,q)\in\mathbb{N}\times\mathbb{R}}{|f''(q;\theta_k)|}<\infty$.
\end{assumption}

Assumption \ref{aspt:diff_quantile} is a common setting in continuous optimization and implies the smoothness of the DRM objectives. Assumption \ref{aspt:step_size} is standard in multi-timescale SA analysis and kernel smoothing. Assumption \ref{aspt:dist_grad} is made naturally to avoid computational overflow. Finally, Assumption \ref{aspt:density} includes the necessary conditions to control the bias and variance in the kernel density estimation and to ensure that the quantile gradients are well defined.

In the subsequent theorem, we will show that the recursions (\ref{eq:iter_d})-(\ref{eq:iter_theta}) 
track three coupled ordinary differential equations (ODEs):
\begin{equation}
\begin{aligned}
    \dot{D}_i(t) &= -\nabla_\theta F(q_i(t);\theta(t)) -f(q_i(t);\theta(t)) D_i(t), &\text{for}\ i=1,\cdots,N, \\ 
    \dot{q}_i(t) &= {z}_i - F(q_i(t);\theta(t)), &\text{for}\ i=1,\cdots,N,\\
    \dot{\theta}(t) &= -\sum_{i=1}^N D_i(t) (\tilde{w}(z_i) - \tilde{w}(z_{i-1})) + p(t),  \label{eq:ode_couple}
\end{aligned}
\end{equation}
where $\dot{\theta}(t) $ is a projected ODE, with the projection term $p(t)$ defined to ensure that the trajectory remains within the feasible region.
Due to the relationships of three step-sizes in Assumption \ref{aspt:step_size}, the latter ODE(s) can be viewed as static for analyzing the dynamics of the former one(s). Conversely, from the perspective of the slower ODE(s), the faster variables are always close to their stationary points.
Therefore, substituting the quasi-stationary solutions of the faster dynamics into the slower ones, we expect the sequence $\{\theta_k\}$ generated by the recursions (\ref{eq:iter_d})-(\ref{eq:iter_theta}) to follow the ODE 
\begin{align}
    \dot{\theta}(t) = \nabla_\theta\mathcal{J}_N(\theta(t)) + p(t), \label{eq:ode_final}
\end{align}
which corresponds to the surrogate objective $\mathcal{J}_N(\theta)$.

\begin{remark}
The ODE (\ref{eq:ode_couple}) 
is called a projected ODE.
It can be shown that the trajectories of the ODE are continuous even though the vector field in (\ref{eq:ode_couple}) is not; for details, we refer to \cite{ProjODE1,ProjODE2,FH}.
The projected ODE reaches an equilibrium point $ \theta^\ast $ if
$\dot{\theta}(t) = 0 $ for $ \theta ( t ) = \theta^\ast $ and $ \lim_{t\to\infty} \theta ( t ) = \theta^\ast.$
Extending on (\ref{eq:drm_grad_numerical}), we can now, 
like for the discrete approximation (see Remark~\ref{re:proj1}), argue that $\dot{\theta}(t) = 0$
together with $ p ( t ) =0 $ implies that $\theta^\ast $ is a stationary point
of $ \mathcal{J} ( \theta) $.
\end{remark}

Now the following convergence results can be established, for proofs we refer to Appendix~\ref{sec:22}.
First, we establish strong convergence of the recursions (\ref{eq:iter_d})-(\ref{eq:iter_theta}) to the corresponding ODEs in (\ref{eq:ode_couple}), and further to the solution of the surrogate problem.

\begin{theorem}\label{th:main_convergence}
If Assumptions \ref{aspt:diff_quantile}-\ref{aspt:density} hold, then the sequence $\{\theta_k\}_{k\in\mathbb{N}}$ generated by recursions (\ref{eq:iter_d})-(\ref{eq:iter_theta}) converge a.s. to some limit set of the ODE (\ref{eq:ode_final}).
More formally, it holds w.p.1 that 
$
\lim_{t \rightarrow \infty }  {\theta}(t) 
=\lim_{k \rightarrow \infty } \theta_k = \theta^\ast_N ,
$
where $ \theta^\ast_N  $ is a stationary point of the ODE (\ref{eq:ode_final}).

Moreover, if $ \theta^\ast_N$ is an inner point of $ \Theta$, then 
$ \theta^\ast_N$ is a stationary point of $\mathcal{J}_N (\theta) $
(i.e., $ \nabla \mathcal{J}_N (\theta^\ast_N) = 0  )$, if, otherwise, $ \theta^\ast_N $ lies on the boundary $ \Theta$, then $ \nabla \mathcal{J}_N (\theta^\ast_N) $ is in the convex cone to $ \Theta$ at $ \theta^\ast_N$.
\end{theorem}

In addition to the convergence of $\{\theta_k\}$, the first two recursions (\ref{eq:iter_d}) and (\ref{eq:iter_q}) also respectively converge a.s. to the true quantile and the quantile gradient evaluated at the limiting value $\theta^\ast_N$ as expected. We refer to the proofs of Theorem \ref{th:main_convergence} in Appendix \ref{sec:Proof of Strong Convergence} for details.
It is worth noting that Assumption~\ref{aspt:density} applies for given but arbitrary $ N$. 
When we want to study the limit with respect to $ N$, a stronger version of Assumption~\ref{aspt:density} is needed; see Remark~\ref{re:a5}.

Then, to characterize the finite-time performance of the proposed three-timescale algorithm, we consider specific step-sizes of the forms
\begin{equation}\label{eq:ratesa}
\gamma^{D}_k=ak^{-\alpha}, \: \gamma^{q}_k=bk^{-\beta}, \: \gamma^{\theta}_k=rk^{-\gamma}, 
\text{ and } \: h_k=h k^{-\eta},
\end{equation}
for $k\in\mathbb{N}$, where $a,b,r,h>0$,
$\frac{1}{2}<\alpha<\beta<\gamma<1$, and $\eta<2\alpha-1$.
We now establish our result on the convergence rate of the quantile tracker.

\begin{theorem}\label{th:mse_q}
If Assumptions \ref{aspt:diff_quantile}-\ref{aspt:density} hold for given $N$, 
then for all $i=1,\cdots,N$, the sequences $\{q_{k,i}\}_{k \in \mathbb{N}}$ generated by recursion (\ref{eq:iter_q}) satisfy 
    \begin{align*}
        \mathbb{E}\big[\vert q_{k,i}-F^{-1}({z}_i;\theta_k)\vert^2\big] = O \big (\gamma^{q}_k + (\gamma^{\theta}_k)^2 (\gamma^{q}_k)^{-2} \big)
         = O \big  ( k^{- \min \{\beta , 2 \gamma - 2\beta \} } \big).
    \end{align*}
\end{theorem}

Next, we analyze the gradient approximation under the following two additional assumptions.

\begin{assumption} \label{aspt:lip} For all $\theta\in \Theta$,
(a) $\nabla_{\theta}F(\cdot;\theta)$ is Lipschitz continuous, i.e., there exists a constant $L_d>0$ such that $\Vert \nabla_{\theta}F(q;\theta) - \nabla_{\theta}F(q';\theta)\Vert \leq L_d\vert q-q'\vert $;
(b) $f(\cdot;\theta)$ is Lipschitz continuous, i.e., there exists a constant $L_f>0$ such that $\vert f(q;\theta) - f(q';\theta)\vert \leq L_f\vert q-q'\vert$.
\end{assumption} 
\begin{assumption} \label{aspt:hessian}
Let $H(\theta)=\nabla_{\theta}^2 \mathcal{J}_N(\theta)$ and $\lambda(\theta)$ be the largest real part of its eigenvalues. There exists $C_{\lambda}^N>0$, such that $\lambda(\theta)<-C_{\lambda}^N$ for all $\theta\in\Theta$.
\end{assumption}

The following theorem characterizes the error bound for the quantile gradient estimator.

\begin{theorem}\label{th:mse_d}
If Assumptions \ref{aspt:diff_quantile}-\ref{aspt:hessian} hold for given $N$, 
 then for all $i=1,\cdots,N$, the sequences $\{D_{k,i}\}_{ k \in \mathbb{N}}$ generated by recursion (\ref{eq:iter_d}) satisfy 
    \begin{align*}
        \mathbb{E}[\Vert D_{k,i} - \nabla_\theta F^{-1}({z}_i;\theta_k) \Vert^2] = O\big((\gamma^{\theta}_k)^2 (\gamma^{q}_k)^{-2} +h_k^4 + \gamma^{D}_k h_k^{-1} \big)  = O \big  ( k^{- \min \{2 \gamma - 2\beta, 4\eta,\alpha-\eta \} } \big) .
    \end{align*}
\end{theorem}

Finally, we present the result on the mean squared error (MSE) convergence rate for $\theta_k$.

\begin{theorem}\label{th:mse_theta}
Let Assumptions \ref{aspt:diff_quantile}-\ref{aspt:hessian} hold for given $N$. 
If the algorithm in (\ref{eq:iter_theta}) converges to a unique $\theta^\ast_N$ that lies in the interior of $\Theta$, i.e., ${\nabla}_{\theta} \mathcal{J}_N(\theta^\ast_N) = 0$, then the sequence $\{\theta_{k}\}_{ k \in \mathbb{N}}$ generated by recursion (\ref{eq:iter_theta}) satisfy
\begin{align*}
        \mathbb{E}[\Vert \theta_k -\theta^\ast_N\Vert^2 ] =O\big((\gamma^{\theta}_k)^2 (\gamma^{q}_k)^{-2} +h_k^4 + \gamma^{D}_k h_k^{-1} \big)= O \big  ( k^{- \min \{2 \gamma - 2\beta, 4\eta,\alpha-\eta \} } \big) .
    \end{align*}
\end{theorem}

Theorem \ref{th:mse_theta} also implies the algorithm achieves an optimal sublinear convergence rate of $O(k^{-\frac{4}{7}})$ with the previously specified hyperparameter constraints, while not requiring batches of data for each iteration.
The above results are derived for a given $N$, it is important to note that the rates obtained are independent of $N$. As we show below, letting $N$ tend to infinity, we can establish convergence to the solution of the original problem.

\begin{remark}\label{re:a5}
{ Suppose that we replace Assumption~\ref{aspt:density} 
by the following uniform version, namely Assumption~\ref{aspt:density}$^{\ast}$:
For all $ \theta \in \Theta$, the support of $ f ( \cdot , \theta ) $ is a connected set contained in some compact set $ {\cal I}$.
Moreover, constants $\varepsilon_f \in(0,1)$ and $C_f > 0$ exist such that w.p.1,\\
\noindent(a) $\inf_{ ( q , \theta ) \in {\cal I } \times \Theta  } f(q;\theta )>\varepsilon_f $, and $\sup_{(q , \theta ) \in {\cal  I} \times \Theta} f(q;\theta)<C_f $; 
(b) $\sup_{(q,\theta)\in {\cal I } \times \Theta}{|f''(q;\theta)|}<\infty$.
\noindent Then we are able to establish convergence to a stationary point $\theta^*$ of the original objective $\mathcal{J}(\theta)$. Furthermore, if Assumption~\ref{aspt:hessian} is strengthened to Assumption~\ref{aspt:hessian}$^*$, where the constant $C_\lambda$ does not depend on $N$, we can also establish the MSE convergence rate of the sequence $\{\theta_k\}_{k\in\mathbb{N}}$ to the solution of the original problem as $N$ grows. 
}
\end{remark}

The extensions are as shown in the following two theorems. We specify the grid using a uniform partition, i.e., $z_i=(i+1)/(N+2)$, for $i=0,\cdots,N$.

\begin{theorem}\label{th:totalerror}
Under the conditions of Theorem~\ref{th:main_convergence}, with Assumption~\ref{aspt:density} replaced by \ref{aspt:density}$^\ast$, and provided $w(\cdot)$ is continuous at the boundary points $0$ and $1$, we have $$ \lim_{N\to\infty} \sup_{\theta\in\Theta} \Vert \nabla_\theta \mathcal{J}_N(\theta) - \nabla_\theta \mathcal{J}(\theta)\Vert = 0. $$ Furthermore, suppose $\theta^\ast\in\Theta^\circ$ is the unique stationary point (interior solution) of $\mathcal J(\theta)$. Then, it holds that
$\lim_{N\to\infty}\lim_{k\to\infty} \theta_k = \lim_{N\to\infty} \theta^\ast_N = \theta^\ast$ a.s.
\end{theorem}
\begin{theorem}\label{th:totalerrorrate}
Under the conditions of Theorem~\ref{th:mse_theta}, with Assumptions~\ref{aspt:density} and~\ref{aspt:hessian} replaced by \ref{aspt:density}$^\ast$ and \ref{aspt:hessian}$^\ast$, respectively, suppose that $\nabla_\theta F^{-1}(z;\theta)$ is uniformly Lipschitz in $z$, $w(\cdot)$ is Lipschitz, and $\mathcal{J}(\cdot)$ is locally strongly concave at the unique stationary point (interior solution) $\theta^\ast$. Then, the sequence $\{\theta_{k}\}_{ k \in \mathbb{N}}$ generated by recursion (\ref{eq:iter_theta}) satisfy
\begin{align*}
    \mathbb{E}[\Vert \theta_k -\theta^\ast\Vert^2 ] = \underbrace{O\big((\gamma^{\theta}_k)^2 (\gamma^{q}_k)^{-2} +h_k^4 + \gamma^{D}_k h_k^{-1} \big)}_{\mathrm{SA\ error}} + \underbrace{O(N^{-2}).}_{\mathrm{Integration\ error}}
\end{align*}
\end{theorem}

While Theorem~\ref{th:totalerror} is stated under uniqueness and interiority assumptions, analogous results for more general solution types can be obtained similarly; we omit details for brevity. And we highlight that, Theorem~\ref{th:totalerrorrate} reveals an interesting property: in our DRM optimization algorithm, the SA error and numerical integration error are fully decoupled, allowing for independent tuning of hyperparameters to efficiently control both errors. This also indicates that when choosing an integration discretization scheme with higher numerical precision, the convergence rate of our integration error can achieve a higher order. 
However, in single-timescale SA with plug-in estimation, the batch size per update determines the number of grid points used for DRM approximation, making it impossible to run without batching as in our approach.

\section{The QF-Form: Two-Timescale DRM Optimization}\label{section: Two-timescale DRM Optimization}

In this section, we utilize the reformulation of the DRM gradient and simplify the three-timescale algorithm by avoiding estimating the quantile gradient.
We propose a two-timescale variant with some additional constraints. The strong convergence of the algorithm and
rate of convergence results are derived to verify the algorithm.

In Section \ref{section: Three-timescale DRM Optimization}, we take advantage of the integral Lebesgue-Stieltjes form of the DRM gradient and take the distortion function as a measure. The corresponding numerical integration avoids using the derivative of the distortion function but pays an expensive cost in estimating quantile gradients instead.
Recursion (\ref{eq:iter_d}) occupies $N$ times the storage space of the parameter size and requires complex techniques to address the challenges of ratio bias and density estimation. We further observe that the estimated quantile function inherently encodes the density information, which motivates a more efficient approach.

Therefore, instead of using the original Lebesgue-Stieltjes integral in (\ref{eq:drm_grad}), we adopt the modified form in (\ref{eq:QF}). Accordingly, the approximated gradient can be expressed as
\begin{align}
    \nabla_{\theta}\mathcal{J}(\theta)
    \approx 
    \nabla_{\theta}\mathcal{J}_N^q (\theta) := 
    \sum_{i=1}^N \nabla_{\theta} F(q;\theta) \big |_{q=F^{-1}({z}_i;\theta)} \tilde{w}'({z}_i) (F^{-1}(z_i;\theta) - F^{-1}(z_{i-1};\theta)), \label{eq:drm_grad_numerical2}
\end{align}
where the density is excluded, and the distribution gradient can be estimated using the score function method in the same manner as in Section \ref{section: Three-timescale DRM Optimization}.
The only component that remains to be estimated is the quantile function.
Similarly to the recursion (\ref{eq:iter_q}), we track the quantile values at the points of the grid. By plugging in the estimated quantile values to the equation (\ref{eq:drm_grad_numerical}) and performing the gradient ascent, we obtain the two-timescale DRM optimization algorithm below:
\begin{align}
    q_{k+1,i} &= q_{k,i} + \gamma^{q}_k(z_i-\mathbf{1}\{Y_{k}\leq q_{k,i}\}),\ &\text{for}\ i=0,\cdots,N,\label{eq:iter2_q}\\
    \theta_{k+1} &= \varphi_{\Theta}\big(\theta_k + \gamma^{\theta}_{k} \sum_{i=1}^N G_1(X_k,\theta_k,{q}_{k,i})\tilde{w}'({z}_i)(q_{k,i}-q_{k,i-1} )\big),\label{eq:iter2_theta}
\end{align}
where the first two recursions are in the same timescale $\gamma^{q}_k$, while the last one uses $\gamma^{\theta}_{k}$ as the gradient search step-size. 
Since the algorithm avoids estimating the density function, it enables 
more stable updates and is much easier to tune. As shown later, it also achieves a faster convergence rate.

Similar to Section \ref{section: Three-timescale DRM Optimization}, the recursions (\ref{eq:iter2_q}) 
and (\ref{eq:iter2_theta})
are expected to track the coupled ODEs:
\begin{equation}
\begin{aligned}
    \dot{q}_i(t) &= z_i - F(q_i(t);\theta(t)), &\text{for}\ i=0,\cdots,N, \\
    \dot{\theta}(t) &= \sum_{i=1}^N \nabla_{\theta}F(q_i(t);\theta(t)) \tilde{w}'({z}_i) (q_i(t)-q_{i-1}(t)) + p(t),  \label{eq:ode2_couple}
\end{aligned}
\end{equation}
where $p(t)$, as before, denotes the correction force. 
Under Assumption~\ref{aspt:step_size}(d), the last ODE evolves on a slower timescale and can be treated as quasi-static when analyzing the dynamics of the quantile recursion. In particular, recursion~(\ref{eq:iter2_theta}) can be eventually viewed as tracking the ODE:
\begin{align}
    \dot{\theta}(t) = \nabla_{\theta}\mathcal{J}_N^q (\theta(t)) + p(t), \label{eq:ode2_theta}
   \end{align}
whereas other variables are always close to converging, and we can establish the strong convergence of recursions (\ref{eq:iter2_q}) and (\ref{eq:iter2_theta}) to some limit set of ODE (\ref{eq:ode2_theta}).
Analogous to Theorem \ref{th:main_convergence}, we first have the following main convergence result for our two-timescale algorithm.

\begin{theorem}\label{th:main_convergence2}
If Assumptions \ref{aspt:diff_quantile}, \ref{aspt:step_size}(b)-(d),  \ref{aspt:dist_grad}, and \ref{aspt:density}(a) hold, and $w(z)\in C^1[0,1]$, then the sequence $\{\theta_k\}_{k\in\mathbb{N}}$ generated by recursions (\ref{eq:iter2_q}) and (\ref{eq:iter2_theta}) converge a.s. to some limit set of ODE (\ref{eq:ode2_theta}).\end{theorem}

To characterize the finite-time performance of our algorithm in terms of the MSE of $\{\theta_k\}_{k\in\mathbb{N}}$, we adopt the step-size schedule in~(\ref{eq:ratesa}),
while dropping the constraints on $ \gamma_k^D $ and $ h_k$.
In addition, we introduce the following technical assumption, which parallels Assumption~\ref{aspt:hessian}:
\begin{assumption} \label{aspt:jocobi}
Let $\tilde{H}(\theta)=\nabla_{\theta}^{\top} {\nabla}_{\theta} \mathcal{J}^q_N(\theta)$ and $\mathrm{Re}(\tilde{\lambda}(\theta))$ be the largest real part of its eigenvalues. There exists $\tilde{C}_{\lambda}^N>0$, such that $\mathrm{Re}(\tilde{\lambda}(\theta))<-\tilde{C}_{\lambda}^N$ for all $\theta\in\Theta$.
\end{assumption} 

Next, we establish the convergence rates of the quantile trackers and $\theta_k$ for given $N$.

\begin{theorem}\label{th:mse2_q}
    If Assumptions \ref{aspt:diff_quantile}, \ref{aspt:step_size}(b)-(d), \ref{aspt:dist_grad}, and \ref{aspt:density}(a) hold for given $N$, and $w(z)\in C^1[0,1]$, then, for all $ i=1 , \cdots, N$, the sequences $\{q_{k,i}\}_{k\in\mathbb{N}}$ 
    generated by recursion (\ref{eq:iter2_q}) 
       satisfy 
    \begin{align*}
        \mathbb{E}[\vert q_{k,i}-F^{-1}(z_i;\theta_k)\vert^2] = O(\gamma^{q}_k +(\gamma^{\theta}_k)^2(\gamma^{q}_k)^{-2})=O(k^{-\min\{\beta,2\gamma-2\beta\}}).
         \end{align*}
\end{theorem}

\begin{theorem}\label{th:mse2_theta}
    Let Assumptions \ref{aspt:diff_quantile}, \ref{aspt:step_size}(b)-(d), \ref{aspt:dist_grad}, \ref{aspt:density}(a), \ref{aspt:lip}, and 
    \ref{aspt:jocobi} hold; and assume that $w(z)\in C^{1,1}[0,1]$.
    If the algorithm in (\ref{eq:iter2_theta}) converges to a unique $\theta^+_N$ that lies in the interior of $\Theta$, i.e., ${\nabla}_{\theta} \mathcal{J}^q_N(\theta^+_N) = 0$, then the sequence $\{\theta_{k}\}_{k\in\mathbb{N}}$  
    generated by recursion (\ref{eq:iter2_theta}) 
        \begin{align*}
        \mathbb{E}[\Vert \theta_k -\theta^+_N\Vert^2] = O_N(\gamma^{q}_k +(\gamma^{\theta}_k)^2(\gamma^{q}_k)^{-2})=O_N(k^{-\min\{\beta,2\gamma-2\beta\}}). \label{eq:mse_theta_rate2}
    \end{align*}
Moreover, if Assumption \ref{aspt:density}$^\ast$(a) holds, then the convergence rate above does not depend on $N$.
\end{theorem}

Theorem \ref{th:mse2_theta} implies that the QF-form achieves an optimal convergence rate of 
$O(k^{-\frac{2}{3}})$, which is faster than that of the DM-form as a result of the reduced timescales and the elimination of a kernel density estimation technique, provided that the hyperparameters are chosen to be feasible.

Finally, we establish the convergence results to the solution of the original problem. Here, Assumption \ref{aspt:jocobi}$^\ast$ strengthens Assumption \ref{aspt:jocobi} by requiring that the constant $\tilde{C}_\lambda$ involved is uniform in $N$. We still
specify the grid using a uniform partition, i.e., $z_i=(i+1)/(N+2)$, for $i=0,\cdots,N$.

\begin{theorem}\label{th:totalerror2}
Under the conditions of Theorem~\ref{th:main_convergence2}, with Assumption~\ref{aspt:density}(a) replaced by \ref{aspt:density}$^\ast$(a), we have $\lim_{N\to\infty} \sup_{\theta\in\Theta} \Vert \nabla_\theta \mathcal{J}_N^q(\theta) - \nabla_\theta \mathcal{J}(\theta)\Vert = 0$. Suppose $\theta^\ast\in\Theta^\circ$ is the unique stationary point (interior solution) of $\mathcal J(\theta)$. Then, it holds that
$$ \lim_{N\to\infty}\lim_{k\to\infty} \theta_k = \lim_{N\to\infty} \theta^+_N = \theta^\ast \quad a.s. $$
\end{theorem}
\begin{theorem}\label{th:totalerrorrate2}
Under the conditions of Theorem~\ref{th:mse2_theta}, with Assumptions~\ref{aspt:density}(a) and~\ref{aspt:jocobi} replaced by \ref{aspt:density}$^\ast$(a) and \ref{aspt:jocobi}$^\ast$, respectively, suppose that $\nabla_\theta F(q;\theta)\vert_{q=F^{-1}(z;\theta)}$ is uniformly Lipschitz in $z$, and $\mathcal{J}(\cdot)$ is locally strongly concave at the unique stationary point (interior solution) $\theta^\ast$. Then, the sequence $\{\theta_{k}\}_{ k \in \mathbb{N}}$ generated by recursion (\ref{eq:iter2_theta}) satisfy
\begin{align*}
        \mathbb{E}[\Vert \theta_k -\theta^\ast\Vert^2 ] =O\big(\gamma^{q}_k+(\gamma^{\theta}_k)^2 (\gamma^{q}_k)^{-2}\big)+O(N^{-2}) .
    \end{align*}
\end{theorem}

\section{The Hybrid Form\label{sec:hybrid}}

In this section, we present the hybrid algorithm that integrates DM and QF estimators.
The pseudocode is presented in Algorithm~\ref{alg:hybrid_drm}. 
The underlying idea of the hybrid algorithm is as follows: under a given grid, the domain can be partitioned into intervals that contain potential distributional jumps, and we collect the indices of these intervals into the set $\mathcal I_{\mathrm{jump}}$. In these intervals, we apply the DM estimator due to its robustness and superior performance around discontinuities. For the remaining intervals, where the distribution is smooth, with indices collected in the set $\mathcal I_{\mathrm{smooth}}=\{1,\cdots,N\}\setminus\mathcal I_{\mathrm{jump}}$,  we apply the QF estimator to leverage its efficiency for smooth regions. The algorithm then aggregates the updates from both the DM and QF estimators in a unified update step, ensuring stable and efficient convergence while respecting the local characteristics of the distribution under the DRM. 
This hybrid algorithm can be implemented efficiently because, under the grid, the computation of quantiles and their gradients within each interval is nearly independent of that in other intervals. 
From the theory developed in the previous sections, it is easily seen that under Assumptions \ref{aspt:diff_quantile}-\ref{aspt:density}, $\{\theta_k\} $ generated by the hybrid algorithm converges almost surely to the equilibria of the hybrid gradient proxy
\begin{equation*}
\begin{aligned}
\nabla_{\theta}\mathcal{J}^h_N(\theta)= \sum_{i\in\mathcal I_{\mathrm{jump}}}
& -  d_i(\theta) (\tilde{w}(z_i) - \tilde{w}(z_{i-1}) ) \\
&+\sum_{i\in\mathcal I_{\mathrm{smooth}}} \nabla_{\theta} F(q;\theta) \Big |_{q=F^{-1}({z}_i;\theta)} \tilde{w}'({z}_i) (F^{-1}(z_i;\theta) - F^{-1}(z_{i-1};\theta)).
\end{aligned}
\end{equation*}
We state the result formally below.
\begin{theorem}\label{th:main_convergence3} 
If Assumptions \ref{aspt:diff_quantile}-\ref{aspt:density} hold, then the sequence $\{\theta_k\}$ generated by Algorithm \ref{alg:hybrid_drm} converges a.s. to a stationary point of the projected ODE $\dot{\theta}(t)=\nabla_{\theta}\mathcal{J}^h_N(\theta(t))+p(t)$.
\end{theorem}

\begin{algorithm}[h]
\caption{Hybrid Form for DRM Optimization}
\label{alg:hybrid_drm}
\begin{algorithmic}[1]
  \State \textbf{Input:} Distortion function $\tilde w$, grid points $\{z_i\}$, index sets $\mathcal I_{\mathrm{jump}}$ and $\mathcal I_{\mathrm{smooth}}$, step size sequences $\{\gamma_k^{D}\},\{\gamma_k^{q}\},\{\gamma_k^{\theta}\}$, and band width sequence $\{h_k\}$
  \State \textbf{Initialize:} Parameters $\theta_0\in\Theta$, quantile estimtors $\{q_{0,i}\}$; and gradient estimators $\{D_{0,i}:i\!\in\!\mathcal I_{\mathrm{jump}}\}$
  \For{$k = 0,1,\ldots,K$}
      \State Sample $(X_k,Y_k)$ from the data-generating process under $\theta_k$
      \For {$i\in\mathcal I_{\mathrm{jump}} \cup \mathcal I_{\mathrm{smooth}}$ } 
      \State $ q_{k+1,i}= q_{k,i} + \gamma_k^{q}\!\bigl(
                 z_i - \mathbf 1\{Y_k\le q_{k,i}\}\bigr)$
      \EndFor
      \For{ $i\in\mathcal I_{\mathrm{jump}}$ } 
      \State $D_{k+1,i}= D_{k,i} + \gamma_k^{D}\!\bigl(
                 G_1(X_k;\theta_k,q_{k,i})
                 - G_3(Y_k;\theta_k,q_{k,i},h_k)\,D_{k,i}\bigr)$
      \EndFor
      \State Compute DM-form integral $g_k^{\text{DM}}=\sum_{i\in\mathcal{I}_{\mathrm{jump}}} 
                -D_{k+1,i}\,\bigl(\tilde w(z_i)-\tilde w(z_{i-1})\bigr)$
    \State Compute QF-form integral  $g_k^{\text{QF}}=\sum_{i\in\mathcal{I}_{\mathrm{smooth}}}
                G_1(X_k;\theta_k,q_{k,i})
                \tilde w'( z_i)
                (q_{k,i}-q_{k,i-1})$ 
      \State $\theta_{k+1}= 
             \varphi_{\Theta}\!\bigl(\theta_k + \gamma_k^{\theta}\,(g_k^{\text{DM}}+g_k^{\text{QF}})\bigr)$
  \EndFor
  \State \textbf{Output:} Final iterate $\theta_{K+1}$
\end{algorithmic}
\end{algorithm}

\section{Numerical Results}\label{section: Numerical Results}

In this section, we conduct two experiments to evaluate the performance of the three proposed algorithms. 
First, we consider a robust portfolio optimization problem as a classical stochastic optimization task, where the goal is to find the optimal distribution under a given distortion function and distributional constraint. This experiment aims to validate the superiority of our methods over the single-timescale baseline approach, highlight their respective characteristics, and demonstrate the advantage of the proposed hybrid algorithm compared with using either the DM- or QF-form algorithm alone.
In the second experiment, we explore a large-scale application in the context of deep RL. Specifically, we incorporate the proposed DRM optimization algorithm into the policy optimization framework and develop a DRM-based RL algorithm. We then conduct a multi-echelon dynamic inventory management example to illustrate the practical impact and significance of DRM-based optimization under varying risk profiles. This example also demonstrates that the proposed methods can effectively handle realistic risk-sensitive decision-making tasks in RL settings.

All experiments are conducted on a Linux server equipped with four NVIDIA RTX 4090 GPUs and two Intel\textsuperscript{\textregistered} Xeon\textsuperscript{\textregistered} Platinum 8352V CPUs (2.10GHz). 
To ensure a fair comparison, we adopt a unified set of core hyperparameter configurations across all algorithms, whenever applicable, as detailed below:
\begin{align*}
    \gamma^{D}_k = \frac{a}{(k_0 + k)^{\alpha}},\ \ \gamma^{q}_k = \frac{b}{(k_0 + k)^{\beta}},\ \ \gamma^{\theta}_k = \frac{r}{(k_0 + k)^{\gamma}},\ \ \text{and}\ \  h_k = \frac{h}{ (k_0 + k)^{\eta}},
\end{align*}
where $\alpha = 0.70$, $\beta = 0.71$, $\gamma = 0.99$, and $\eta = 0.14$, all of which satisfy Assumption \ref{aspt:step_size} and are close to the optimal values suggested by Theorem \ref{th:mse_theta}. It is worth noting that the QF-form allows for more aggressive hyperparameter choices, thereby achieving higher data efficiency as indicated in Theorem \ref{th:mse2_theta}. Others are selected through trial-and-error by pre-experiments. Please see Appendix \ref{appendix Experimental Settings} for detailed experimental settings.

\subsection{Robust Portfolio Selection with DRM Criteria}\label{Robust Portfolio Selection with DRM Criteria}

In robust portfolio selection, investors typically rely only on limited distributional information (e.g., known moments) to determine portfolio allocation. The key challenge lies in evaluating these portfolios under distributional uncertainty, which leads to an extreme-case risk measure problem \cite[see, e.g.,][]{zhu2009worst}. This problem can be formulated as
\begin{equation}\label{eq:hhkh}
 \max_{F\in\mathcal{P}} \int_0^1 F^{-1}(1 - z)  d w(z) ,
\end{equation}
where $\mathcal{P}$ denotes the set of distribution functions that satisfy certain moment constraints.
Note that (\ref{eq:hhkh}) is the non-parametric version of (\ref{eq:drm_def_2}). 
Such distributionally robust optimization with partial information has attracted sustained attention in 
multiple fields \cite[see, e.g.,][]{chen2011tight, lam2016robust} and also provides a rigorous ground truth for algorithmic evaluations.

In our experiments, we consider the common scenario in which only the first two moments are known, specifically setting the mean and variance equal to zero and one, respectively. To facilitate gradient-based optimization, we parametrize the problem within the family of Gaussian mixtures and impose the normalization explicitly to ensure that the resulting Gaussian mixture always satisfies the prescribed moment constraints. Let $\theta = \{(w_j, \mu_j, \ln\sigma_j)\}_{j=1}^d\in\mathbb{R}^{3d}$ denote trainable parameters. 
When evaluating, we first normalize the weights by Softmax transformation, i.e., $w'_j=\exp(w_j)/\sum_{j'=1}^d \exp(w_{j'})$, ensuring that the mixture weights are non-negative and sum to one. Next, we define the weighted mean $\mu_{\text{mix}} = \sum_{j=1}^n w'_j \mu_j$ and then centralize each component by $\tilde{\mu}_j = \mu_j - \mu_{\text{mix}}$. We compute the weighted variance before normalization as $\sigma_{\text{mix}}^2 = \sum_{j=1}^d w'_j (\sigma_j^2 + \tilde{\mu}_j^2)$. Finally, we standardize each component by setting $\mu_j' = \tilde{\mu}_j / \sigma_{\text{mix}}$ and $\sigma_j' = \sigma_j / \sigma_{\text{mix}}$, yielding the normalized Gaussian mixture $f(x;\theta) = \sum_{j=1}^d w'_j \cdot \mathcal{N}(x \mid \mu_j', (\sigma_j')^2)$ for sample generation, where $\mathcal{N}$ denotes the Gaussian density.
This parameterization transforms (\ref{eq:hhkh}) into a parametric type of problem (\ref{eq:drm_def_2}).
We then take the following DRMs as our testing instances:
\begin{itemize}
    \item[(a)] S-shape function with parameter $\alpha=5$, and $$ w_1(z;\alpha)=\frac{\exp(2\alpha z) -1 }{(\exp (\alpha) - 1)(\exp(2\alpha z-\alpha) +1)}; $$
        \item[(b)] Wang transformation \citep{wang2000class} with parameter $\alpha = -0.85$, and $$  w_2(z;\alpha)=\Phi(\Phi^{-1}(z)-\alpha); $$
       \item[(c)] CVaR with parameter $\alpha =0.7$, and $w_3(z;\alpha)=\min\{\frac{z}{1-\alpha},1\}$;
    \item[(d)] Discontinuous function with parameter $\alpha=5$, and
    $$ w_4(z;\alpha) = \frac{4}{5}w_1(z;\alpha) +\frac{1}{15} \big( \mathbf{1}\{z>0.3\} + \mathbf{1}\{z>0.5\} + \mathbf{1}\{z>0.7\}\big) . $$
\end{itemize}
The solution of (\ref{eq:hhkh}) under our moment conditions admits the closed form \citep{shao2024extreme}:
\begin{align}
    F^{-1}_\ast(z) = \big(\int_{0}^{1} \vert w'_*(1-z')-1\vert^2 dz'\big)^{-\frac{1}{2}} (w'_*(1-z)-1),  \label{eq:realsolu}
\end{align}
where $w_*$ is the concave envelope of the distortion function $w$.

Since no prior work has applied multi-timescale SA to DRM optimization, we construct a single-timescale baseline following our QF-form, which is the most numerically stable among forms discussed in Section \ref{section: Problem Formulation & Preliminaries}. It employs our recursion~(\ref{eq:iter2_theta}) but replaces all $q_{k,i}$ with the empirical quantiles estimated from the current batch of samples in each iteration. We then compare our three proposed algorithms (DM-form, QF-form, Hybrid-form) against the baseline (Batching). 
It is worth noting that our algorithms can operate effectively either with or without batching; however, in order for the empirical estimation in the baseline to function, we collect $4$ simulated samples as a batch for all algorithms before each update. Furthermore, we also test the baseline with $5$ (Batching$^+$) and $10$ (Batching$^*$) times the number of samples per batch, and the learning rate is scaled proportionally.

\begin{figure}[t]
\centering 
\includegraphics[trim=0cm 0.3cm 0cm 0.3cm, width=0.9\linewidth]{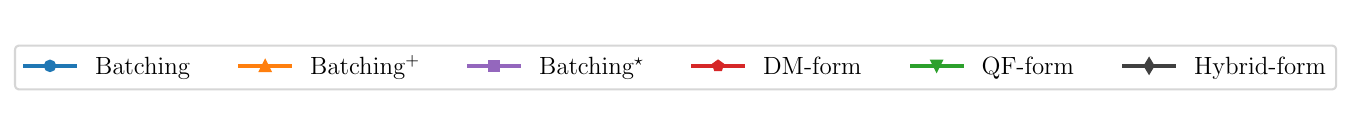}
\includegraphics[trim=0.2cm 0.cm 0.2cm 0.cm, clip, width=\linewidth]{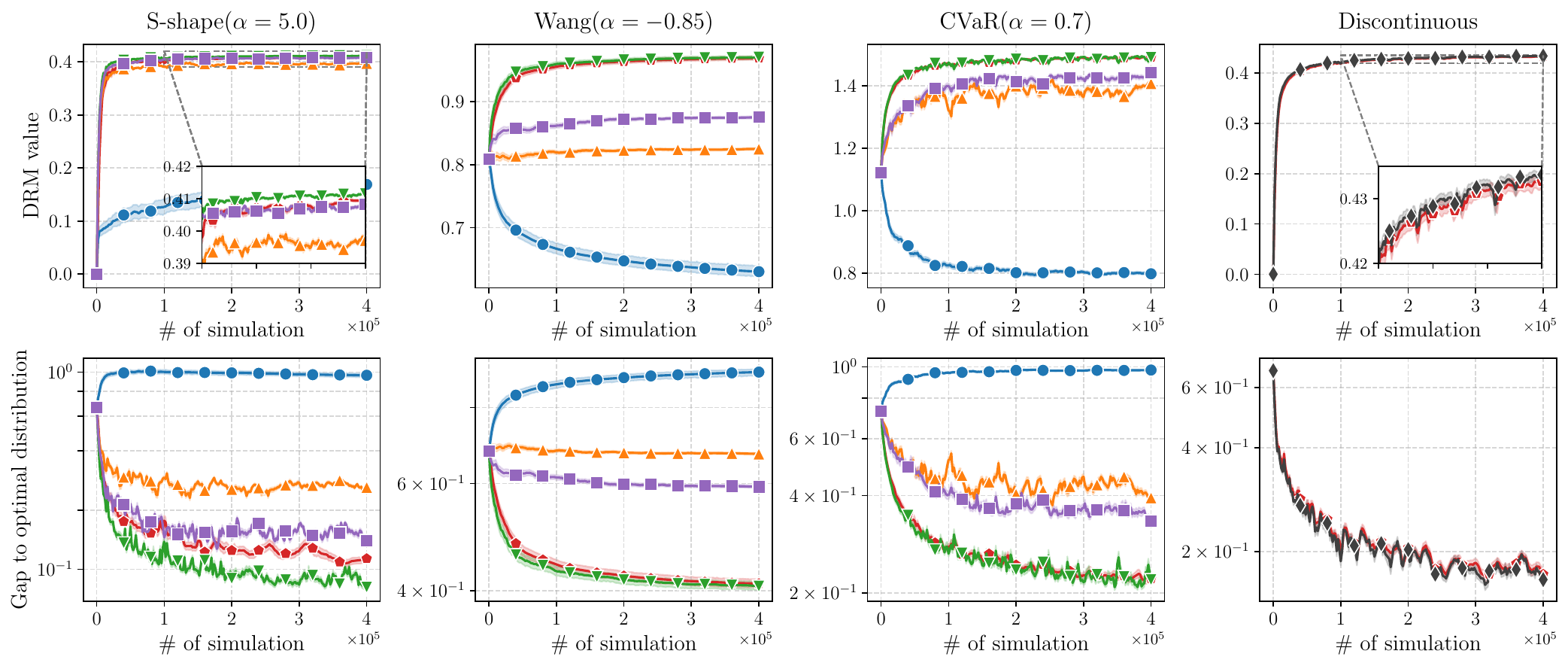}
\vspace{-0.5cm}
\caption{Visualization of the optimization procedure under different distortion functions. The top row shows DRM learning curves, and the bottom row shows the Wasserstein-2 distances between fitted and optimal quantile functions.
}
\label{fig:mtscompare}
\end{figure}

The corresponding DRM learning curves are shown in the first row of Figure \ref{fig:mtscompare}, and the second row reports the Wasserstein-2 distances between the quantile functions of the optimized model and the true solution (\ref{eq:realsolu}). The remaining gap is due to the distributional assumption when parameterizing the problem (\ref{eq:hhkh}). Each curve is based on 100 independent replications, with solid lines denoting the mean performance and shaded areas representing 95\% confidence bands. Because QF- and Hybrid-form are equivalent in the first three experiments, we omit the latter in those cases. In the last experiment, where discontinuous distortion functions render other methods inapplicable, only the results of DM- and Hybrid-form are reported. Across all settings, our proposed methods consistently and significantly outperform the single-timescale baselines.

It is critical that an accurate estimation of integral objectives like DRM requires a sufficiently fine discretization grid, which corresponds to the batch sizes in single-timescale SA. Small batch sizes may thus yield incorrect approximations of objectives, leading to incorrect behaviors such as the blue curves in Figure \ref{fig:mtscompare}, whereas large ones hinder the efficiency of SA \cite[see, e.g.,][]{FH}. This underscores the necessity of employing multiple recursions to carry historical information for optimization. The separation between approximation and estimation error, as demonstrated in Theorems \ref{th:totalerrorrate} and \ref{th:totalerrorrate2}, plays a crucial role in this regard, as it enables fast and reliable optimization without relying on batching. Our three algorithms exhibit comparable sample efficiency under identical settings, with minor differences due to the simpler structure and easier hyperparameter tuning of the QF-form and Hybrid-form. Moreover, the results highlight that both DM- and Hybrid-form variants are capable of addressing a wider class of problems, underscoring their flexibility and generality.

\begin{table}[t]
\caption{Comparison of mean computation time and $95\%$ confidence interval (CI) per iteration across proposed algorithms.}
\label{tab:alg_horizontal}
\centering
\setlength{\tabcolsep}{2pt}  
\renewcommand{\arraystretch}{1.0} 
\begin{adjustbox}{max width=\linewidth}
\begin{tabular}{cccccccc}
\toprule
$\#$ of grid  & $\#$ of model
& \multicolumn{2}{c}{DM-form} 
& \multicolumn{2}{c}{QF-form} 
& \multicolumn{2}{c}{Hybrid-form} \\
points ($N$) & parameters ($3d$) & Mean (ms) & 95\% CI (ms) & Mean (ms) & 95\% CI (ms) & Mean (ms) & 95\% CI (ms) \\
\midrule
\multirow{3}{*}{$1\times10^2$}   & $3\times10^1$    & 2.541 & [2.525, 2.556] & 1.080 & [1.078, 1.082] & 3.179 & [3.162, 3.196] \\
   & $3\times10^2$   & 3.823 & [3.817, 3.829] & 1.096 & [1.095, 1.096] & 3.186 & [3.185, 3.187] \\
   & $3\times10^3$  & 5.657 & [5.649, 5.665] & 1.692 & [1.689, 1.696] & 4.739 & [4.734, 4.745] \\
\midrule
\multirow{3}{*}{$1\times10^3$}   & $3\times10^1$    & 4.293 & [4.283, 4.303] & 1.140 & [1.139, 1.141] & 3.214 & [3.213, 3.215] \\
 & $3\times10^2$   & 6.026 & [6.017, 6.035] & 1.176 & [1.175, 1.176] & 3.278 & [3.277, 3.278] \\
 & $3\times10^3$  & 11.028 & [10.998, 11.058] & 1.782 & [1.779, 1.785] & 4.841 & [4.838, 4.844] \\
\bottomrule
\end{tabular}
\end{adjustbox}
\end{table}

To further highlight the differences among the three proposed algorithms, we also examine their computation time per iteration, as reported in Table \ref{tab:alg_horizontal}. All statistics are based on 10000 iterations. We vary both the number of grid points in the DRM integral and the number of parameters in the Gaussian mixture model, which influence the scale of the approximated problem. The results indicate that the computation time of QF-form is the least sensitive to the problem scale, while DM-form is the most affected. Hybrid-form introduces a minor, implementation-induced fixed overhead, which becomes negligible as the problem size increases. Ultimately, its computational cost depends on the number of discontinuous points on the distortion function rather than the total number of grid points, thereby achieving a substantial reduction in runtime compared with DM-form.

\subsection{Risk-sensitive Inventory Management via DRM-based Deep RL}\label{Risk-sensitive Inventory Management via DRM-based Deep RL}
To demonstrate the scalability of our method, we integrate the DRM optimization into deep RL, which is a general framework for solving Markov Decision Processes (MDPs), and propose a novel algorithm, namely DRM-based Proximal Policy Optimization (DPPO).
An MDP consists of the state space $\mathcal{S}$ and action space $\mathcal{A}$, as well as the transition kernel $p(s'|s, a)$, reward function $r(s, a, s')$, and reward discount factor $\eta\in(0,1)$. We consider the stochastic decision policy parameterized directly as a conditional density $\pi(\cdot|s;\theta)$ over $\mathcal{A}$ for every $s\in\mathcal{S}$. A trajectory $\tau=\{s_0,a_0,s_1,\cdots,s_T\}$ is generated from the interaction between the MDP dynamics and parametric policy, where $s_{t+1}\sim p(\cdot |s_{t}, a_{t})$, $a_t\sim\pi(\cdot| s_{t};\theta)$, $s_0$ follows some initializing density $p(\cdot)$, and $T$ is the time horizon. The discounted cumulative reward is written as $R(\tau)=\sum_{t=0}^{T-1}\iota^t r(s_{t+1},a_{t},s_{t})$, which can be viewed as a function of $\tau$ and follows a distribution denoted by $ F(\cdot;\theta)$.
With the score function technique \cite[see, e.g.,][]{jiang2023quantile}, we can rewrite the CDF gradient and obtain $G_1(\tau;\theta,q)$ in our algorithms as follows:
\begin{align*}
G_1(\tau;\theta,q) = -\mathbf{1}\{R(\tau)\leq q\}\sum_{t=0}^{T-1}\nabla_{\theta}\ln\pi(a_t|s_t;\theta), \text{ and } -\nabla_{\theta} F(q;\theta) =\mathbb{E}[G_1(\tau;\theta,q)] ,
\end{align*} 
compare with (\ref{eq:lr_111}).
To further increase the sample efficiency of DRM optimization, we apply a similar IS technique to that used in the state-of-the-art mean-based deep RL algorithms  \citep{schulman2017proximal}, which allows multiple parameter updates before generating new simulated data. 
We define $[k]=K_0\lfloor\frac{k}{K_0}\rfloor$, where $K_0>0$ is an integer, meaning that trajectories are generated only at the $[k]$-th iteration and are reused for parameter updates in the subsequent $K_0$ iterations.
An IS ratio, defined as 
$
\rho_k=\frac{\Pi(\tau_{[k]};\theta_k)}{\Pi(\tau_{[k]};\theta_{[k]})}=\prod_{t=0}^{T-1}\frac{\pi(a_{[k],t}|s_{[k],t};\theta_k)}{\pi(a_{[k],t}|s_{[k],t};\theta_{[k]})}
$
is used to correct the updates for the distributional discrepancy induced by the policy update in a probabilistic sense \citep{glynn1989importance}. 
Since the IS technique may lead to increased estimation variance when $\theta_k$ significantly differs from $\theta_{[k]}$, we also follow the mechanism in \citep{schulman2017proximal} and leverage the IS ratio as a criterion to determine when to regenerate samples, and trigger new sample collection whenever $\rho_k\notin [1-\varepsilon,1+\varepsilon]$, where $\varepsilon$ is the tolerance. DPPO is designed on top of the Hybrid-form framework in Algorithm \ref{alg:hybrid_drm}, ensuring high efficiency while remaining compatible with various DRMs. A complete pseudocode of DPPO is provided in Algorithm \ref{alg:hybrid_drmrl}.

\begin{algorithm}[h]
\caption{DRM-based Proximal Policy Optimization (DPPO)}
\label{alg:hybrid_drmrl}
\begin{algorithmic}[1]
  \State \textbf{Input:} Policy network $\pi(\cdot|\cdot;\theta)$, sampling interval $K_0$, distortion function $\tilde w$, grid points $\{z_i\}$, index sets $\mathcal I_{\mathrm{jump}}$ and $\mathcal I_{\mathrm{smooth}}$, step size sequences $\{\gamma_k^{D}\},\{\gamma_k^{q}\},\{\gamma_k^{\theta}\}$, and band width sequence $\{h_k\}$
  \State \textbf{Initialize:} Parameters $\theta_0\in\Theta$, quantile estimtors $\{q_{0,i}\}$; and gradient estimators $\{D_{0,i}:i\!\in\!\mathcal I_{\mathrm{jump}}\}$
  \For{$k = 0,1,\ldots,K$}
      \If { $k=[k]$ } 
      \State Generate one episode $\tau_k=\{s_0^k,a_0^k,\cdots,s_k^T\}$ following policy $\pi(\cdot|\cdot;\theta_k)$
      \EndIf
      \State Compute the IS ratio $\rho_k$
      \If {$1-\varepsilon\leq \rho_k \leq 1+\varepsilon$}
      \For{ $i\in\mathcal I_{\mathrm{jump}} \cup \mathcal I_{\mathrm{smooth}}$  }
      \State $ q_{k+1,i}= q_{k,i} + \gamma_k^{q}\!\bigl(
                 z_i - \rho_k\mathbf 1\{R(\tau_{[k]})\le q_{k,i}\}\bigr)$
      \EndFor
      \For{ $i\in\mathcal I_{\mathrm{jump}}$  } 
      \State $ D_{k+1,i}= D_{k,i} + \gamma_k^{D}\rho_k\!\bigl(
                 G_1(\tau_{[k]};\theta_k,q_{k,i})
                 - G_3(R(\tau_{[k]});\theta_k,q_{k,i},h_k)\,D_{k,i}\bigr) $
      \EndFor
      \State Compute DM-form integral $g_k^{\text{DM}}=\sum_{i\in\mathcal{I}_{\mathrm{jump}}} 
                -D_{k+1,i}\,\bigl(\tilde w(z_i)-\tilde w(z_{i-1})\bigr)$
    \State Compute QF-form integral  $$ g_k^{\text{QF}}=\rho_k\sum_{i\in\mathcal{I}_{\mathrm{smooth}}}
                G_1(\tau_{[k]};\theta_k,q_{k,i})
                \tilde w'(z_i)
                (q_{k,i}-q_{k,i-1}) $$ 
      \State $\theta_{k+1}= 
             \varphi_{\Theta}\!\bigl(\theta_k + \gamma_k^{\theta}\,(g_k^{\text{DM}}+g_k^{\text{QF}})\bigr)$
    \EndIf\EndFor
  \State \textbf{Output:} Final iterate $\theta_{K+1}$
\end{algorithmic}
\end{algorithm}

We consider the multi-echelon dynamic inventory management problem within a supply chain during $T$ periods, where the two ends represent the customer and manufacturer. We use $S_t^j$, $U_t^j$ and $I_t^j$ to denote per echelon $ j $ the products shipped, the lost sales, and on-hand inventory  at the end of period $t$. The reordering quantity of echelon $j$ is denoted as $Q_t^j$, with $Q_t^0$ as the randomly generated exogenous demand.
In each period $t$, the agent has to select the best replenishment quantities $Q_t=\{Q_t^j\}_{j=1}^M$ for intermediate echelons. The shipped quantities are given by $S_t^j=Q_t^{j-1}-[Q_t^{j-1}-I_{t-1}^j-S_{t-L^j}^{j+1}]^{+}$, for $j=1,\cdots, M$, where $[x]^+=\max\{0,x\}$, $L_j$ denotes the transit duration of echelon $j$, and the manufacturer is assumed to satisfy all demands from the last echelon, i.e., $S_t^{M+1} = Q_t^{M}$.
Then the lost sales and on-hand inventory are computed by $U_t^j=[Q_t^{j-1}-S_t^j]^{+}$ and $I_t^j=[I_{t-1}^j+S_{t-L^j}^{j+1}-S_t^j]^{+}$. The profit of each echelon is calculated by $P_t^j=p^j S_t^j-p^{j+1} S_t^{j+1}-h^j I_t^j-l^j U_t^j$,
where $p^j$, $h^j$, and $l^j$ are the unit price, holding cost, and stock-out penalty, respectively. The RL states cover the inventories, lost sales, shipped and ordering quantities in past $L=\max_j L_j$ periods, and the step rewards are determined by current overall profits, i.e., $\sum_{j=1}^M P_t^j$.

\begin{figure}[tb]
\centering 
\subfigure[0.1-quantile curves.]{
\label{fig:3a}
\includegraphics[trim=0.35cm 0.35cm 0.35cm 0.1cm,  width=0.3\linewidth]{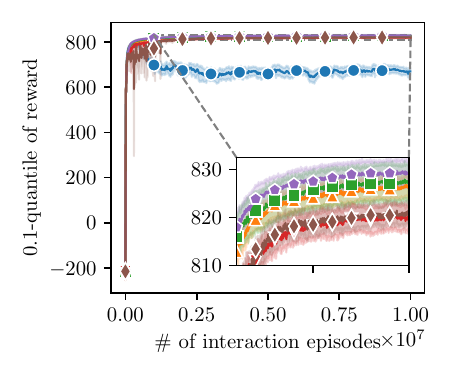}}\hfill
\subfigure[0.5-quantile curves.]{
\label{fig:3b}
\includegraphics[trim=0.35cm 0.35cm 0.35cm 0.1cm,  width=0.3\linewidth]{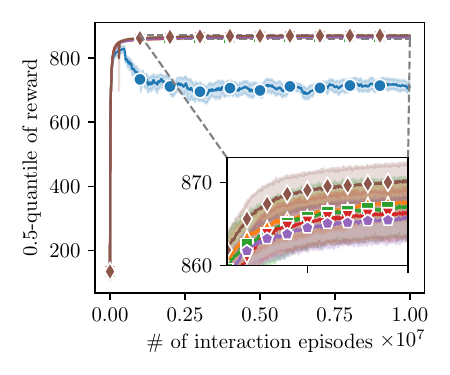}}\hfill
\subfigure[0.9-quantile curves.]{
\label{fig:3c}
\includegraphics[trim=0.35cm 0.35cm 0.35cm 0.1cm,  width=0.3\linewidth]{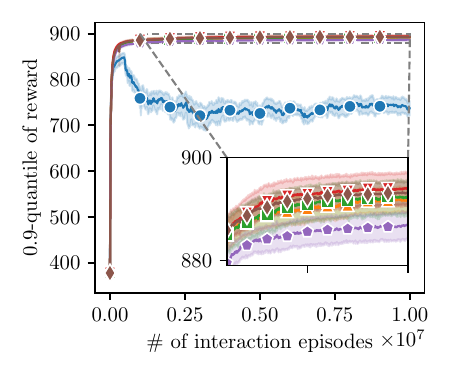}}\hfill
\subfigure[Mean curves.]{
\label{fig:3d}
\includegraphics[trim=0.35cm 0.35cm 0.35cm 0.1cm,  width=0.3\linewidth]{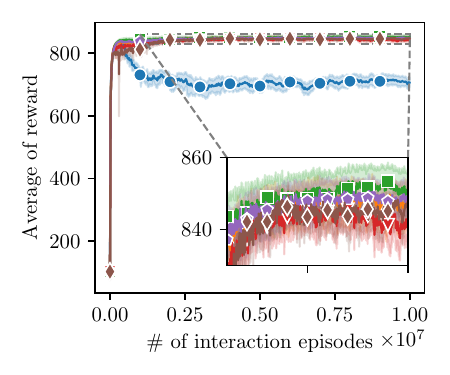}}\hfill
\subfigure[KDE plots.]{
\label{fig:3e}
\includegraphics[trim=0.35cm 0.5cm 0.35cm 0.1cm,  width=0.305\linewidth]{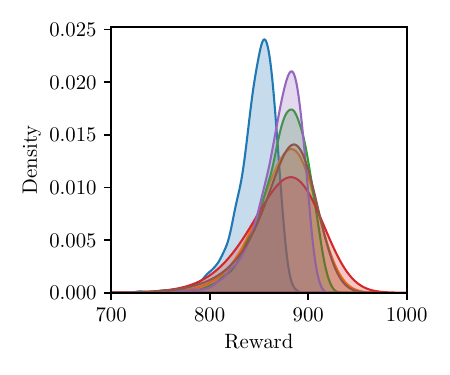}}\hfill
\includegraphics[trim=-0.8cm 0.cm 0.cm 0.cm,  width=0.3\linewidth]{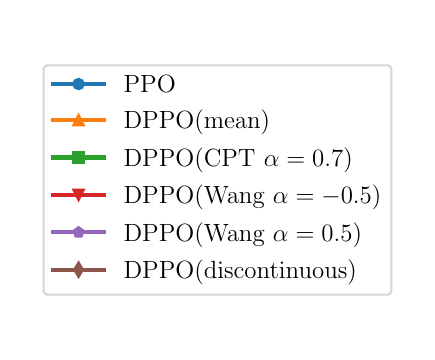}
\caption{Performance visualization of PPO and DPPO variants in the inventory management example.
}
\label{fig:3}
\end{figure}

We compare the performance of several DPPO variants with different distortion functions against standard Proximal Policy Optimization \cite[PPO,][]{schulman2017proximal}, which is the most widely used baseline in deep RL. Both PPO and DPPO (mean) are risk-neutral, aiming to maximize the expected return. DPPO (CPT) variant also symmetrically emphasizes both tails of the distribution but assigns greater weight to extreme outcomes, using the distortion function $w_5(z;\alpha)=\frac{z^\alpha}{(z^\alpha+(1-z)^\alpha)^{1/\alpha}}$. The DPPO (Wang) variants, with parameters $\alpha=-0.5$ and $\alpha=0.5$, correspond to risk-seeking and risk-averse preferences, respectively. The DPPO (discontinuous) variant employs the distortion function (d) introduced in Section \ref{Robust Portfolio Selection with DRM Criteria}, which focuses on outcomes near the median and places additional weight on the $0.3$-, $0.5$-, and $0.7$- quantiles.

Figures \ref{fig:3a}-\subref{fig:3d} show learning curves averaged over 20 independent runs, with solid lines indicating the averaged outcome and shaded areas representing 95\% confidence intervals. All DPPO variants consistently outperform PPO. While PPO exhibits noticeable improvement during the early stage of training, its performance tends to deteriorate as training approaches the optimal level. This decline is commonly associated with the reliance on a critic network to approximate the expected return given a state–action pair, which introduces instability due to the nonconvex nature of the underlying optimization subproblem. In contrast, DPPO tracks auxiliary variables by solving convex quadratic programs for quantiles and their gradients, which admits a unique solution and thereby avoids such numerical instabilities.

Internal comparisons among DPPO variants reveal that risk preferences can substantially influence the learning efficiency of deep RL. In Figure \ref{fig:3a}, the risk-averse DPPO (Wang, $\alpha=0.5$) achieves the highest learning efficiency at the 0.1-quantile, whereas the risk-seeking DPPO (Wang, $\alpha=-0.5$) exhibits the slowest. The opposite pattern is observed in Figure \ref{fig:3c}.
In Figure \ref{fig:3b}, DPPO(discontinuous), which emphasizes central outcomes, exhibits the steepest performance gain at the median. Figure \ref{fig:3d} shows that CPT and Wang ($\alpha=-0.5$), both of which place greater emphasis on the left tail of the outcome distribution, achieve faster improvement when optimizing the mean return. This phenomenon will be further explained later, where it is linked to the characteristics of the multi-echelon inventory problem.

Figure \ref{fig:3e} visualizes the performance of the best policy from each of the 20 training runs, evaluated under each algorithm’s respective risk measure. Kernel density estimates (KDEs) based on 10,000 simulated episodes reveal that DPPO variants such as Wang ($\alpha=0.5$) and CPT, which place greater emphasis on tail losses, yield return distributions with substantially lower variance, whereas that from the risk-seeking Wang ($\alpha=-0.5$) exhibits the largest spread. Notably, all return distributions are right-skewed, reflecting the structural properties of the inventory problem: profits are limited by the demand distribution, whereas losses from overstocking can be unbounded. This asymmetry provides a natural explanation for the superior learning efficiency of metrics that emphasize the left tail of the distribution.

\begin{table}[th]
\caption{Best performance comparison of PPO and DPPO variants in the inventory management example.}
\label{tab:deep rl}
\centering
\setlength{\tabcolsep}{2pt}  
\renewcommand{\arraystretch}{1.0} 
\small 
\begin{adjustbox}{max width=\linewidth}
\begin{tabular}{lccccccc}
\toprule
Algorithm & Mean & 0.1-quantile & 0.3-quantile & 0.5-quantile & 0.7-quantile & 0.9-quantile\\
\midrule
PPO  & 825.43 & 797.45 & 818.21 & 829.14 & 838.20 & 849.07\\
DPPO (mean)  & 849.32 & 827.01 & 853.76 & 867.91 & 879.34 & 891.77\\
DPPO (CPT $\alpha=0.7$)  & \textbf{853.07} & 828.14 & 853.46 & 867.27 & 878.90 & 892.19\\
DPPO (Wang $\alpha=-0.5$)  & 843.16 & 820.38 & 850.34 & 866.63 & 879.72 & \textbf{894.11}\\
DPPO (Wang $\alpha=0.5$)  & 851.71 & \textbf{830.10} & 853.74 & 865.87 & 875.76 & 886.73\\
DPPO (discontinuous)  & 848.18 & 822.05 & \textbf{855.16} & \textbf{870.37} & \textbf{881.57} & 893.15
\\
\bottomrule
\end{tabular}
\end{adjustbox}
\end{table}

We further summarize the performance of trained agents as shown in Table \ref{tab:deep rl}. For each algorithm, we first identify the best-performing policy from each independent training run, then evaluate it over 10000 simulated episodes and report the average performance across runs. Under the mean criterion, DPPO (CPT) achieves the highest value due to the distributional characteristics of the problem.  DPPO (Wang) variants achieve the best performance at the 0.1- and 0.9-quantiles, respectively, consistent with their corresponding risk preferences, whereas DPPO (discontinuous) achieves optimal performance at the $0.3$-, $0.5$-, and $0.7$-quantiles. This outcome aligns with the additional weights that the discontinuous distortion function assigns to these quantiles, thereby validating the ability of our approach to effectively optimize even in the presence of discontinuous distortion functions. Overall, these findings corroborate our theoretical predictions and underscore the importance of incorporating distribution-sensitive risk measures in sequential decision-making scenarios.

\section{Conclusions}\label{section: Conclusions}
In this paper, we propose efficient multi-timescale SA methods for optimizing objective functions formulated in integral forms, with a particular focus on DRM optimization. We derive two distinct forms of the DRM gradient and design corresponding algorithms with complementary characteristics, which are ultimately integrated into a more powerful hybrid algorithm. The theoretical contributions of this work include rigorous proofs of strong convergence and convergence rates for the proposed algorithms. Numerical experiments demonstrate that our family of algorithms substantially outperforms baseline methods in classical financial scenarios. To further validate scalability on large-scale and complex tasks, we extend our approach to deep RL and introduce a risk-sensitive algorithm, DPPO, which effectively controls output uncertainty and exhibits strong practical relevance in inventory management applications.
The codes that support this study can be found in \url{https://github.com/JinyangJiangAI/SA-Methods-for-DRM-Optimization}.

\acks{This work was supported in part by the National Natural Science Foundation of China (NSFC) under Grants 72325007, 72250065, and 72022001, as well as the Key Project of Xiangjiang Laboratory under Grant 3XJ02004.
The work of Jiaqiao Hu was supported by the U.S. National Science Foundation under Grant CMMI-2027527.}

\newpage

\appendix

\section{Gradient Expressions \label{sec:yije}}

We derive the following gradient estimators based on the GLR \citep{peng2020maximum} method:
\begin{align*}
    -\nabla_{\theta}F(y;\theta) = \mathbb{E}[G_1(X;\theta,y)],\quad f(y;\theta) = \mathbb{E}[G_2(X;\theta,y)],
\end{align*}
where $x_i$ is an entry arbitrarily chosen from $x$ and 
\begin{align}
     & G_1(x;\theta,y) \nonumber \\
     & = \mathbf{1}\{\mathcal{L}(x;\theta)\leq y\} 
     \left (-\nabla_{\theta}\ln f_X(x;\theta) + \left (\frac{\partial \mathcal{L}(x;\theta)}{\partial x_i}\right )^{-1}  \right . \nonumber\\
     & \qquad \left. \left [ \left (
     \frac{\partial \ln f_X(x;\theta)}{\partial x_i} - \frac{\partial^2 \mathcal{L}(x;\theta)}{\partial x_i^2} \left (\frac{\partial \mathcal{L}(x;\theta)}{\partial x_i}\right )^{-1} \right )
     \nabla_{\theta}\mathcal{L}(x;\theta) + \frac{\partial \nabla_{\theta}\mathcal{L}(x;\theta)}{\partial x_i} 
     \right ] \right ), 
     \label{eq:glr_1}\\
     & G_2(x;\theta,y)  \nonumber \\
     & =  \mathbf{1}\{\mathcal{L}(x;\theta)\leq y\} \left(\frac{\partial \mathcal{L}(x;\theta)}{\partial x_i}\right )^{-1} \left (\frac{\partial \ln f_X(x;\theta)}{\partial x_i} - \frac{\partial^2 \mathcal{L}(x;\theta)}{\partial x_i^2} \left(\frac{\partial \mathcal{L}(x;\theta)}{\partial x_i}\right)^{-1} \right), \label{eq:glr_2}
\end{align}
which are based on strong differentiability of $\mathcal{L}(\cdot;\theta)$.

\section{Proofs of Theorems}\label{sec:22}

Let $(\Omega,\mathcal{F},P)$ be a probability space. We simplify the notations $\{q_{k,i}\}_{i=1}^N$ and $\{D_{k,i}\}_{i=1}^N$ as $q_k$ and $D_k$, respectively, and define $\mathcal{F}_{k}=\sigma\{\theta_0,q_0 ,\cdots,\theta_k,q_k \}$, for $k=0,1,\cdots$, as the filtration generated by our algorithms. When applicable, the history of $D_k$ is also included in the filtration.

\subsection{Auxiliary Results}

The following lemma supports the characterization of the limiting behavior of the separate quantile gradient estimators.

\begin{lemma}\label{lem:ode_d}
If Assumption \ref{aspt:diff_quantile} hold, then for all $\bar{\theta}\in\Theta$, and $i=1,\cdots,N$, $ \nabla_\theta F^{-1}({z}_i;\bar{\theta})$ is the unique global asymptotically stable equilibrium of the ODE 
\begin{align}
    \dot{D}_i(t) = -\nabla_\theta F(F^{-1}({z}_i;\bar{\theta});\theta)\big|_{\theta=\bar{\theta}} - f(F^{-1}({z}_i;\bar{\theta});\bar{\theta})D_i(t). \label{eq:ode_d}
\end{align}
\end{lemma}
\proof 
With Assumption \ref{aspt:diff_quantile} and $f(\cdot;\theta)$ being positive and continuous, the gradient $$\nabla_\theta F^{-1}({z}_i;\bar{\theta}) = - \nabla_\theta F(F^{-1}({z}_i;\bar{\theta});\theta)|_{\theta=\bar{\theta}} \big/ f(F^{-1}({z}_i;\bar{\theta});\bar{\theta})$$ is well-defined and the unique solution of $\dot{D}_i(t) = 0$, i.e., the unique equilibrium point of ODE (\ref{eq:ode_d}). Consider the Lyapunov function $V_1(x)=\Vert x -\nabla_\theta F^{-1}({z}_i;\bar{\theta}) \Vert^2$, and the derivative $\dot{V_1}(x)= 2(x - \nabla_\theta F^{-1}({z}_i;\bar{\theta}))^{\top} (-\nabla_\theta F(F^{-1}({z}_i;\bar{\theta});\theta)\big|_{\theta=\bar{\theta}} - f(F^{-1}(z_i;\bar{\theta});\bar{\theta})x)$ is negative for any $x\neq \nabla_\theta F^{-1}(z_i;\bar{\theta})$. Hence, it is global asymptotically stable by the Lyapunov Stability Theory \citep{liapounoff2016probleme}.
\endproof

For the proofs of the main convergence theorems, we rely on the following convergence theorem for single-timescale SA with bias term, which is extended from Theorem 5.2.1 in \cite{kushner2003stochastic}. The assumptions required are largely decomposed and simplified for a more flexible application.
For the projected ODE, we denote by $ - C ( x )$, for $ x \not \in \Theta$, the set of vectors that project $ x $ onto $ \Theta$.

\begin{theorem}\label{th:single_convergence}
Consider a recursion $x_{k+1}=x_k + \gamma_k (h(x_k)+m_k+b_k)$, where $h(\cdot)$ is continuous and $\sup_k \Vert h(x_k)\Vert <\infty$ w.p.1; $\{\gamma_k\}$ satisfies $\gamma_k>0$, $\sum_k \gamma_k =\infty$; $\{m_k\}$ is a martingale difference sequence such that $\mathbb{E}[m_{k+1}|\mathcal{F}_k']=0$ for $k\geq0$, and $\sum_{k=0}^{\infty} \gamma_k^2\mathbb{E}[\Vert m_{k+1}\Vert^2]<\infty$, with sigma fields denoted as $\mathcal{F}_k'=\sigma\{x_0,\cdots,x_k\}$; and $b_k\rightarrow 0$ w.p.1. If the generated sequence $\{x_k\}$ is bounded, then $\{x_k\}$ converges a.s. to some limit set of the ODE $\dot{x}(t) = h(x(t))$ or the unique global asymptotically stable equilibrium if one exists. Otherwise, a projection onto a given compact convex set can be applied to ensure boundedness, in which case $\{x_k\}$ converges to counterparts of the projected ODE.
\end{theorem}

\proof The proof proceeds in two main steps.

\textbf{Convergence without projection.}
Let $t_k=\sum_{r=0}^{k-1} \gamma_r$, $t_0=0$ and $[t]=\max\{k: t_k\leq t\}$. Define the piecewise constant interpolation $x^0(\cdot)$ on $\mathbb{R}$ by $x^0(t)=x_0$ for $t<0$; and $x^0(t)=x_k$ for $t\in[t_k,t_{k+1})$. 
Denote the shifted processes $x^k(\cdot)$ as $x^k(t)=x^k(t_k+t)$ on $\mathbb{R}$, i.e., 
$x^k(t) = x_k + \sum_{r =k}^{[t_k+t]-1} \gamma_r h(x_r) + m^k(t) + b^k(t) = x^k(0) + \int_{0}^t h(x^k(s))ds + m^k(t) + b^k(t) + \rho^k(t),$
where $\rho^k(t) = \sum_{r =k}^{[t_k+t]-1} \gamma_r h(x_r) - \int_{0}^t h(x^k(s))ds$, $m^k(t)=\sum_{r =k}^{[t_k+t]-1} \gamma_r m_r$, and  $b^k(t)=\sum_{r =k}^{[t_k+t]-1} \gamma_r b_r$. 
Since $\Vert \rho^k(t)\Vert = \Vert \int_{\sum_{r =k}^{[t_k+t]-1} \gamma_r}^t h(x^k(s))ds\Vert<\gamma_{[t_k+t]}\sup_k \vert h(x_k)\vert$, $\rho^k(t)$ converge to zero uniformly on each prefixed bounded interval $[0,T]$ as $k\rightarrow\infty$ w.p.1.
Let $M_k = \sum_{r =0}^{k-1}\gamma_r m_r$. Note that $\mathbb{E}[ m_r m_s] =\mathbb{E}[ m_r \mathbb{E}[m_s \vert \mathcal{F}_r']] = 0$, for all $r < s$. Thus, $\sup_k \mathbb{E}[\Vert M_k\Vert^2] = \sup_k\mathbb{E}\big[\sum_{r=0}^{k-1} \gamma_r^2 \mathbb{E}[\Vert m_{r}\Vert^2\vert \mathcal{F}_r']\big] \leq \sum_{k=0}^{\infty} \gamma_k^2\mathbb{E}[\Vert m_{k+1}\Vert^2]<\infty$, and $\{M_k\}$ is an $L^2$-bounded martingale sequence. From the martingale convergence theorem \citep{durrett2019probability}, $\{M_k\}$ converges a.s., thereby $m^k(t) =  M_{[t_k+t]} - M_k$  converge to zero uniformly on each bounded interval as $k\rightarrow\infty$ w.p.1.
Note that $b_k\rightarrow 0$ w.p.1, a similar result can be obtained for $b^k(t)$.
Along with the boundedness of $\{x_k\}$ and the fact that $\sup_k \vert h(x_k)\vert<\infty$ a.s., we can conclude that $\{x^k(\cdot)\}$ is equicontinuous in the extended sense, yielding a.s. convergence of its subsequence $\{x^{k_r}(\cdot)\}$ to some continuous limit $x(\cdot)$ by Arzelà–Ascoli Theorem \cite[see, e.g.,][]{royden2010real}.
With Lebesgue Dominated Convergence Theorem \cite[see, e.g.,][]{bartle2014elements} and the continuity of $h(\cdot)$, we have
$ x(t) = x(0) + \lim_{k_r\rightarrow\infty} \int_{0}^t h(x^{k_r}(s))ds = x(0) + \int_{0}^t h(x(s))ds$, which admits the ODE $\dot{x}(t)=h(x(t))$.
Because the continuous-time process $\{x^k(\cdot)\}$ is constructed
by piecewise constant interpolation of $\{x_k\}$, $\{x_k\}$ must converge to some limit set of the ODE $\dot{x}(t)=h(x(t))$ w.p.1. If the ODE has a unique global asymptotically stable equilibrium, then the limit set reduces to this single point. 

\textbf{Extension to the projected case.} When the boundedness of $\{x^k\}$ is ensured by a projection onto a given compact convex set $A$, we rewrite the recursion as $x_{k+1}=x_k + \gamma_k (h(x_k)+m_k+b_k+p_k)$, where $p_k\in-C(x_{k+1})$, and $\gamma_kp_k$ is the minimal force required to keep $x_{k+1}$ within $A$. The shifted process $x^k(t)$ contains an extra term $p^k(t) = \sum_{r =k}^{[t_k+t]-1} \gamma_r p_r$.
Now, our main task is to establish the equicontinuity of the projection term.
Since $p_k\in-C(x_{k+1})$ and $x_k\in A$, we have $0\leq p_k^{\top}(x_k-x_{k+1})=\gamma_{k}p_k^{\top}(-  (h(x_k)+m_k+b_k)-p_k)\leq \gamma_{k}(\Vert p_k\Vert  \Vert h(x_k)+m_k+b_k\Vert -
\Vert p_k\Vert^2)$, which implies $ \Vert p_k\Vert\leq \Vert h(x_k)\Vert+\Vert m_k\Vert+\Vert b_k\Vert$.
Note that by Markov's Inequality, we have
$\sum_{k=1}^{\infty}P(\gamma_k\Vert m_k\Vert > \epsilon) \leq  \epsilon^{-2} \sum_{k=1}^{\infty} \gamma_k^2 \mathbb{E}[\Vert m_k\Vert^2]<\infty$.
By the Borel-Cantelli Lemma, we conclude that
$P(\gamma_k\Vert m_k\Vert > \epsilon \text{ i.o.}) = 0$.
This means a.s., for all sufficiently large $ k $, $ \gamma_k\Vert m_k\Vert \leq \epsilon $. Since $ \epsilon > 0 $ is arbitrary chosen, it follows that $\gamma_k\Vert m_k\Vert \to 0$ as $k\to\infty$ w.p.1.
Therefore, together with $\sup_k\Vert h(x_k)\Vert <\infty$ and $b_k\to 0$ w.p.1, $\gamma_k\Vert p_k\Vert \to 0$ as $k\to\infty$ w.p.1.
Suppose $\{p^k(\cdot)\}$ is not equicontinuous on $[0,T]$ for a constant $T>0$. Then, we have by definition that there exists a lower bound $\varepsilon_p>0$, integers $\{k_s\}$ and bounded durations $\{\delta_s, \xi_s\}$, where $k_s\rightarrow\infty$, $\delta_s\in[0,T]$ and $\xi_s\rightarrow 0+$ as $s\rightarrow\infty$, such that
$0<\varepsilon_p\leq\Vert p^{k_s}(\delta_s+\xi_s)-p^{k_s}(\delta_s)\Vert \leq \sum_{r = [t_{k_s}+\delta_s]}^{[t_{k_s}+\delta_s+\xi_s]-1} \gamma^{\theta}_r \Vert p_r\Vert \rightarrow 0$,
as $s\rightarrow\infty$, since $\xi_s\rightarrow 0+$ as $s\to \infty$, and $\gamma_r\Vert p_r\Vert\to0$ a.s. as $r\rightarrow\infty$. This raises a contradiction and suggests that $\{P^k(\cdot)\}$ is equicontinuous in the extended sense, and consequently, so is $\{x^k(\cdot)\}$.
By arguing as in the proof without projection, we conclude that the sequence $\{x_k\}$ converges to some limit set of the projected ODE $\dot{x}(t)=h(x(t))+p(t)$ w.p.1, which completes the proof.
\endproof

The main difficulty of applying the result of Theorem \ref{th:single_convergence} lies in showing the boundedness of the sequences generated by recursions (\ref{eq:iter_d})-(\ref{eq:iter_theta}) or (\ref{eq:iter2_q})-(\ref{eq:iter2_theta}), which is verified in Lemma \ref{lem:q_stability} and Theorem \ref{th:d_stability}. The following lemma follows directly from noting that the update in (\ref{eq:iter_q}) is bounded by one.

\begin{lemma}
\label{lem:q_stability}
If Assumptions \ref{aspt:diff_quantile} and \ref{aspt:step_size}(b) hold, then the sequence $\{q_k\}$ generated by recursion (\ref{eq:iter_q}) or (\ref{eq:iter2_q}) is bounded w.p.1, i.e., $\sup_{k,i}\lvert q_{k,i}\rvert < \infty$ w.p.1. Moreover, if Assumption \ref{aspt:density}(a) also holds, then the sequence is square-integrable, i.e., $\sup_{k,i}\mathbb{E}[\lvert q_{k,i}\rvert^2] < \infty$.
\end{lemma}
\proof  
Note that the pathwise boundedness of the scalar sequence $\{q_{k,i}\}_k$ follows from Assumption \ref{aspt:step_size}(b), the essential boundedness of the CDF, and the boundedness of the ${z}_i$-quantile function, which is primarily ensured by the continuity of the quantile function over the compact parameter set $\Theta$ \citep{jiang2023quantile}.
For square integrability, we take the square and conditional expectation of the recursion (\ref{eq:iter_q}) or (\ref{eq:iter2_q}), i.e.,
\begin{align*}
    \mathbb{E}[q_{k+1,i}^2|\mathcal{F}_k]&\leq q_{k,i}^2+2\gamma^q_k q_{k,i}(F(F^{-1}(z_i;\theta_k);\theta_k)-F(q_{k,i};\theta_k)) + (\gamma_k^q)^2 \\
    &= (1-2\gamma^q_kf(\hat{q};\theta_k))q_{k,i}^2+2\gamma^q_k q_{k,i}f(\hat{q};\theta_k)F^{-1}(z_i;\theta_k) + (\gamma_k^q)^2\\
    &\leq (1-\gamma^q_k\varepsilon_f^N)q_{k,i}^2+\gamma^q_k C_f\sup_{\theta\in\Theta}(F^{-1}(z_i;\theta))^2  + (\gamma_k^q)^2,
\end{align*}
where the equality follows from the intermediate value theorem and the second inequality is justified by Assumption \ref{aspt:density}(a). By Assumption \ref{aspt:step_size}(b), there exists an integer 
$K_{1}>0$ such that $\gamma_k^q\varepsilon_f^N\in(0,1)$ for $k>K_{1}$.
Then, by taking expectation we have for $k>K_{1}$ that 
\begin{align*}
    \mathbb{E}[q_{k+1,i}^2]
    &\leq (1-\gamma^q_k\varepsilon_f^N) \mathbb{E}[q_{k,i}^2]+\gamma^q_k C_f\sup_{\theta\in\Theta}(F^{-1}(z_i;\theta))^2  + (\gamma_k^q)^2 \\
    & \leq \max\{\mathbb{E}[q_{k,i}^2], (\varepsilon_f^N)^{-1}C_f\sup_{\theta\in\Theta}(F^{-1}(z_i;\theta))^2\}+ (\gamma_k^q)^2\\
 & \leq \max\{\max_{0\leq k\leq K_{1}}\mathbb{E}[q_{k,i}^2], (\varepsilon_f^N)^{-1}C_f\sup_{\theta\in\Theta}(F^{-1}(z_i;\theta))^2\}+ \sum_{k=0}^\infty (\gamma_k^q)^2,
\end{align*}
which implies $\sup_k\mathbb{E}[q_{k+1,i}^2]<\infty$ since $\sup_{\theta\in\Theta}(F^{-1}(z_i;\theta))^2$ and $\sum_{k=0}^\infty (\gamma_k^q)^2$ are finite by the compactness of $\Theta$, Assumption \ref{aspt:diff_quantile}, and \ref{aspt:step_size}(b). 
\endproof

Next, we present the following lemma on kernel estimation, which will be used later to show that, under appropriate conditions, the updates in (\ref{eq:iter_d}) remain bounded. For readability, we abbreviate $G_1(X_{k};\theta_k,q_{k,i})$ and $G_3(Y_{k};\theta_k, q_{k,i},h_k)$ as $G_{1,k,i}$ and $G_{3,k,i}$, respectively, in the remainder of the appendices.

\begin{lemma}\label{lem:kde}
If Assumption \ref{aspt:density} holds, then $$\mathbb{E}[G_{3,k,i}|\mathcal{F}_k]-f(q_{k,i};\theta_k) = O(h_k^2) \text{ and } \mathbb{E}[G_{3,k,i}^2|\mathcal{F}_k]=O(h_k^{-1}).$$
\end{lemma}
\proof 
By the definition of $\mathbb{E}[G_{3,k,i}|\mathcal{F}_k]$, we have
\begin{align}
     \mathbb{E}[G_{3,k,i}|\mathcal{F}_k] \nonumber
     &= \int_{\mathbb{R}}  \mathcal{K}(z) f(zh_k+q_{k,i};\theta_k) dz \\
    & = \int_{\mathbb{R}}  \mathcal{K}(z) \big(f(q_{k,i};\theta_k) + zh_k f'(q_{k,i};\theta_k) + \frac{z^2 h_k^2}{2}f''(\xi_z;\theta_k)\big) dz \nonumber\\
    &= f(q_{k,i};\theta_k) + \frac{ h_k^2}{2} \int_{\mathbb{R}}  z^2 \mathcal{K}(z) f''(\xi_z;\theta_k) dz = f(q_{k,i};\theta_k) + O(h_k^2), \label{eq:G2_mean}
\end{align}
where $\xi_z\in (q_{k,i}, zh_k+ q_{k,i})$, 
the fifth equality comes from the definition of our kernel function, and the last one uses Assumption \ref{aspt:density}(b) and the property of the kernel function. The order of the conditional second moment $\mathbb{E}[G_{3,k,i}^2|\mathcal{F}_k]$ can be derived similarly as below
\begin{align}
    \mathbb{E}[G_{3,k,i}^2|\mathcal{F}_k]   
    &= \int_{\mathbb{R}}  \frac{1}{h_k^2} \mathcal{K}^2(\frac{y-q_{k,i}}{h_k}) f(y;\theta_k) dy \nonumber\\
    &= \frac{1}{h_k} \int_{\mathbb{R}}  \mathcal{K}^2(z) f(zh_k+q_{k,i};\theta_k) dz 
    = O(h_k^{-1}), \label{eq:G2_second_moment}
\end{align}
where the last equality is obtained by Assumption \ref{aspt:density}(a)  and the square integrability of the kernel function. 
\endproof

\begin{theorem}\label{th:d_stability}
If Assumptions \ref{aspt:step_size}(a), (e), \ref{aspt:dist_grad} and \ref{aspt:density} hold, then the sequence $\{D_k\}$ generated by recursion (\ref{eq:iter_d}) is square-integrable and bounded w.p.1, i.e., $\sup_{k,i} \mathbb{E} [\Vert D_{k,i}\Vert^2] < \infty$ and $\sup_{k,i}\Vert D_{k,i}\Vert <\infty$ w.p.1.
\end{theorem}
\proof 
The proof is structured in two main steps.

\textbf{Justification of square-integrability.} The recursion (\ref{eq:iter_d}) can be transformed into $D_{k+1,i}=(1-\gamma^D_k G_{3,k,i})D_{k,i} + \gamma^D_k G_{1,k,i}$. Taking square on both sides, we have $\Vert D_{k+1,i}\Vert ^2\leq \big(|1-\gamma^D_k G_{3,k,i}| \Vert D_{k,i}\Vert + \gamma^D_k C_{g}\big)^2$, where the second term comes on the right-hand side from Assumption \ref{aspt:dist_grad}.
Next, taking conditional expectations on both sides and using the Cauchy–Schwarz inequality, we can obtain
\begin{align}
    \mathbb{E} [\Vert D_{k+1,i}\Vert^2|\mathcal{F}_k]  
    \leq (\sqrt{\mathbb{E} [(1-\gamma^D_k G_{3,k,i})^2|\mathcal{F}_k]} \Vert D_{k,i}\Vert + \gamma^D_k C_{g})^2.\label{eq:d_conditional_bound}
\end{align}
With Lemma \ref{lem:kde}, we can obtain 
\begin{align}
\mathbb{E} [(1-\gamma^D_k G_{3,k,i})^2|\mathcal{F}_k] & = 1 - 2\gamma^D_k \mathbb{E}[G_{3,k,i}|\mathcal{F}_k] + (\gamma^D_k)^2 \mathbb{E}[G_{3,k,i}^2|\mathcal{F}_k]
\nonumber \\
& = 1 - 2\gamma^D_k f(q_{k,i};\theta_k) + 2\gamma^D_k O(h_k^2) + (\gamma^D_k)^2 O(h_k^{-1}). 
\nonumber
\end{align}
By Assumption \ref{aspt:step_size}(a), we have $\gamma^D_k\rightarrow 0$ as $k\rightarrow 0$.
Then with Assumptions \ref{aspt:step_size}(e) and \ref{aspt:density}, there exist constants $C_1 > 0$ and $K_{2} > 0$ such that for $k\geq K_{2}$, 
$\mathbb{E} [(1-\gamma^D_k G_{3,k,i})^2|\mathcal{F}_k] 
\leq 1- 2 \gamma^D_k \varepsilon_f^N + 2\gamma^D_k C_1 ( h_k^2 + \gamma^D_k  h_k^{-1}) \leq 1-\gamma^D_k \varepsilon_f^N$, w.p.1.
Thus, by combining this bound with the inequality (\ref{eq:d_conditional_bound}), we have w.p.1 that for all $k>K_{2}$, $\mathbb{E} [\Vert D_{k+1,i}\Vert^2|\mathcal{F}_k] \leq (\sqrt{1-\gamma^D_k \varepsilon_f^N} \Vert D_{k,i}\Vert + \gamma^D_k C_{g})^2$.
Taking expectations at both sides and applying the Cauchy–Schwarz inequality, we can obtain that for all $k>K_{2}$,
\begin{align*}
    \mathbb{E} [\Vert D_{k+1,i}\Vert^2] &\leq (\sqrt{1-\gamma^D_k \varepsilon_f^N} \sqrt{\mathbb{E} [ \Vert D_{k,i}\Vert^2]} + \gamma^D_k C_{g})^2
    \\&\leq \max_{x\in(0,1)}( \sqrt{\mathbb{E} [ \Vert D_{k,i}\Vert^2]} \sqrt{1-x} + C_{g}(\varepsilon_{f}^N)^{-1} x)^2 \\
    &\leq \max\{\mathbb{E} [ \Vert D_{k,i}\Vert^2], 4 C_{g}^2 (\varepsilon_{f}^N)^{-2}\} 
    \leq \max\{\max_{0\leq k\leq K_{2}}\mathbb{E} [ \Vert D_{k,i}\Vert^2], 4 C_{g}^2 (\varepsilon_{f}^N)^{-2}\},
\end{align*}
which implies $\sup_k \mathbb{E} [\Vert D_{k,i}\Vert^2] < \infty$ since $K_{2}$ is finite. 

\textbf{Justification of pathwise boundedness.}
Next, we rewrite the recursion (\ref{eq:iter_d}) as 
\begin{align}
D_{k+1,i} &= (1-\gamma^D_k \mathbb{E}[G_{3,k,i}|\mathcal{F}_k]) D_{k,i} + \gamma^D_k \mathbb{E}[G_{1,k,i}|\mathcal{F}_k] + \gamma^D_k (V_{1,k,i} + V_{2,k,i}), \label{eq:iter_d_martingale}
\end{align}
where $V_{1,k,i} = G_{1,k,i} - \mathbb{E}[G_{1,k,i}|\mathcal{F}_k]$, $V_{2,k,i} =( \mathbb{E}[G_{3,k,i}|\mathcal{F}_k] - G_{3,k,i}) D_{k,i}$. 
By Assumptions \ref{aspt:step_size}(e), \ref{aspt:density}(a) and Lemma \ref{lem:kde}, there exists a constant $K_{3}>0$ such that  for all $k\geq K_{3}$, $\mathbb{E}[G_{3,k,i}|\mathcal{F}_k] \in[\frac{\varepsilon_f^N}{2},C_f+\frac{\varepsilon_f^N}{2}]$. Then, with Assumption \ref{aspt:step_size}(a), we have another $K_{4}\geq K_{3}$ such that $\gamma^D_k \mathbb{E}[G_{3,k,i}|\mathcal{F}_k]\in(0,1)$ for all $k\geq K_4$.
Therefore, we can expand the equality (\ref{eq:iter_d_martingale}) and take norms on both sides, i.e.,
\begin{align}
    \Vert D_{k+1,i}\Vert &\leq \Vert D_{K_{4},i}\Vert \prod_{r=K_{4}}^k (1-\gamma^D_r \mathbb{E}[G_{3,r,i}|\mathcal{F}_r]) + C_{g}\sum_{r=K_{4}}^k \gamma^D_r  \prod_{s=r+1}^k (1-\gamma^D_s \mathbb{E}[G_{3,s,i}|\mathcal{F}_s]) \nonumber  \\
    &\quad + \sum_{t=1}^2 \big\Vert \sum_{r=K_{4}}^k \gamma^D_r V_{t,r,i} \prod_{s=r+1}^k (1-\gamma^D_s \mathbb{E}[G_{3,s,i}|\mathcal{F}_s])  \big\Vert.\label{eq:iter_d_norm}
\end{align}
With Assumption \ref{aspt:step_size}(a), we can obtain 
\begin{align*}
    \prod_{r=K_{4}}^k (1-\gamma^D_r \mathbb{E}[G_{3,r,i}|\mathcal{F}_r])\leq  \exp\big\{-\sum_{r=K_{4}}^k \gamma^D_r \mathbb{E}[G_{3,r,i}|\mathcal{F}_r]\big\}\leq  \exp\big\{-\frac{\varepsilon_f^N}{2} \sum_{r=K_{4}}^k \gamma^D_r \big\}\rightarrow 0
\end{align*}
as $k\rightarrow \infty$. For the second term on the right-hand side of (\ref{eq:iter_d_norm}), by noticing that for all $k\geq K_{4}$, $\frac{\gamma^D_{k} \varepsilon_f^N}{2}\in(0,1) $, we have w.p.1, 
\begin{align*}
    \sum_{r=K_{4}}^k &\gamma^D_r \prod_{s=r+1}^k (1-\gamma^D_s \mathbb{E}[G_{3,s,i}|\mathcal{F}_s]) 
\leq  \frac{2}{\varepsilon_f^N}  \sum_{r=K_{4}}^k \frac{\gamma^D_r \varepsilon_f^N}{2} \prod_{s=r+1}^k (1-\frac{\gamma^D_s \varepsilon_f^N}{2})\\
&\leq 
\frac{2}{\varepsilon_f^N} \sum_{r=K_{4}+1}^k \frac{\gamma^D_r \varepsilon_f^N}{2} \prod_{s=r+1}^k (1-\frac{\gamma^D_s \varepsilon_f^N}{2})\leq \cdots\leq \frac{2}{\varepsilon_f^N}  \frac{\gamma^D_{k} \varepsilon_f^N }{2} \leq  \gamma^D_{K_{4}}.
\end{align*}
Then, note that $\sum_{r=K_{4}}^{k} \gamma^D_{r}  V_{1,r,i}$ is a martingale. We have 
\begin{align*}
    \mathbb{E}\big[\big\Vert\sum_{r=K_{4}}^k \gamma^D_{r}  V_{1,r,i} \big\Vert^2 \big] 
  &= \sum_{r=K_{4}}^k (\gamma^D_{r})^2 \mathbb{E}\big[ \Vert G_{1,r,i} - \mathbb{E}[G_{1,r,i}|\mathcal{F}_r] \Vert^2 \big] \\
  & \leq \sum_{r=K_{4}}^k (\gamma^D_{r})^2 \mathbb{E}\big[ \Vert G_{1,r,i} \Vert^2 \big] \leq \sum_{r=K_{4}}^k (\gamma^D_{r})^2 C_{g}^2 <\infty,
\end{align*}
where the cross terms in the equality are eliminated since $\mathbb{E}[V_{1,r,i}V_{1,s,i}]= \mathbb{E}[V_{1,r,i}\mathbb{E}[V_{1,s,i}|\mathcal{F}_r]]= 0$ for $r<s$, and the second and last inequalities come from Assumption \ref{aspt:dist_grad} and \ref{aspt:step_size}(a), respectively. Thus, $\sum_{r=K_{4}}^{k} \gamma^D_{r}  V_{1,r,i}$ is $L^2$-bounded and converges to a finite random vector \citep{durrett2019probability}. Let $\tilde{\gamma}_r = \prod_{s=K_{4}}^r (1-\gamma^D_s \mathbb{E}[G_{3,s,i}|\mathcal{F}_s])^{-1}$. Since for all $r\geq K_{4}$, $\gamma^D_s \mathbb{E}[G_{3,s,i}|\mathcal{F}_s]\in(0,1)$, $\{\tilde{\gamma}_r\}$ is a positive non-decreasing sequence and $\lim_{r\to\infty}\tilde{\gamma}_r = \infty$. Then the third term on the right-hand side of (\ref{eq:iter_d_norm}) can be written as 
$\sum_{r=K_{4}}^k \gamma^D_r V_{1,r,i} \prod_{s=r+1}^k (1-\gamma^D_s \mathbb{E}[G_{3,s,i}|\mathcal{F}_s]) =  (\tilde{\gamma}_k)^{-1} \sum_{r=K_{4}}^k \tilde{\gamma}_r \gamma^D_r V_{1,r,i}$, which converges to zero as $k\rightarrow\infty$ by Kronecker's lemma \citep{shiryaev2016probability}.
For the last term in (\ref{eq:iter_d_norm}), we have
\begin{align*}
  \mathbb{E}\big[\big\Vert\sum_{r=K_{4}}^k& \gamma^D_{r}  V_{2,r,i} \big\Vert^2 \big] 
  = \sum_{r=K_{4}}^k  (\gamma^D_{r})^2  \mathbb{E} [(\mathbb{E}[G_{3,r,i}|\mathcal{F}_r] - G_{3,r,i})^2 \Vert D_{r,i}\Vert^2] \\
  &\leq  \sum_{r=K_{4}}^k  (\gamma^D_{r})^2  
  \mathbb{E} [\mathbb{E}[G_{3,r,i}^2|\mathcal{F}_r] \Vert D_{r,i}\Vert ^2] 
  \leq C_2 \sup_k \mathbb{E} [ \Vert D_{k,i}\Vert ^2]  \sum_{r=K_{4}}^k (\gamma^D_{r})^2 h_r^{-1}  < \infty,
\end{align*}
where $C_2>0$ is a constant, the second inequality uses Lemma \ref{lem:kde} and Assumption \ref{aspt:density}(a), and the last comes from Assumptions \ref{aspt:step_size}(a), (e) and the fact that $\sup_k \mathbb{E} [ \Vert D_{k,i}\Vert ^2]<\infty$. Since $\sum_{r=K_{4}}^{k} \gamma^D_{r} V_{2,r,i}$ is an $L^2$-bounded martingale, the same argument as that for the third term in (\ref{eq:iter_d_norm}) can be used to show the last term also converges to zero as $k\rightarrow \infty$. Hence, we have the conclusion that $\sup_k \Vert D_{k,i}\Vert = \max\{\max_{0\leq k\leq K_{4}}\Vert D_{k,i}\Vert , \sup_{k\geq K_{4}} \Vert D_{k,i}\Vert\}<\infty$, w.p.1. The results can be extended to uniform ones with respect to $i$, given a fixed $N$, i.e., $\sup_{k,i} \mathbb{E} [\Vert D_{k,i}\Vert^2] < \infty$ and $\sup_{k,i}\Vert D_{k,i}\Vert <\infty$ w.p.1.
\endproof

\subsection{Proof of Theorem~\ref{th:main_convergence}}\label{sec:Proof of Strong Convergence}

Now we have the following main convergence result for our three-timescale algorithm, whose proof is based on iteratively verifying the conditions given in Theorem \ref{th:single_convergence}. If the concavity of $\mathcal{J}_N(\theta)$ on $\Theta$ is further provided, then the conclusion can be strengthened to Corollary \ref{cor:main_convergence}. 
The proof is organized into the following three parts, corresponding to different timescales.

\textbf{Analysis under the step-size $\gamma^{q}_k$.} Let $m^q_{k,i}=F(q_{k,i};\theta_k)-\mathbf{1}\{Y_{k}\leq q_{k,i}\}$. Rewrite the last two recursions as below: 
\begin{equation}
\begin{aligned}
    q_{k+1,i} &= q_{k,i}+\gamma^{q}_{k}(z_i - F(q_{k,i};\theta_k) + m^q_{k,i}), &\text{for}\ i=1,\cdots,N, \\
    \theta_{k+1} &= \theta_k + \gamma^{q}_{k}\frac{\gamma^{\theta}_{k}}{\gamma^{q}_{k}} \sum_{i=1}^N -D_{k,i}(\tilde{w}(z_i) - \tilde{w}(z_{i-1}) ) + \gamma^{q}_{k}\frac{\gamma^{\theta}_{k}}{\gamma^{q}_{k}} P_k, \label{eq:iter_unified_scale_q}
\end{aligned}
\end{equation}
where $P_k\in-C(\theta_{k+1})$ is the projection term.
We expect recursion (\ref{eq:iter_unified_scale_q}) to track the coupled ODEs 
$\dot{q}_i(t) = z_i - F(q_i(t);\theta(t))$ and $\dot{\theta}(t) = 0$, which is equivalent to the ODE
$\dot{q}_i(t) = z_i - F(q_i(t);\bar{\theta})$ 
and $\theta(t)\equiv \bar{\theta}$ given some fixed $\bar{\theta}\in\Theta$. The noise $\{m^q_{k,i}\}$ is a square-integrable martingale difference sequence by the definition of CDF, thereby $\sum_{k=0}^{\infty} (\gamma^{q}_k)^2 \mathbb{E}[\vert m^q_{k,i} \vert^2]<\infty$ by Assumption \ref{aspt:step_size}(b).
Since $P_k\in-C(\theta_{k+1})$ and $\theta_k\in\Theta$, we have $0\leq P_k^{\top}(\theta_k-\theta_{k+1})=\gamma^{\theta}_{k}P_k^{\top}(\sum_{i=1}^N -D_{k,i}(\tilde{w}(z_i) - \tilde{w}(z_{i-1}) )-P_k)\leq \gamma^{\theta}_{k}(\Vert P_k\Vert \sup_i \Vert D_{k,i}\Vert -
\Vert P_k\Vert^2)$, which implies $\sup_k \Vert P_k\Vert\leq \sup_{k,i} \Vert D_{k,i}\Vert<\infty$ w.p.1. 
With Assumption \ref{aspt:step_size}(d) and the boundedness of $D_{k,i}$ and $P_k$,  $\frac{\gamma^{\theta}_{k}}{\gamma^{q}_{k}} \sum_{i=1}^N -D_{k,i}(\tilde{w}(z_i) - \tilde{w}(z_{i-1}) ) + \frac{\gamma^{\theta}_{k}}{\gamma^{q}_{k}} P_k$ is asymptotically negligible w.p.1.
The remaining conditions in Theorem \ref{th:single_convergence} can be verified by Assumption \ref{aspt:step_size}(b) in combination with Lemma 1 in \cite{jiang2023quantile}. Thus, the outputs of recursions (\ref{eq:iter_unified_scale_q}) a.s. converge to some limit set $\{ \{F(z_i;\bar{\theta})\}, \bar{\theta}: \bar{\theta}\in\Theta  \}$.

\textbf{Analysis under the step-size $\gamma^D_k$.}
Let $m^{D}_{k,i} = V_{1,k,i} + V_{2,k,i}$, and $b_{k,i}=(f(q_{k,i};\theta_k)-\mathbb{E}[G_{3,k,i}|\mathcal{F}_k])D_{k,i}$. Rewrite the three recursions into a unified timescale $\gamma^D_k$ as below:
\begin{equation}
\begin{aligned}
    D_{k+1,i} &= D_{k,i}+\gamma^{D}_{k}(-\nabla_{\theta}F(q_{k,i};\theta_k) - f(q_{k,i};\theta_k)D_{k,i} + m^D_{k,i} + b_{k,i}), &\text{for}\ i=1,\cdots,N,  \\
    q_{k+1,i} &= q_{k,i}+\gamma^{D}_{k}\frac{\gamma^{q}_{k}}{\gamma^{D}_{k}}(z_i - F(q_{k,i};\theta_k) + m^q_{k,i}), &\text{for}\ i=1,\cdots,N, \\
    \theta_{k+1} &= \theta_k + \gamma^{D}_{k}\frac{\gamma^{\theta}_{k}}{\gamma^{D}_{k}} \sum_{i=1}^N -D_{k,i}(\tilde{w}(z_i) - \tilde{w}(z_{i-1}) ) + \gamma^{D}_{k}\frac{\gamma^{\theta}_{k}}{\gamma^{D}_{k}} P_k, \label{eq:iter_unified_scale_D}
\end{aligned}
\end{equation}
where the corresponding coupled ODEs are $\dot{D}_i(t)=-\nabla_{\theta}F(q_i(t);\theta(t))-f(q_i(t);\theta(t))D_i(t)$, $\dot{q}_i(t) = 0$ and $\dot{\theta}(t)=0$. With the conclusion for recursions (\ref{eq:iter_unified_scale_q}), they are equivalent to the linear ODE (\ref{eq:ode_d}), $q_i(t)\equiv F^{-1}(z_i;\bar{\theta})$, and $\theta(t)\equiv\bar{\theta}$ given some $\bar{\theta}\in\Theta$. 
With Lemma \ref{lem:q_stability} and Theorem \ref{th:d_stability}, we have the boundedness of sequences generated by three recursions, which also implies $-\nabla_{\theta}F(q_{k,i};\theta_k) - f(q_{k,i};\theta_k)D_{k,i}$ remains bounded on almost every sample path. 
The noise term $\{\frac{\gamma^{q}_{k}}{\gamma^{D}_{k}}m^q_{k,i}\}$ is a square-integrable martingale difference sequence, since $\mathbb{E}[(\frac{\gamma^{q}_{k}}{\gamma^{D}_{k}}m^q_{k,i})^2]\leq  \big(\frac{2\gamma^{q}_{0}}{\gamma^{D}_{0}}\big)^2$ by Assumption \ref{aspt:step_size}(d). And there exists a constant $C_3>0$ such that
\begin{align*}
    \mathbb{E}[\Vert m^D_{k,i}\Vert^2]&\leq 4C_{g}^2 + \mathbb{E}[( \mathbb{E}[G_{3,k,i}|\mathcal{F}_k] - G_{3,k,i})^2 \Vert D_{k,i}\Vert^2 ]\\
    &\leq 4C_{g}^2 + \mathbb{E}[ \mathbb{E}[\Vert G_{3,k,i}\Vert^2|\mathcal{F}_k] \Vert D_{k,i}\Vert^2]  \leq 4C_{g}^2 + C_3 h_k^{-1} \mathbb{E}[\Vert D_{k,i}\Vert^2],
\end{align*}
where the last inequality comes from Lemma \ref{lem:kde}.
Therefore, $\sum_{k=0}^{\infty} (\gamma^{D}_k)^2 \mathbb{E}[(\frac{\gamma^{q}_{k}}{\gamma^{D}_{k}}m^q_{k,i})^2]<\infty$ by Assumption \ref{aspt:step_size}(a), and $\sum_{k=0}^{\infty} (\gamma^{D}_k)^2 \mathbb{E}[\Vert m^D_{k,i}\Vert^2]<\infty$ by Assumption \ref{aspt:step_size}(e) and Theorem \ref{th:d_stability}.
Assumption \ref{aspt:step_size}(e), Theorem \ref{th:d_stability}, and Lemma \ref{lem:kde} imply that $b_{k,i}$ shrinks to zero as $k\rightarrow\infty$. 
Analogously to the first part analysis, $\frac{\gamma^{q}_{k}}{\gamma^{D}_{k}}(z_i - F(q_{k,i};\theta_k))$ and $\frac{\gamma^{\theta}_{k}}{\gamma^{D}_{k}} \sum_{i=1}^N -D_{k,i}(\tilde{w}(z_i) - \tilde{w}(z_{i-1}) ) + \frac{\gamma^{\theta}_{k}}{\gamma^{D}_{k}} P_k$ are also asymptotically negligible. 
The remaining conditions in Theorem \ref{th:single_convergence} can be verified by Assumption \ref{aspt:step_size}(a) and Lemma \ref{lem:ode_d}.
Therefore, the recursions (\ref{eq:iter_unified_scale_D}) converge to some limit set $\{\{d_i(\bar{\theta}) \}, \{F(z_i;\bar{\theta})\}, \bar{\theta}: \bar{\theta}\in\Theta\}$ a.s.

\textbf{Analysis under the step-size $\gamma^{\theta}_k$.} Finally, we focus on the last recursion (\ref{eq:iter_theta})
and transform it into
\begin{align}
    \theta_{k+1} &= \theta_k + \gamma^{\theta}_{k} \nabla_\theta\mathcal{J}(\theta_k) + \gamma^{\theta}_{k}\sum_{i=1}^N \big(d_{i}(\theta_k)-D_{i,k}\big)(\tilde{w}(z_i) - \tilde{w}(z_{i-1}) ) + \gamma^{\theta}_{k} P_k. \label{eq:iter_theta_proj}
\end{align}
Note that the sequences generated by recursions (\ref{eq:iter_d})-(\ref{eq:iter_theta}) a.s. converge to $\{\{d_i(\bar{\theta})\}, \break \{F(z_i;\bar{\theta})\}, \bar{\theta}: \bar{\theta}\in\Theta \}$. Thus, we have $\Vert \sum_{i=1}^N(D_{i,k} - d_i(\theta_k))(\tilde{w}(z_i)-\tilde{w}(z_{i-1}))\Vert\leq \sup_i\Vert D_{i,k} - d_i(\bar{\theta})\Vert + \sup_i\Vert d_i(\bar{\theta}) - d_i(\theta_k)\Vert$, which a.s. converges to zero as $k\to\infty$.
Note that $d_i(\cdot)$ is bounded on compact $\Theta$ by Assumption \ref{aspt:diff_quantile}, and consequently, $\sup_k \Vert\nabla_\theta\mathcal{J}(\theta_k)\Vert <\infty$ w.p.1.
The remaining conditions in Theorem \ref{th:single_convergence} can be verified by Assumption \ref{aspt:step_size}(c). Thus, we can conclude that the output of the recursion (\ref{eq:iter_theta}) converges to some limit set of projected ODE (\ref{eq:ode_final}) w.p.1, which completes the proof.
\endproof

\begin{corollary}\label{cor:main_convergence}
If Assumptions \ref{aspt:diff_quantile}-\ref{aspt:density} hold and ${
\mathcal{J}}_N(\theta)$ is strictly concave on $\Theta$, then the sequences $\{D_k\}$, $\{q_k\}$ and $\{\theta_k\}$ generated by recursions (\ref{eq:iter_d})-(\ref{eq:iter_theta}) almost surely converge to $\{\{d_i(\theta^*_N)\}$, $ \{F^{-1}(z_i;\theta^*_N)\}, \theta^* _N\}$, where $\theta^*_N=\arg\max_{\theta\in\Theta} \mathcal{J}_N(\theta)$ is the unique optimal solution.
\end{corollary}
\proof
Denote the normal cone to $\Theta$ at $\theta$ as $C(\theta)$. If $\theta^*_N\in\Theta^{\circ}$, then $\nabla_{\theta}\mathcal{J}_N(\theta^*_N)=0$ and $C(\theta^*_N)=\{0\}$;  if $\theta^*_N\in\partial\Theta$, then $\nabla_{\theta}\mathcal{J}_N(\theta^*_N)$ must lie in $C(\theta^*_N)$, so $p(t)=-\nabla_{\theta}\mathcal{J}_N(\theta^*_N)$. By concavity of $\mathcal{J}_N(\theta)$ on $\Theta$, $\theta^*_N$ is the unique equilibrium point. Take $V_2(x)=\Vert x-\theta^*_N\Vert^2$ as the Lyapunov function, and the derivative is $\dot{V}_2(x)=2(x-\theta^*_N)^{\top}(\nabla_{\theta}\mathcal{J}_N(x)+p(t))$. Since $\mathcal{J}_N(\theta)$ is strictly concave, $(\theta^*_N-x)^{\top}\nabla_{\theta} \mathcal{J} (x)>\mathcal{J}_N(\theta^*_N)-\mathcal{J}_N(x)>0$ for any $x\neq\theta^*_N$. Since $p(t)\in-C(x)$, we have $(x-\theta^*_N)^{\top}p(t)\leq0$. Thus, $\theta^*_N$ is global asymptotically stable by the Lyapunov Stability Theory \citep{liapounoff2016probleme}. Thus, based on existing results in Theorems \ref{th:single_convergence} and \ref{th:main_convergence}, we can conclude that recursions (\ref{eq:iter_d})-(\ref{eq:iter_theta}) converge to $\{\{d_i(\theta^*_N)\}, \{F^{-1}(z_i;\theta^*_N)\}, \theta^*_N \}$ w.p.1. 
\endproof

\subsection{Proof of Theorem \ref{th:mse_q}}

For notational convenience, let $\Delta\tilde{w}_i = \tilde{w}(z_i) - \tilde{w}(z_{i-1})<\infty$.
From the first part of the proof of Theorem \ref{th:main_convergence}, we can check that $\Vert P_k\Vert\leq \Vert\sum_{i=1}^N -D_{k,i}\Delta\tilde{w}_i\Vert <\sup_i \Vert D_{k,i}\Vert<\infty$ w.p.1 and $\mathbb{E}[\Vert P_k\Vert^2]\leq\mathbb{E}[\Vert\sum_{i=1}^N -D_{k,i}\Delta\tilde{w}_i\Vert^2]\leq \sup_i \mathbb{E}[\Vert D_{k,i} \Vert^2]<\infty$.
Thus, we have $\Vert \theta_{k+1}-\theta_k\Vert\leq 2\gamma^{\theta}_k \Vert\sum_{i=1}^N -D_{k,i} \Delta\tilde{w}_i\Vert$. 
By Assumption \ref{aspt:diff_quantile},  
$F^{-1}(z_i;\cdot)$ is Lipschitz, and there exist a constant $C_4>0$ such that $\vert F^{-1}(z_i;\theta_k) - F^{-1}(z_i;\theta_{k+1})\vert \leq \gamma^{\theta}_k C_4  \Vert\sum_{i=1}^N -D_{k,i}\Delta\tilde{w}_i\Vert$.
Define $\xi_{k}=q_{k,i} - F^{-1}(z_i;\theta_k)$ and rewrite the recursion (\ref{eq:iter_q}) as 
$$\xi_{k+1}=\xi_k + \gamma^{q}_k(z_i-\mathbf{1}\{Y_{k}\leq q_{k,i}\}) + F^{-1}(z_i;\theta_k) - F^{-1}(z_i;\theta_{k+1}).$$
By taking square on both sides, we have
\begin{align*}
    \xi_{k+1}^2\leq &\ \xi_k^2  + (\gamma^{q}_k)^2 + (\gamma^{\theta}_k)^2 C_4^2  \Vert\sum_{i=1}^N -D_{k,i}\Delta\tilde{w}_i\Vert^2 + 2\gamma^{\theta}_k C_4  \Vert\sum_{i=1}^N -D_{k,i}\Delta\tilde{w}_i\Vert\vert \xi_k \vert  \\
    &\ \quad + 2\xi_k \gamma^{q}_k(z_i-\mathbf{1}\{Y_{k}\leq q_{k,i}\}) + 2\gamma^{q}_k(z_i-\mathbf{1}\{Y_{k}\leq q_{k,i}\})(F^{-1}(z_i;\theta_k) - F^{-1}(z_i;\theta_{k+1})).
\end{align*}
With Assumption \ref{aspt:density}(a) 
and the intermediate value theorem, we have
\begin{align*}
    \mathbb{E}[\xi_{k+1}^2|\mathcal{F}_k]\leq&\ \xi_{k}^2+(\gamma^{q}_k)^2+ (\gamma^{\theta}_k)^2 C_4^2  \Vert\sum_{i=1}^N D_{k,i}\Delta\tilde{w}_i\Vert^2 + 2\vert \xi_k\vert \gamma^{\theta}_k  C_4  \Vert\sum_{i=1}^N D_{k,i}\Delta\tilde{w}_i\Vert
    \\
    &\ \quad -2\gamma^{q}_k\xi_k^2 f(\hat{q};\theta_k)-2\gamma^{q}_k\xi_k f(\hat{q};\theta_k) (F^{-1}(z_i;\theta_k) - F^{-1}(z_i;\theta_{k+1}))\\
    \leq&\ \xi_{k}^2(1-2\gamma^{q}_k {{\varepsilon}_f^N}) + (\gamma^{q}_k)^2+2(\gamma^{\theta}_k)^2 C_4^2  \sup_i \Vert\sum_{i=1}^N D_{k,i}\Delta\tilde{w}_i\Vert^2 
    \nonumber \\
    &\  \quad+ 2\vert \xi_k\vert \gamma^{\theta}_k C_4  \Vert\sum_{i=1}^N D_{k,i}\Delta\tilde{w}_i\Vert(1+\gamma^{q}_k C_f),
\end{align*}
where $\theta_{k+1}$ is $\mathcal{F}_k$-measurable, and $\hat{q}$ lies in the interval between $q_{k,i}$ and $F^{-1}(z_i;\theta_k)$. 
We denote $\tilde{C}_4=C_4 \sup_{k,i} \sqrt{\mathbb{E}[\Vert D_{k,i}\Vert^2]}$, and by Theorem \ref{th:d_stability}, we have $\tilde{C}_4<\infty$. By taking expectation with respect to $\mathcal{F}_k$ and applying the Cauchy-Schwarz inequality, we obtain
\begin{align*}
    \mathbb{E}[\xi_{k+1}^2]  &= \mathbb{E}[\mathbb{E}[\xi_{k+1}^2|\mathcal{F}_k]]
    \leq \mathbb{E}[\xi_k^2](1-2\gamma^{q}_k {{\varepsilon}_f^N}) + (\gamma^{q}_k)^2 
    + (\gamma^{\theta}_k)^2 \tilde{C}_4^2 + 2 \mathbb{E}[\vert \xi_k\vert]\gamma^{\theta}_k \tilde{C}_4(1+\gamma^{q}_k C_f)\\
    & \leq \mathbb{E}[\xi_k^2](1-2\gamma^{q}_k {{\varepsilon}_f^N}) + (\gamma^{q}_k)^2 + (\gamma^{\theta}_k)^2 \tilde{C}_4^2 
    + 2  \gamma^{\theta}_k \tilde{C}_4 (1+\gamma^{q}_k C_f)\big(\mathbb{E}[\xi_k^2]\big)^{\frac{1}{2}}\\
    & \leq \mathbb{E}[\xi_k^2](1-2\gamma^{q}_k {{\varepsilon}_f^N}) + (\gamma^{q}_k)^2 + (\gamma^{\theta}_k)^2 \tilde{C}_4^2 
    + \frac{( \gamma^{\theta}_k \tilde{C}_4(1+\gamma^{q}_k C_f) )^2}{\gamma^{q}_k {{\varepsilon}_f^N}}+\gamma^{q}_k {{\varepsilon}_f^N}\mathbb{E}[\xi_k^2]\\
    & = \mathbb{E}[\xi_k^2](1-\gamma^{q}_k {{\varepsilon}_f^N}) + (\gamma^{q}_k)^2 + \frac{(\gamma^{\theta}_k)^2}{\gamma^{q}_k}\tilde{C}_4^2\big(\gamma^{q}_k+\frac{(1+(\gamma^{q}_k)C_f)^2}{{{\varepsilon}_f^N}}\big)\\
    & \leq \mathbb{E}[\xi_k^2](1-\gamma^{q}_k {{\varepsilon}_f^N}) + (\gamma^{q}_k)^2 + \frac{(\gamma^{\theta}_k)^2}{\gamma^{q}_k} C_5,
\end{align*}
where $C_5 = \tilde{C}_4^2\big(\gamma^{q}_1+\frac{(1+\gamma^{q}_1 C_f)^2}{{{\varepsilon}_f^N}}\big)$. By assuming $\gamma^{q}_k{{\varepsilon}_f^N}<1$ without loss of generality and repeatedly applying the inequality above, we have
\begin{align}
    \mathbb{E}[\xi_{k+1}^2] &\leq \prod_{i=1}^k (1-\gamma^{q}_i {{\varepsilon}_f^N})\mathbb{E}[\xi_1^2] \nonumber \\
    & \qquad \qquad
    +\sum_{i=1}^{k-1}\bigg[\prod_{j=i+1}^k (1-\gamma^{q}_j {{\varepsilon}_f^N}) \bigg]\gamma^{q}_i\big(\gamma^{q}_i+\frac{(\gamma^{\theta}_i)^2}{(\gamma^{q}_i)^2}C_5\big) + \big((\gamma^{q}_{k+1})^2
     +\frac{(\gamma^{\theta}_{k+1})^2}{\gamma^{q}_{k+1}}C_5\big)\nonumber\\
    &= \prod_{i=1}^k (1-\gamma^{q}_i {{\varepsilon}_f^N})\big(\mathbb{E}[\xi_1^2]+\frac{(\gamma^{q}_{1})^2+(\gamma^{\theta}_{1})^2(\gamma^{q}_{1})^{-1}C_5}{1-\gamma^{q}_1 {{\varepsilon}_f^N}}\big) 
    \nonumber \\
    & \qquad \qquad +\sum_{i=2}^{k-1}\bigg[\prod_{j=i+1}^k (1-\gamma^{q}_j{{\varepsilon}_f^N}) \bigg]\gamma^{q}_i\big(\gamma^{q}_i+\frac{(\gamma^{\theta}_i)^2}{(\gamma^{q}_i)^2}C_5\big)
    + \big((\gamma^{q}_{k})^2+\frac{(\gamma^{\theta}_{k})^2}{\gamma^{q}_{k}}C_5\big). \label{order1}
\end{align}
Since $\sum_{i=2}^{k-1} \left[\prod_{j=i+1}^k (1-cj^{-\beta}) \right] i^{-\beta}i^{-p}= O(k^{-p})$ for $c, p>0$ \cite[see, e.g.,][]{jiang2023quantile, hu2025quantile}, the second term on the right-hand side of inequality (\ref{order1}) is in the order of $O(k^{-\beta})+O(k^{2\beta-2\gamma})=O(\gamma^{q}_k)+O((\gamma^{\theta}_k)^2(\gamma^{q}_k)^{-2})$. Note that
$\mathbb{E}[\xi_1^2]\leq\mathbb{E}[\xi_0^2](1-\gamma^{q}_0 {{\varepsilon}_f^N})+(\gamma^{q}_0)^2+(\gamma^{\theta}_0)^2(\gamma^{q}_0)^{-1}\tilde{C}_4^2\big(\gamma^{q}_0+(1+\gamma^{q}_0C_f)^2({{\varepsilon}_f^N})^{-1}\big)$ and $\prod_{i=0}^k (1-\gamma^{q}_i{{\varepsilon}_f^N})\leq \exp\{-{{\varepsilon}_f^N}\sum_{i=0}^k\gamma^{q}_i\}\leq \exp\{-{{\varepsilon}_f^N}k^{1-\beta}\}$,
so the first term on the right-hand side of inequality (\ref{order1}) decays exponentially so that it can be absorbed into the second term.
Finally, we can conclude that $\mathbb{E}[\vert q_{k,i} - F^{-1}(z_i;\theta_k)\vert^2] = O(\gamma^{q}_k)+O((\gamma^{\theta}_k)^2(\gamma^{q}_k)^{-2})$.
\endproof

\subsection{Proof of Theorem~\ref{th:mse_d}} 

Let $\eta_k = D_{k,i} - \nabla_\theta F^{-1}(z_i;\theta_k)$. The recursion (\ref{eq:iter_d}) can be transformed into
\begin{align*}
    \eta_{k+1} = \eta_k +\gamma^{D}_k (G_{1,k,i} - G_{3,k,i} D_{k,i}) + \nabla_\theta F^{-1}(z_i;\theta_k) - \nabla_\theta F^{-1}(z_i;\theta_{k+1}).
\end{align*}
By taking square on both sides, we have
\begin{align*}
    \Vert \eta_{k+1}\Vert^2 \leq\ \Vert \eta_{k}\Vert^2 &+ (\gamma^{D}_k)^2  \Vert G_{1,k,i} - G_{3,k,i} D_{k,i}\Vert^2 + \Vert \nabla_\theta F^{-1}(z_i;\theta_k) - \nabla_\theta F^{-1}(z_i;\theta_{k+1}) \Vert^2\\
    &+ 2\gamma^{D}_k (\nabla_\theta F^{-1}(z_i;\theta_k) - \nabla_\theta F^{-1}(z_i;\theta_{k+1})+\eta_k)^{\top} (G_{1,k,i} - G_{3,k,i} D_{k,i})
    \\
    & + 2 (\nabla_\theta F^{-1}(z_i;\theta_k) - \nabla_\theta F^{-1}(z_i;\theta_{k+1}))^{\top} \eta_k .
\end{align*}
Since $F^{-1}(z_i;\theta)$ is twice differentiable by Assumption \ref{aspt:hessian}, $\nabla_{\theta} F^{-1}(z_i;\theta)$ is Lipschitz continuous in $\theta$, i.e., there exists a constant $C_6>0$ such that
$\Vert \nabla_\theta F^{-1}(z_i;\theta_k) - \nabla_\theta F^{-1}(z_i;\theta_{k+1}) \Vert \leq C_6 \Vert \theta_{k+1}-\theta_k \Vert \leq  2C_6 \gamma^{\theta}_k \Vert\sum_{i=1}^N -D_{k,i}\Delta\tilde{w}_i\Vert$.
And by Assumption \ref{aspt:dist_grad} and Lemma \ref{lem:kde}, we have
$\mathbb{E}[\Vert G_{1,k,i} - G_{3,k,i} D_{k,i}\Vert^2\vert \mathcal{F}_k] \leq C_g^2 +  \mathbb{E}[G_{3,k,i}^2\vert \mathcal{F}_k]   \Vert  D_{k,i}\Vert^2 \leq C_g^2 + C_3 h_k^{-1} \Vert  D_{k,i}\Vert^2 $. Let $\delta_k = \nabla_{\theta}F(F^{-1}(z_i;\bar{\theta});\theta_k)\vert_{\bar{\theta}=\theta_k}-f(F^{-1}(z_i;\theta_k);\theta_k)D_{k,i} -\nabla_{\theta}F(q_{k,i};\theta_k) + f(q_{k,i};\theta_k)D_{k,i}$. Therefore, with Assumption \ref{aspt:density}(a), we take conditional expectations on both sides and obtain
\begin{align*}
    \mathbb{E}[\Vert \eta_{k+1}&\Vert^2 \vert \mathcal{F}_k] \\
    \leq\Vert \eta_{k}\Vert^2 &+ (\gamma^{D}_k)^2 (C_g^2 + \frac{C_3}{h_k} \Vert  D_{k,i}\Vert^2) 
    + 4 C_6 \gamma^{\theta}_k \Vert\sum_{i=1}^N D_{k,i}\Delta\tilde{w}_i\Vert \Vert \eta_k\Vert + 4C_6^2(\gamma^{\theta}_k)^2 \Vert\sum_{i=1}^N D_{k,i}\Delta\tilde{w}_i\Vert^2\\
    &+ 4\gamma^{D}_k C_6  \gamma^{\theta}_k \Vert\sum_{i=1}^N D_{k,i}\Delta\tilde{w}_i\Vert\bigg[\big\Vert\nabla_{\theta}F(F^{-1}(z_i;\bar{\theta});\theta_k)\vert_{\bar{\theta}=\theta_k}-f(F^{-1}(z_i;\theta_k);\theta_k)D_{k,i}\big\Vert \\
    &+ \Vert \delta_k \Vert + \big\vert f(q_{k,i};\theta_k) - \mathbb{E}[\Vert G_{3,k,i}\Vert\vert \mathcal{F}_k] \big\vert \cdot \Vert D_{k,i}\Vert\bigg] - 2\gamma^{D}_k \eta_k^{\top}\eta_k f(F^{-1}(z_i;\theta_k);\theta_k)\\
    &+ 2\gamma^{D}_k \Vert \eta_k\Vert \big[ \Vert \delta_k \Vert 
    + \vert f(q_{k,i};\theta_k) - \mathbb{E}[\Vert G_{3,k,i}\Vert\vert \mathcal{F}_k] \vert \cdot \Vert D_{k,i}\Vert\big]\\
    \leq\Vert \eta_{k}\Vert^2 &+ (\gamma^{D}_k)^2 (C_g^2 + \frac{C_3}{h_k}\Vert  D_{k,i}\Vert^2) + 4 C_6 \gamma^{\theta}_k \Vert\sum_{i=1}^ND_{k,i}\Delta\tilde{w}_i\Vert\Vert \eta_k\Vert+ 4C_6^2(\gamma^{\theta}_k)^2 \Vert\sum_{i=1}^N D_{k,i}\Delta\tilde{w}_i\Vert^2\\
    &+ 4\gamma^{D}_k C_6 \gamma^{\theta}_k \Vert\sum_{i=1}^N D_{k,i}\Delta\tilde{w}_i\Vert \big(C_f \Vert\eta_k\Vert  +  (L_d+L_f\Vert D_{k,i}\Vert) \vert \xi_k \vert+ C_7 h_k^2 \Vert D_{k,i}\Vert\big) \\
    &+ 2\gamma^{D}_k \Vert \eta_k\Vert \big(-{{\varepsilon}_f^N} \Vert\eta_k\Vert  +  (L_d+L_f\Vert D_{k,i}\Vert) \vert \xi_k \vert+ C_7 h_k^2 \Vert D_{k,i}\Vert\big),
\end{align*}
where the second inequality can be justified by Assumptions \ref{aspt:density}(a) 
and \ref{aspt:lip}, and $C_7>0$ such that $\vert f(q_{k,i};\theta_k) - \mathbb{E}[\Vert G_{3,k,i}\Vert\vert \mathcal{F}_k] \vert \leq C_7 h_k^2$ by Lemma \ref{lem:kde}. By Theorem \ref{th:d_stability}, there exist $C_8, C_9, C_{10}>0$ such that $L_d+L_f \sqrt{\mathbb{E}[\Vert  D_{k,i}\Vert^2]}\leq L_d+L_f \sup_{k,i} \sqrt{\mathbb{E}[\Vert  D_{k,i}\Vert^2]}< C_8$, $ C_7 h_k^2 \Vert D_{k,i}\Vert\leq  C_7 h_k^2 \sup_{k,i} \sqrt{\mathbb{E}[\Vert  D_{k,i}\Vert^2]} < C_9 h_k^2$, and $C_g^2 + C_3 h_k^{-1} \Vert  D_{k,i}\Vert^2\leq C_g^2 + C_3 h_k^{-1} \sup_{k,i} \mathbb{E}[\Vert  D_{k,i}\Vert^2]< C_{10} h_k^{-1}$. We denote $\tilde{C}_6=2C_6 \sup_{k,i} \sqrt{\mathbb{E}[\Vert D_{k,i}\Vert^2]}<\infty$. Then, we can further obtain
\begin{align*}
    &\!\!\!\!\!\!\!\!\!\!\!\!\!\!\!\!\!\! \mathbb{E}[\Vert \eta_{k+1}\Vert^2]  \\
    \leq\ (1&-2\gamma^{D}_k{{\varepsilon}_f^N})\mathbb{E}[\Vert\eta_{k}\Vert^2]    + \bigg(2 \tilde{C}_6 \gamma^{\theta}_k (1 + \gamma^{D}_k C_f) + 2\gamma^{D}_k (C_8 \sqrt{\mathbb{E}[\xi_k^2]}+ C_9 h_k^2)  \bigg) \sqrt{\mathbb{E}[\Vert\eta_{k}\Vert^2]}\\
    &+ C_{10} (\gamma^{D}_k)^2  h_k^{-1}  + \tilde{C}_6^2(\gamma^{\theta}_k)^2 + 2\gamma^{D}_k \tilde{C}_6 \gamma^{\theta}_k (C_8 \sqrt{\mathbb{E}[\xi_k^2]}+ C_9 h_k^2)\\
    \leq\ (1&-\gamma^{D}_k{{\varepsilon}_f^N})\mathbb{E}[\Vert\eta_{k}\Vert^2]    + \bigg(2 \tilde{C}_6 \gamma^{\theta}_k (1 + \gamma^{D}_k C_f) + 2\gamma^{D}_k (C_8 \sqrt{\mathbb{E}[\xi_k^2]}+ C_9 h_k^2)  \bigg)^2 (\gamma^{D}_k {{\varepsilon}_f^N})^{-1}\\
    &+ C_{10} (\gamma^{D}_k)^2  h_k^{-1}  + \tilde{C}_6^2(\gamma^{\theta}_k)^2 + 2\gamma^{D}_k \tilde{C}_6 \gamma^{\theta}_k (C_8 \sqrt{\mathbb{E}[\xi_k^2]}+ C_9 h_k^2).
\end{align*}
With Assumption \ref{aspt:step_size}, it can be simplified as 
\begin{align*}
    \mathbb{E}[\Vert \eta_{k+1}&\Vert^2] \leq(1-\gamma^{D}_k{{\varepsilon}_f^N})\mathbb{E}[\Vert\eta_{k}\Vert^2] + O\big[ \big(\gamma^{\theta}_k (1 + \gamma^{D}_k )+ \gamma^{D}_k (\sqrt{\mathbb{E}[\xi_k^2]}+h_k^2)  \big)^2 (\gamma^{D}_k)^{-1}\\
    &\quad\quad\quad\quad\!\!   + (\gamma^{D}_k)^2 h_k^{-1}  + (\gamma^{\theta}_k)^2 + \gamma^{D}_k \gamma^{\theta}_k ( \sqrt{\mathbb{E}[\xi_k^2]}+ h_k^2) \big]\\
    \leq&\ (1-\gamma^{D}_k{{\varepsilon}_f^N})\mathbb{E}[\Vert\eta_{k}\Vert^2] + O\big[ (\gamma^{\theta}_k)^2 (\gamma^{D}_k)^{-1} + \gamma^{D}_k (\gamma^{q}_k+(\gamma^{\theta}_k)^2 (\gamma^{q}_k)^{-2} +h_k^4) + (\gamma^{D}_k)^2 h_k^{-1} \big]\\
    \leq&\ (1-\gamma^{D}_k{{\varepsilon}_f^N})\mathbb{E}[\Vert\eta_{k}\Vert^2] + O\big[ \gamma^{D}_k \big((\gamma^{\theta}_k)^2 (\gamma^{q}_k)^{-2} +h_k^4 + \gamma^{D}_k h_k^{-1} \big) \big].
\end{align*}
By a similar recursive derivation to that in inequality (\ref{order1}), we can conclude the final result as $\mathbb{E}[\Vert D_{k,i} - \nabla_\theta F^{-1}(z_i;\theta_k) \Vert^2] = O\big((\gamma^{\theta}_k)^2 (\gamma^{q}_k)^{-2} +h_k^4 + \gamma^{D}_k h_k^{-1} \big).$
\endproof

\subsection{Proof of Theorem~\ref{th:mse_theta}} 

Let $\Delta_k = \theta_k-\theta^\ast_N$, $\eta_{k,i}= D_{k,i} - \nabla_\theta F^{-1}(z_i;\theta_k)$. From equality (\ref{eq:iter_theta_proj}), we have 
\begin{equation*}
\resizebox{\linewidth}{!}{$%
\begin{aligned}
   &\Vert\Delta_{k+1}\Vert^2\\
   &= \Vert\Delta_k + \gamma^{\theta}_{k}\sum_{i=1}^N - D_{k,i} \Delta\tilde{w}_i+ \gamma^{\theta}_{k}P_k\Vert^2 
    = \Vert\Delta_k + \gamma^{\theta}_{k} \nabla_\theta\mathcal{J}_N(\theta_k) + \gamma^{\theta}_{k}\sum_{i=1}^N - \eta_{k,i} \Delta\tilde{w}_i+ \gamma^{\theta}_{k}P_k\Vert^2\\
    &\leq \Vert\Delta_k\Vert^2 + (\gamma^{\theta}_{k})^2 \Vert\sum_{i=1}^N - D_{k,i} \Delta\tilde{w}_i+ P_k\Vert^2 + 2\gamma^{\theta}_{k}\Delta_k^{\top}\nabla_\theta\mathcal{J}_N(\theta_k) - 2 \gamma^{\theta}_{k}\Delta_k^{\top}\sum_{i=1}^N  \eta_{k,i} \Delta\tilde{w}_i+ \gamma^{\theta}_{k}\Delta_k^{\top}P_k\\
    &\leq \Vert\Delta_k\Vert^2 + 4(\gamma^{\theta}_{k})^2 \Vert\sum_{i=1}^N -D_{k,i}\Delta\tilde{w}_i\Vert^2 + 2\gamma^{\theta}_{k}\Delta_k^{\top}H(\hat{\theta})\Delta_k - 2 \gamma^{\theta}_{k}\Delta_k^{\top}\sum_{i=1}^N  \eta_{k,i} \Delta\tilde{w}_i+ \gamma^{\theta}_{k}\Delta_k^{\top}P_k,
\end{aligned}
$}%
\end{equation*}
where the last equality is based on the fact provided in the proof of Theorem~\ref{th:main_convergence} that $\Vert P_k\Vert \leq \Vert\sum_{i=1}^N -D_{k,i}\Delta\tilde{w}_i\Vert$ and Taylor expansion of $\nabla_\theta\mathcal{J}_N(\theta_k)$ around $\theta^\ast_N$, and $\hat{\theta}$ lies on the line segment between $\theta_k$ and $\theta^\ast_N$. Under Assumption \ref{aspt:hessian}, and by applying Rayleigh-Ritz
inequality \cite[see, e.g.,][]{rugh1996linear}, we can obtain
\begin{align*}
    \Vert\Delta_{k+1}\Vert^2 \leq (1-2\gamma^{\theta}_{k} C_{\lambda}^N)\Vert\Delta_k\Vert^2 &+ 4(\gamma^{\theta}_{k})^2 \Vert\sum_{i=1}^N D_{k,i}\Delta\tilde{w}_i\Vert^2 - 2 \gamma^{\theta}_{k}\Delta_k^{\top}\sum_{i=1}^N  \eta_{k,i} \Delta\tilde{w}_i+ 2\gamma^{\theta}_{k}\Delta_k^{\top}P_k,
\end{align*}
Taking expectations on both sides and applying Cauchy–Schwarz inequality, we can obtain 
\begin{align}
    \mathbb{E}[\Vert\Delta_{k+1}\Vert^2] \leq (1-2\gamma^{\theta}_{k} C_{\lambda}^N) \mathbb{E}[\Vert\Delta_k\Vert^2] &+ 4(\gamma^{\theta}_{k})^2 \sup_i\mathbb{E}[\Vert D_{k,i}\Vert^2]  \nonumber \\
    &+ 2 \gamma^{\theta}_{k}\sqrt{\mathbb{E}[\Vert\Delta_k\Vert^2]} \big( \sup_i\sqrt{\mathbb{E}[\Vert\eta_{k,i}\Vert^2]}  + \sqrt{\mathbb{E}[\Vert P_k\Vert^2]}\big)\nonumber\\
    \leq (1-\gamma^{\theta}_{k} C_{\lambda}^N) \mathbb{E}[\Vert\Delta_k\Vert^2] &+ 4(\gamma^{\theta}_{k})^2 \sup_i\mathbb{E}[\Vert D_{k,i}\Vert^2]\nonumber   \\
    &+  2\gamma^{\theta}_{k} (C_{\lambda}^N)^{-1} \big( \sup_i \mathbb{E}[\Vert\eta_{k,i}\Vert^2]   +  \mathbb{E}[\Vert P_k\Vert^2] \big), \label{eq: tmp1}
\end{align}
Now we derive a bound for $\mathbb{E}[\Vert P_k \Vert^2]$. Since the interior of $\Theta$ is not empty, so that there exists constant $C_{\Theta}>0$ such that $\mathcal{B}(\theta^\ast_N, 2C_{\Theta})\subseteq\Theta$, where $\mathcal{B}(\theta^\ast_N,\delta)$ represents a round neighborhood of $\theta^\ast_N$ with radius $\delta$. Let $\mathcal{E}_k=\{\theta_{k}+\gamma^{\theta}_k \sum_{i=1}^N -D_{k,i}\Delta\tilde{w}_i\notin \mathcal{B}(\theta^\ast_N, 2C_{\Theta})\}$. By the definition of the projection function $\varphi_{\Theta}(\cdot)$, the occurrence of $\mathcal{E}_k^c$ implies $P_k=0$. Note that $\Vert P_k \Vert\leq \Vert \sum_{i=1}^N -D_{k,i}\Delta\tilde{w}_i \Vert \leq \sup_i\Vert  D_{k,i}  \Vert$ as proved in Theorem \ref{th:main_convergence}. Thus, we have 
\begin{align*}
    & \mathbb{E} [\Vert P_k \Vert^2] =\mathbb{E}[\Vert P_k \Vert^2|\mathcal{E}_k]P(\mathcal{E}_k)
    \leq  \mathbb{E}[\Vert  \sum_{i=1}^N -D_{k,i}\Delta\tilde{w}_i  \Vert^2]P(\mathcal{E}_k)  
     \\&\leq \mathbb{E}[\Vert  \sum_{i=1}^N -D_{k,i}\Delta\tilde{w}_i  \Vert^2]\big(P(\Vert (\theta_{k}+\gamma^{\theta}_k \sum_{i=1}^N -D_{k,i}\Delta\tilde{w}_i)-\theta_k \Vert\geq C_{\Theta})+P(\Vert \theta_{k}-\theta^\ast_N \Vert\geq C_{\Theta})\big).
\end{align*}
Using Markov's inequality, we further have 
\begin{align}
    \mathbb{E} [\Vert P_k \Vert^2] &\leq  \big((\gamma^{\theta}_k)^2 \sup_i\mathbb{E}[\Vert  D_{k,i}  \Vert^2]^2+\mathbb{E}[\Vert \Delta_k \Vert^2]  \mathbb{E}[\Vert  \sum_{i=1}^N -D_{k,i}\Delta\tilde{w}_i  \Vert^2]\big)C_{\Theta}^{-2}. \label{eq: tmp2}
\end{align}
Let $C_{11}=\sup_i\mathbb{E}[\Vert D_{k,i}\Vert^2]$, which is finite by Theorem \ref{th:d_stability}. 
By substitutions into inequality (\ref{eq: tmp1}), we can obtain 
\begin{align*}
    \mathbb{E}[\Vert\Delta_{k+1}\Vert^2] \leq \bigg(1&-\gamma^{\theta}_{k} (C_{\lambda}^N- 2(C_{\lambda}^N)^{-1}C_{\Theta}^{-2}\mathbb{E}[\Vert  \sum_{i=1}^N -D_{k,i}\Delta\tilde{w}_i  \Vert^2]) \bigg) \mathbb{E}[\Vert\Delta_k\Vert^2] + 4(\gamma^{\theta}_{k})^2 C_{11} \\ 
    &+  2\gamma^{\theta}_{k} (C_{\lambda}^N)^{-1} \big( \sup_i \mathbb{E}[\Vert\eta_{k,i}\Vert^2]   +  (\gamma^{\theta}_k)^2 C_{11}^2C_{\Theta}^{-2} \big),
\end{align*}
By assumption, we have $\theta_k\rightarrow\theta^\ast_N\in\Theta^\circ$ a.s. and $\sum_{i=1}^N \nabla_\theta F^{-1}(z_i;\theta^\ast_N) \Delta\tilde{w}_i = 0$.
The continuity of $\nabla_\theta F^{-1}(z_i;\theta)$ on compact $\Theta$ implies 
$\Vert  \sum_{i=1}^N \nabla_\theta F^{-1}(z_i;\theta_k) \Delta\tilde{w}_i \Vert \rightarrow 0$ a.s. and the boundedness of $\nabla_\theta F^{-1}(z_i;\theta)$, for $i=1,\cdots,N$, thereby indicating $\mathbb{E}[\Vert  \sum_{i=1}^N \nabla_\theta F^{-1}(z_i;\theta_k) \break \Delta\tilde{w}_i \Vert ] \rightarrow 0 $ by the dominated convergence theorem. By Theorem~\ref{th:mse_d}, 
$$\mathbb{E}[\Vert \sum_{i=1}^N -D_{k,i}\Delta\tilde{w}_i \Vert]\leq \sum_{i=1}^N\mathbb{E}[ \Vert  \eta_{k,i}\Vert ]\Delta\tilde{w}_i + \mathbb{E}[\Vert  \sum_{i=1}^N \nabla_\theta F^{-1}(z_i;\theta_k) \Delta\tilde{w}_i \Vert ]\rightarrow 0
$$ as $k\to \infty$, and thus, $\mathbb{E}[\Vert \sum_{i=1}^N -D_{k,i}\Delta\tilde{w}_i \Vert^2]\to 0$. Hence, by Theorem \ref{th:mse_d}, there exists an integer $K_{5}>0$ such that, for $k>K_{5}$,
\begin{align*}
    \mathbb{E}[\Vert\Delta_{k+1}\Vert^2] &\leq \big(1-\frac{\gamma^{\theta}_{k} C_{\lambda}^N}{2} \big) \mathbb{E}[\Vert\Delta_k\Vert^2] +   \gamma^{\theta}_{k} \cdot O\big((\gamma^{\theta}_k)^2 (\gamma^{q}_k)^{-2} +h_k^4 + \gamma^{D}_k h_k^{-1}\big) ,
\end{align*}
With a similar technique to that in inequality (\ref{order1}), we only expand from the $k$-th to $K_{5}$-th iteration and can conclude the final result as $\mathbb{E}[\Vert \theta_k -\theta^\ast_N\Vert^2] = O\big((\gamma^{\theta}_k)^2 (\gamma^{q}_k)^{-2} +h_k^4 + \gamma^{D}_k h_k^{-1}\big)$.
\endproof

\subsection{Proof of Theorem \ref{th:totalerror}}
By definitions and Assumptions \ref{aspt:dist_grad}, \ref{aspt:density}$^\ast$(a), we have 
\begin{align}
    \Vert \nabla_\theta &\mathcal{J}_N(\theta) - \nabla_\theta \mathcal{J}(\theta)\Vert
    \leq \sum_{i=1}^N \sup_{z\in[z_{i-1},z_i]}\big\Vert  \nabla_\theta F^{-1}(z_i;\theta) - \nabla_\theta F^{-1}(z;\theta)\big\Vert \cdot \vert\tilde{w}(z_{i}) - \tilde{w}(z_{i-1})\vert \nonumber\\ &\ \ \quad\quad\quad\quad\quad\quad\quad\quad\ \ \! + \sup_{z\in[0,z_0]\cup[z_N,1]} \big\Vert \nabla_\theta F^{-1}(z;\theta)\big\Vert (\vert\tilde{w}(z_0)-\tilde{w}(0)\vert + \vert\tilde{w}(1)-\tilde{w}(z_N)\vert)\nonumber\\
    \leq&  \sup_i\sup_{z\in[z_{i-1},z_i]}\big\Vert  \nabla_\theta F^{-1}(z_i;\theta) - \nabla_\theta F^{-1}(z;\theta)\big\Vert + C_g\varepsilon_f^{-1} (\vert\tilde{w}(z_0)-\tilde{w}(0)\vert + \vert\tilde{w}(1)-\tilde{w}(z_N)\vert). \label{eq:tmperrorJ}
\end{align}
Since $\nabla_\theta F^{-1}(z;\theta)$ is continuous on the compact set $[0,1]\times\Theta$, it is uniformly continuous. There exists a modulus $\lim_{\delta\to 0+}\omega(\delta)= 0 $ such that $\big\Vert  \nabla_\theta F^{-1}(z;\theta) - \nabla_\theta F^{-1}(z';\theta)\big\Vert \leq \omega(\vert z-z'\vert)$ for all $\theta$ and $z$, $z'$. Thus, together with the continuity of $w(\cdot)$ at the boundary, we have 
$\sup_{\theta\in\Theta} \Vert \nabla_\theta \mathcal{J}_N(\theta) - \nabla_\theta \mathcal{J}(\theta)\Vert\leq \omega(\sup_i\vert z_i-z_{i-1}\vert) + C_g\varepsilon_f^{-1} (\vert w(z_0)-w(0)\vert + \vert w(z_N)-w(1)\vert)\to 0$
as $N\to\infty$.
Next, we establish the convergence of $\theta^\ast_N$ to $\theta^*$. With Assumption \ref{aspt:diff_quantile}, we have the continuity of $\nabla \mathcal{J}(\theta)$ on the compact $\Theta$. Since $\theta^\ast$ is the unique root of this continuous vector field, there exist constants $r, c_r >0$ such that $\Vert\nabla_\theta\mathcal J(\theta)\Vert\ge c_r>0$ for every $\theta\in\Theta\setminus\mathcal{B}(\theta^\ast, r)$. By noticing the uniform convergence of the gradient proxy, one may choose $N_r>0$ such that $\sup_{\theta\in\Theta}\Vert\nabla_\theta\mathcal J_N(\theta)-\nabla_\theta\mathcal J(\theta)\Vert<c_r/2$ for all $N\ge N_r$. Thus, for such $N$ and every $\theta\notin\mathcal B(\theta^{\ast},r)$ we have $\Vert\nabla_\theta\mathcal J_N(\theta)\Vert\ge c_r/2>0$, implying $\theta^\ast_N\in\mathcal B(\theta^{\ast},r)$. As $r$ can be made arbitrarily small, it follows that $\lim_{N\to\infty}\theta_N^\ast=\theta^*$, and the final conclusion follows from Theorem~\ref{th:main_convergence}, which holds for any $N$.
\endproof

\subsection{Proof of Theorem~\ref{th:totalerrorrate}}
Since $\nabla_\theta F^{-1}(z;\theta)$ is uniformly Lipschitz and $w(\cdot)$ is Lipschitz continuous, it follows from (\ref{eq:tmperrorJ}) that $$\sup_{\theta\in\Theta}\Vert \nabla_\theta \mathcal{J}_N(\theta) - \nabla_\theta \mathcal{J}(\theta)\Vert = O(\max\{\sup|z_i-z_{i-1}|, z_0\})=O(N^{-1}).$$
Next, by Taylor expansion around $\theta^*$, we have
$\nabla_\theta\mathcal{J}_N(\theta^\ast) = \nabla_\theta\mathcal{J}_N(\theta^\ast_N) + \nabla^2\mathcal{J}(\hat{\theta})(\theta^\ast-\theta^\ast_N) = \nabla^2\mathcal{J}(\hat{\theta})(\theta^\ast-\theta^\ast_N)$,
where $\hat{\theta}$ lies on the line segment joining $\theta^\ast$ and $\theta^\ast_N$. 
Due to the local strong concavity of $\mathcal{J}(\cdot)$ at $\theta^\ast$, the Hessian $\nabla_\theta^2\mathcal{J}(\hat{\theta})$ is invertible and its inverse is uniformly bounded in a neighborhood around $\theta^\ast$. Therefore,
$\Vert \theta^\ast-\theta^\ast_N \Vert = \Vert (\nabla^2\mathcal{J}(\hat{\theta})(\theta^\ast-\theta^\ast_N))^{-1} \nabla_\theta\mathcal{J}_N(\theta^\ast)\Vert \leq \Vert (\nabla^2\mathcal{J}(\hat{\theta})(\theta^\ast-\theta^\ast_N))^{-1} \Vert \Vert \nabla_\theta\mathcal{J}_N(\theta^\ast)\Vert= O(\Vert \nabla_\theta\mathcal{J}_N(\theta^\ast)\Vert)$.
Using the previous gradient approximation result, we have
$\Vert \theta^\ast-\theta^\ast_N \Vert   =  O(\Vert  \nabla_\theta\mathcal{J}_N(\theta^\ast) \Vert)=O(\Vert  \nabla_\theta\mathcal{J}_N(\theta^\ast) -\nabla_\theta\mathcal{J}(\theta^\ast)\Vert) = O(N^{-1})$.
Finally, noting that the rate in Theorem~\ref{th:mse_theta} does not depend on $N$, the overall convergence rate result follows directly by combining the two bounds.
\endproof

\subsection{Proof of Theorem \ref{th:main_convergence2}} 
Analogous to the proof of Theorem~\ref{th:main_convergence}, the current proof is organized into two parts, each analyzing a different timescale using Theorem~\ref{th:single_convergence}.

\textbf{Analysis under the step-size $\gamma^{q}_k$.} Let $m^q_{k,i}=F(q_{k,i};\theta_k)-\mathbf{1}\{Y_{k}\leq q_{k,i}\}$. We rewrite coupled recursions as below:
\begin{equation}
\begin{aligned}
    q_{k+1,i} &= q_{k,i}+\gamma^{q}_{k}(z_i - F(q_{k,i};\theta_k) + m^q_{k,i}), \quad\quad\quad\quad\quad\quad\quad\quad\quad\quad\text{for}\ i=0,\cdots,N, \\
    \theta_{k+1} &= \theta_k + \gamma^{q}_{k}\frac{\gamma^{\theta}_{k}}{\gamma^{q}_{k}}  \sum_{i=1}^N G_1(X_k,\theta_k,q_{k,i})\tilde{w}'(z_i)(q_{k,i}-q_{k,i-1} ) + \gamma^{q}_{k}\frac{\gamma^{\theta}_{k}}{\gamma^{q}_{k}} P_k, \label{eq:iter2_unified_scale_q}
\end{aligned}
\end{equation}
which are expected to track coupled ODEs 
$\dot{q}_i(t) = z_i - F(q_i(t);\theta(t))$, and $\dot{\theta}(t) = 0$, equivalent to $\dot{q}_i(t) = z_i - F(q_i(t);\bar{\theta})$, and $\theta(t)\equiv \bar{\theta}$ given some fixed $\bar{\theta}\in\Theta$. 
Since $\{m^q_{k,i}\}$ is a square-integrable martingale difference sequence by the boundedness of CDF, $\sum_{k=1}^\infty (\gamma^{q}_{k})^2\mathbb{E}[|m^q_{k,i}|^2]<\infty$ by Assumption \ref{aspt:step_size}(b).
Since $P_k\in-C(\theta_{k+1})$ and $\theta_k\in\Theta$, we have 
\begin{equation}
   \begin{aligned}
    0&\leq P_k^{\top}(\theta_k-\theta_{k+1})=\gamma^{\theta}_{k}P_k^{\top}(- \sum_{i=1}^N G_1(X_k,\theta_k,q_{k,i}) \tilde{w}'(z_i)(q_{k,i}-q_{k,i-1} )-P_k)\\
    &\leq -\gamma^{\theta}_{k} \Vert P_k\Vert^2 + \gamma^{\theta}_{k} \Vert P_k\Vert  C_g \sup_{[0,1]} \vert\tilde{w}'(z)\vert \sum_{i=1}^N \vert q_{k,i} - q_{k,i-1}\vert, \label{eq:p_bound2}
\end{aligned} 
\end{equation}
where the last inequality utilize Assumption \ref{aspt:dist_grad}. 
In all subsequent theorems, we assume that the quantile estimates satisfy $ q_{k,i-1} \leq q_{k,i} $, for $ i = 1, \dots, N $, which implies $\sum_{i=1}^N \vert q_{k,i} - q_{k,i-1}\vert=\vert q_{k,N} - q_{k,0}\vert\leq 2\sup_i \vert q_{k,i}\vert$. If this does not hold, one may instead use the sorted quantile sequence $ \{q_{k,(i)}\} $ in place of $ \{q_{k,i}\} $ when computing the recursion in equation~\eqref{eq:iter2_q}. If $q_{k,i} $ converges to the real $z_i$-quantile, the effect of the sorting operation becomes negligible for sufficiently large $ k $.
With Lemma \ref{lem:q_stability} and the fact that $\tilde{w}(z)\in C^1[0,1]$, the inequality (\ref{eq:p_bound2}) implies $\sup_k \Vert P_k\Vert\leq\sup_k \Vert \sum_{i=1}^N G_1(X_k,\theta_k,\tilde{q}_{k,i}) \tilde{w}'(z_i)(q_{k,i}-q_{k,i-1} ) \Vert\leq 2C_g \sup_{[0,1]} \vert\tilde{w}'(z)\vert \sup_{k,i}\vert q_{k,i}\vert<\infty$ w.p.1. 
And we further know by Assumption \ref{aspt:step_size}(d) that the second term in the last recursion of (\ref{eq:iter2_unified_scale_q}) is asymptotically negligible.
The remaining conditions in Theorem \ref{th:single_convergence} can be verified by Assumption \ref{aspt:step_size}(b) together
with Lemma 1 in \cite{jiang2023quantile}. Thus, the outputs of recursions (\ref{eq:iter2_unified_scale_q}) a.s. converge to some limit set $\{ \{F(z_i;\bar{\theta})\}, \bar{\theta}: \bar{\theta}\in\Theta  \}$.

\textbf{Analysis under the step-size $\gamma^{\theta}_k$.} We now consider the second recursion (\ref{eq:iter2_theta}). For readability, define $g_i(\theta_k) = \nabla_{\theta}F(F^{-1}(z_i;\theta_k);\theta)\big|_{\theta=\theta_k} \tilde{w}'(z_i) (F^{-1}(z_i;\theta_k)-F^{-1}(z_{i-1};\theta_k))$ and
$m^{\theta}_k=\sum_{i=1}^N  (G_1(X_k,\theta_k,q_{k,i})-\nabla_{\theta}F(q_{k,i};\theta_k) )\tilde{w}'(z_i)(q_{k,i}-q_{k,i-1} )$. We decompose the update term in (\ref{eq:iter2_theta}) and rewrite it as
\begin{equation*}
\resizebox{1.02\linewidth}{!}{$%
\begin{aligned}
    \ \theta_{k+1} &= \theta_k +  \gamma^{\theta}_{k} \nabla_\theta\mathcal{J}_N^q(\theta_k)+ \gamma^{\theta}_{k} \sum_{i=1}^N  \nabla_{\theta}F(q_{k,i};\theta_k) \tilde{w}'(z_i)(q_{k,i}-q_{k,i-1} ) - g_i(\theta_k) + \gamma^{\theta}_{k} m^{\theta}_k + \gamma^{\theta}_{k} P_k.
\end{aligned}
$}%
\end{equation*}
Since $\mathbb{E}[m^{\theta}_k|\mathcal{F}_k]=0$, we have $\mathbb{E}[(m^{\theta}_k)^{\top} m^{\theta}_{k'}] = \mathbb{E}[(m^{\theta}_k)^{\top} \mathbb{E}[m^{\theta}_{k'}|\mathcal{F}_{k'}] ] = 0$ for all $k<k'$. The bias term $\sum_{i=1}^N  \nabla_{\theta}F(q_{k,i};\theta_k) \tilde{w}'(z_i)(q_{k,i}-q_{k,i-1} ) - g_i(\theta_k)$ diminishes w.p.1 because the sequences generated by recursions (\ref{eq:iter2_q}) and (\ref{eq:iter2_theta}) a.s. converge to $\{\{F(z_i;\bar{\theta})\}, \bar{\theta}: \bar{\theta}\in\Theta \}$ and $\nabla_\theta F(\cdot;\theta_k)$ is continuous.
By Assumptions \ref{aspt:step_size}(c), \ref{aspt:dist_grad}, and Lemma 1 in \cite{jiang2023quantile}, we have $\sup_k \mathbb{E}[\Vert \sum_{k'=0}^k \gamma^{\theta}_{k'} m^{\theta}_{k'}\Vert^2] < \infty$, since
\begin{align*}
    \mathbb{E}\big[\Vert m^{\theta}_{k}\Vert^2\big] &\leq (2C_g \sup_{[0,1]} \vert\tilde{w}'(z)\vert)^2\mathbb{E}\big[ \big(\sum_{i=1}^N \vert q_{k,i}-q_{k,i-1}\vert\big)^2\big]\\ 
    & \leq (2C_g \sup_{[0,1]} \vert\tilde{w}'(z)\vert)^2 4 \mathbb{E}\big[ \sup_{k,i}\vert q_{k,i}\vert^2\big]<\infty,
\end{align*}
where the last inequality follows from the proof of Lemma \ref{lem:q_stability}.
For the rest of the proof, we can argue as in the proof of Theorem \ref{th:main_convergence} and conclude that the output of the recursion (\ref{eq:iter2_theta}) converges to some limit set of ODE (\ref{eq:ode2_theta}) w.p.1.
\endproof

\subsection{Proof of Theorem \ref{th:mse2_q}} 

From the first part of the proof of Theorem \ref{th:main_convergence2}, we have
$$ \Vert P_k\Vert\leq \Vert \sum_{i=1}^N G_1(X_k,\theta_k,q_{k,i}) \tilde{w}'(z_i)(q_{k,i}-q_{k,i-1} ) \Vert\leq C_g \sup_{[0,1]} \vert\tilde{w}'(z)\vert \vert q_{k,N} -  q_{k,0}\vert.$$ It implies $\Vert \theta_{k+1}-\theta_k\Vert\leq 2 \gamma^{\theta}_k C_g \sup_{[0,1]} \vert\tilde{w}'(z)\vert \vert q_{k,N} -  q_{k,0}\vert $.
By Assumption \ref{aspt:diff_quantile}, $F^{-1}(z_i;\cdot)$ is Lipschitz, and there exist a constant $C_{12}>0$ such that $\vert F^{-1}(z_i;\theta_k) - F^{-1}(z_i;\theta_{k+1})\vert \leq \gamma^{\theta}_k C_{12}  \vert q_{k,N} -  q_{k,0}\vert$. Define $\xi_{k}=q_{k,i} - F^{-1}(z_i;\theta_k)$. Similar to Theorem \ref{th:mse_q}, we have 
\begin{align*}
    \mathbb{E}[\xi_{k+1,i}^2|\mathcal{F}_k]
    \leq\ \xi_{k}^2(1&-2\gamma^{q}_k {{\varepsilon}_f^N}) + (\gamma^{q}_k)^2+2(\gamma^{\theta}_k)^2 C_{12}^2  \vert q_{k,N} -  q_{k,0}\vert^2 \\& + 2\vert \xi_{k,i}\vert \gamma^{\theta}_k C_{12}  \vert q_{k,N} -  q_{k,0}\vert(1+\gamma^{q}_k C_f),
\end{align*}
where $\theta_{k+1}$ is $\mathcal{F}_k$-measurable, and $\hat{q}$ lies in the interval between $q_{k,i}$ and $F^{-1}(z_i;\theta_k)$. 
Define $\tilde{C}_{12}=2C_{12} \sup_{k,i}  \sqrt{\mathbb{E}[\vert q_{k,i}\vert^2]}$.
Since $ \mathbb{E}[ \vert q_{k,N} -  q_{k,0}\vert]\leq 2\sup_{k,i} \sqrt{\mathbb{E}[\vert q_{k,i}\vert^2]}<\infty$ by Lemma \ref{lem:q_stability}, we further obtain
\begin{align*}
    \mathbb{E}[\xi_{k+1}^2]  & 
    \leq \mathbb{E}[\xi_{k,i}^2](1-2\gamma^{q}_k {{\varepsilon}_f^N}) + (\gamma^{q}_k)^2 
    + (\gamma^{\theta}_k)^2 \tilde{C}_{12}^2 + 2 \mathbb{E}[\vert \xi_{k,i}\vert]\gamma^{\theta}_k \tilde{C}_{12}(1+\gamma^{q}_k C_f)\\
    &\leq \mathbb{E}[\xi_{k,i}^2](1-\gamma^{q}_k {{\varepsilon}_f^N}) + (\gamma^{q}_k)^2 + C_{13} (\gamma^{\theta}_k)^2 (\gamma^{q}_k)^{-1} ,
\end{align*}
where $C_{13} = \tilde{C}_{12}^2\big(\gamma^{q}_1+(1+\gamma^{q}_1 C_f)^2/{{\varepsilon}_f^N}\big)$. 
Finally, it follows that $\mathbb{E}[\vert q_{k,i} - F^{-1}(z_i;\theta_k)\vert^2] = O(\gamma^{q}_k + (\gamma^{\theta}_k)^2 (\gamma^{q}_k)^{-2})$, by an argument similar to that in Theorem~\ref{th:mse_q}.
\endproof

\subsection{Proof of Theorem \ref{th:mse2_theta}} 
Denote  
$G_k = \sum_{i=1}^N G_1(X_k,\theta_k,q_{k,i}) \tilde{w}'(z_i)(q_{k,i}-q_{k,i-1} )$ and $\bar{G}_k = \mathbb{E}[G_k\vert \mathcal{F}_k]$ i.e., $m^{\theta}_k = G_k - \bar{G}_k$.
Let $\Delta_k = \theta_k-\theta^+_N$. Rewrite recursion (\ref{eq:iter2_theta}) as 
\begin{align*}
    \Delta_{k+1} & = \Delta_k + \gamma^{\theta}_k {\nabla}_{\theta} \mathcal{J}^q_N(\theta_k) + \gamma^{\theta}_k (\bar{G}_k-{\nabla}_{\theta} {\mathcal{J}}^q_N(\theta_k)) + \gamma^{\theta}_k m^{\theta}_k + \gamma^{\theta}_k P_k.
\end{align*}
By squaring both sides, we have 
\begin{align}
    &\Vert\Delta_{k+1}\Vert^2 \leq  \Vert \Delta_k\Vert^2 + (\gamma^{\theta}_k)^2 \Vert G_k + P_k\Vert^2 + 2\gamma^{\theta}_k \Delta_k^{\top} {\nabla}_{\theta} \mathcal{J}_N^q(\theta_k) \nonumber\\
    &\quad\quad\quad\quad\quad\quad\quad\quad\!   + 2\gamma^{\theta}_k \Delta_k^{\top}(\bar{G}_k-{\nabla_{\theta} \mathcal{J}}^q_N(\theta_k)) + 2\gamma^{\theta}_k \Delta_k^{\top}m^{\theta}_k + 2\gamma^{\theta}_k \Delta_k^{\top}P_k \nonumber\\
    &\leq  (1-2\gamma^{\theta}_k\tilde{C}_{\lambda}^N)\Vert \Delta_k\Vert^2 + 4 (\gamma^{\theta}_k)^2 \Vert G_k \Vert^2 + 2\gamma^{\theta}_k \Delta_k^{\top}m^{\theta}_k + 2\gamma^{\theta}_k \Delta_k^{\top}P_k\nonumber\\
    &\quad\quad\  +2\gamma^{\theta}_k \Vert \Delta_k\Vert   \big\Vert\sum_{i=1}^N   \tilde{w}'(z_i)  \nabla_{\theta} F(q_{k,i};\theta_k)\big(\xi_{k,i} - \xi_{k,i-1}\big) \big\Vert
    +2\gamma^{\theta}_k \Vert \Delta_k\Vert \bigg(\sup_{[0,1]} \vert\tilde{w}'(z)\vert\nonumber\\
    &\quad\quad\ \cdot \sum_{i=1}^N  \Vert \nabla_{\theta} F(q_{k,i};\theta_k) -\nabla_{\theta} F(q;\theta_k)\vert_{q=F^{-1}(z_i;\theta_k)}\Vert \cdot
    \vert F^{-1}(z_i;\theta_k) - F^{-1}(z_{i-1};\theta_k)\vert\bigg). 
    \label{eq:tmpx}
\end{align}
The second inequality is based on Taylor expansion of ${\nabla}_{\theta}\mathcal{J}_N^q(\theta_k)$ around $\theta^+_N$ and Rayleigh-Ritz
inequality \cite[see, e.g.,][]{rugh1996linear}, i.e., $\Delta_k^{\top} {\nabla}_{\theta} \mathcal{J}_N^q(\theta_k) =  \Delta_k^{\top} \tilde{H} (\hat{\theta})\Delta_k\leq -\tilde{C}_{\lambda}^N\Vert \Delta_k\Vert^2$, where $\hat{\theta}$ lies on the line segment between $\theta_k$ and $\theta^+_N$.
Note that $\mathbb{E}[\Delta_k^{\top}m^{\theta}_k] = \mathbb{E}[\Delta_k^{\top}\mathbb{E}[m^{\theta}_k|\mathcal{F}_k]] =  \mathbb{E}[\Delta_k^{\top}\cdot 0] = 0$.
Taking expectations on both sides, we can obtain
\begin{align*}
    \mathbb{E}[\Vert\Delta_{k+1}\Vert^2] 
    \leq \ (1&-2\gamma^{\theta}_k\tilde{C}_{\lambda}^N) \mathbb{E}[\Vert \Delta_k\Vert^2] + 4 (\gamma^{\theta}_k)^2 \mathbb{E}[\Vert G_k \Vert^2] + 2\gamma^{\theta}_k \sqrt{\mathbb{E}[\Vert \Delta_k\Vert^2]}\sqrt{\mathbb{E}[\Vert P_k\Vert^2]}\\
    & + 2\gamma^{\theta}_k \sqrt{\mathbb{E}[\Vert \Delta_k\Vert^2]} \bigg(\mathbb{E}\big[\big\Vert\sum_{i=1}^N   \tilde{w}'(z_i)  \nabla_{\theta} F(q_{k,i};\theta_k)\big(\xi_{k,i} - \xi_{k,i-1}\big) \big\Vert\big]^{\frac{1}{2}} \\
    & + \sup_{[0,1]} \vert\tilde{w}'(z)\vert L_d\mathbb{E}\bigg[\bigg( \sum_{i=1}^{N-1} \vert F^{-1}(z_i;\theta_k) - F^{-1}(z_{i-1};\theta_k)\vert\cdot \vert\xi_{k,i} \vert \bigg)^2 \bigg]^{\frac{1}{2}}\bigg),
\end{align*}
where $\mathbb{E}\big[\big( \sum_{i=1}^{N-1} \vert F^{-1}(z_i;\theta_k) - F^{-1}(z_{i-1};\theta_k)\vert  \vert\xi_{k,i} \vert \big)^2 \big]\leq N\sup_i \mathbb{E}[\xi_{k,i}^2]  \sum_{i=1}^N\mathbb{E}[\vert F^{-1}(z_i;\theta_k) - F^{-1}(z_{i-1};\theta_k)\vert^2] = O_N(\sup_i \mathbb{E}[\xi_{k,i}^2])$ for every fixed $N$, because $F^{-1}(z;\theta)$ is continuous on $[0,1]$.
Note that we also have
\begin{align*}
     \mathbb{E}\bigg[\bigg\Vert\sum_{i=1}^N \tilde{w}'(z_i)&   \nabla_{\theta}  F(q_{k,i};\theta_k) \big(\xi_{k,i} - \xi_{k,i-1}\big) \bigg\Vert^2 \bigg]^{\frac{1}{2}} 
    = \mathbb{E}\bigg[ \bigg\Vert \tilde{w}'(z_1)  \nabla_{\theta}  F(q_{k,1};\theta_k) \xi_{k,0} \\
    & \qquad\ +\tilde{w}'(z_N)  \nabla_{\theta}  F(q_{k,N};\theta_k) \xi_{k,N} + \sum_{i=1}^{N-1}    \big(\tilde{w}'(z_{i})    - \tilde{w}'(z_{i+1})\big)  \nabla_{\theta}  F(q_{k,i};\theta_k) \xi_{k,i} \\
    & \qquad\ + \sum_{i=1}^{N-1}    \tilde{w}'(z_{i+1})  \big(\nabla_{\theta}  F(q_{k,i};\theta_k) -   \nabla_{\theta}  F(q_{k,i+1};\theta_k)\big) \xi_{k,i}\bigg\Vert^2 \bigg]^{\frac{1}{2}}  \\
    &\leq  C_{14}\sup_i \sqrt{\mathbb{E}\big[\vert \xi_{k,i} \vert^2\big]} + \sup_{[0,1]} \vert\tilde{w}'(z)\vert L_d \mathbb{E}\bigg[\bigg( \sum_{i=1}^{N-1} \vert q_{k,i} - q_{k,i+1}\vert\cdot \vert\xi_{k,i} \vert \bigg)^2 \bigg]^{\frac{1}{2}},
\end{align*}
where the inequality follows from Assumptions \ref{aspt:dist_grad} and \ref{aspt:lip},
and $C_{14}=C_g(2\sup_{[0,1]} \vert\tilde{w}'(z)\vert     + \sum_{i=1}^{N-1}    \big\vert\tilde{w}'(z_{i})   - \tilde{w}'(z_{i+1}) \big\vert )< \infty$ since $w(z)\in C^{1,1}[0,1]$. Similarly, with Lemma \ref{lem:q_stability}, we obtain the rough bound for fixed $N$ that $\mathbb{E}\big[\big( \sum_{i=1}^{N-1} \vert q_{k,i} - q_{k,i+1}\vert\cdot \vert\xi_{k,i} \vert \big)^2 \big] \leq N\sup_i \mathbb{E}[\xi_{k,i}^2]  \sum_{i=1}^N\mathbb{E}[\vert q_{k,i} - q_{k,i+1}\vert^2] = O_N(\sup_i \mathbb{E}[\xi_{k,i}^2])$. 
Therefore, to sum up, there exists a constant $C_{15}^N>0$ such that $\sqrt{\mathbb{E}[\Vert \bar{G}_k - \nabla_\theta\mathcal{J}^q_N(\theta_k)\Vert^2]}\leq C_{15}^N\sup_{i} \sqrt{\mathbb{E}[\xi_{k,i}^2]}$.
By substitution, we can obtain
\begin{align*}
    & \mathbb{E}[\Vert\Delta_{k+1}\Vert^2] \leq  (1-2\gamma^{\theta}_k\tilde{C}_{\lambda}^N) \mathbb{E}[\Vert \Delta_k\Vert^2] \\
    &\quad\quad\quad\quad\quad\quad\quad\ \! + 4 (\gamma^{\theta}_k)^2 \mathbb{E}[\Vert G_k \Vert^2] + 2\gamma^{\theta}_k \sqrt{\mathbb{E}[\Vert \Delta_k\Vert^2]} \bigg(\sqrt{\mathbb{E}[\Vert P_k\Vert^2]} + C_{15}^N  \sup_i\sqrt{\mathbb{E}[\vert \xi_{k,i}\vert^2]} \bigg)\\
    &\leq  (1-\gamma^{\theta}_k\tilde{C}_{\lambda}^N) \mathbb{E}[\Vert \Delta_k\Vert^2] + 4 (\gamma^{\theta}_k)^2 \mathbb{E}[\Vert G_k \Vert^2] + 2\gamma^{\theta}_k (\tilde{C}_{\lambda}^N)^{-1} \bigg( \mathbb{E}[\Vert P_k\Vert^2]  + (C_{15}^N)^2  \sup_i \mathbb{E}[\vert \xi_{k,i}\vert^2]  \bigg).
\end{align*}
With the same construction as in inequality (\ref{eq: tmp2}), there exists constant $C_{\Theta}'>0$ such that $\mathcal{B}(\theta^+_N, 2C_{\Theta}')\subseteq\Theta$. Note that 
$$ 0\leq P_k^{\top}(\theta_k-\theta_{k+1}) = -P_k^{\top}(\gamma^{\theta}_k G_k + \gamma^{\theta}_k P_k)\leq \gamma^{\theta}_k\Vert P_k\Vert \Vert G_k\Vert - \gamma^{\theta}_k\Vert P_k\Vert^2 $$ by the definition of $P_k$. Thus, we have 
$\mathbb{E} [\Vert P_k \Vert^2] \leq  \mathbb{E}[\Vert G_k\Vert^2]  \cdot\big((\gamma^{\theta}_k)^2 \mathbb{E}[\Vert G_k\Vert^2]+\mathbb{E}[\Vert \Delta_k \Vert^2]\big)(C_{\Theta}')^{-2}$, which implies 
\begin{align*}
    \mathbb{E}[\Vert\Delta_{k+1}\Vert^2] 
    \leq  \big(1&-\gamma^{\theta}_k(\tilde{C}_{\lambda}^N-2(\tilde{C}_{\lambda}^N)^{-1}(C_{\Theta}')^{-2}\mathbb{E}[\Vert G_k\Vert^2])\big) \mathbb{E}[\Vert \Delta_k\Vert^2] + 4 (\gamma^{\theta}_k)^2 \mathbb{E}[\Vert G_k \Vert^2] \\
    & + 2\gamma^{\theta}_k (\tilde{C}_{\lambda}^N)^{-1} \bigg(  (\gamma^{\theta}_k)^2 \mathbb{E}[\Vert G_k\Vert^2]^2(C_{\Theta}')^{-2} + (C_{15}^N)^2  \sup_i \mathbb{E}[\vert \xi_{k,i}\vert^2]  \bigg).
\end{align*}
We can derive from previous formulas that
\begin{align}
    &\mathbb{E}[\Vert G_k\Vert^2] \leq \bigg(\sqrt{\mathbb{E}[\Vert \nabla_{\theta} \mathcal{J}_N^q(\theta_k)\Vert^2]} + \sqrt{ \mathbb{E}[\Vert\bar{G}_k-\nabla_{\theta} \mathcal{J}_N^q(\theta_k)\Vert^2]} +  \sqrt{\mathbb{E}[\Vert m^{\theta}_k \Vert^2]} \bigg)^2 \nonumber\\
    &\leq  \bigg( \sqrt{\mathbb{E}[\Vert \nabla_{\theta} \mathcal{J}_N^q(\theta_k)\Vert^2]} +  C_{15}^N \sup_i  \sqrt{\mathbb{E}[\vert \xi_{k,i}\vert^2]}  +  2C_g\sup_{[0,1]}\vert\tilde{w}'(z)\vert\sup_{i}\sqrt{\mathbb{E}[\Vert q_{k,i} \Vert^2]}\bigg)^2. \label{eq:tmpxx}
\end{align}
By Theorem \ref{th:main_convergence2} and definitions, we have $\theta_k\rightarrow\theta^+$ a.s. and ${\nabla}_{\theta}\mathcal{J}^q_N(\theta^+) = 0$.
The continuity of ${\nabla}_{\theta}\mathcal{J}^q_N(\theta)$ on compact $\Theta$ implies $\Vert {\nabla}_{\theta} \mathcal{J}_N^q(\theta_k) \Vert \rightarrow 0$
a.s. and the boundedness of ${\nabla}_{\theta}\mathcal{J}_N^q(\theta)$, thereby indicating 
$\mathbb{E}[\Vert  {\nabla}_{\theta}\mathcal{J}_N^q(\theta_k) \Vert ] \rightarrow 0$ 
by the dominated convergence theorem. With Theorem \ref{th:mse2_q}, the second term in the bracket declines to zero as $k$ grows. The last term is due to the estimation variance of $\nabla_{\theta}F(q_{k, i};\theta_k)$ and can be arbitrarily small with a sufficiently large batch size used for estimation at each iteration. Hence, there exists an integer $K_{6}>0$ such that for $k>K_{6}$, we can have $2 (\tilde{C}_{\lambda}^N )^{-1}(C_{\Theta}')^{-2}\mathbb{E}[\Vert G_k\Vert^2]  \leq \frac{1}{2} \tilde{C}_{\lambda}^N$, either by increasing the batch size or by enlarging the projection region, which yields a larger constant $C_\Theta'$.
Consequently, 
\begin{align*}
    \mathbb{E}[\Vert\Delta_{k+1}\Vert^2] 
    \leq &(1-\frac{1}{2}\gamma^{\theta}_k\tilde{C}_{\lambda}^N) \mathbb{E}[\Vert \Delta_k\Vert^2] + \gamma^{\theta}_k O_N(\gamma^{\theta}_k   +  (\gamma^{\theta}_k)^2+\gamma^{q}_k +(\gamma^{\theta}_k)^2(\gamma^{q}_k)^{-2}).
\end{align*}
With Assumption \ref{aspt:step_size}(d), we have $$\mathbb{E}[\Vert\Delta_{k+1}\Vert^2]  \leq (1-\frac{1}{2}\gamma^{\theta}_k\tilde{C}_{\lambda}^N) \mathbb{E}[\Vert \Delta_k\Vert^2] + \gamma^{\theta}_k O_N(\gamma^{q}_k +(\gamma^{\theta}_k)^2(\gamma^{q}_k)^{-2}),$$ which leads to the conclusion that $\mathbb{E}[\| \theta_k - \theta^+_N \|^2] = O_N(\gamma^{q}_k + (\gamma^{\theta}_k)^2 (\gamma^{q}_k)^{-2})$, by an argument analogous to that in Theorem~\ref{th:mse_q}.

Moreover, if the density $f(y,\theta)$ is bounded away from $0$, i.e., $\inf_{ ( q , \theta ) \in {\cal I } \times \Theta  } f(q;\theta )>\varepsilon_f $, then, together with continuity, $F^{-1}(z;\theta)$ is Lipschitz with $L_q>0$. Thus, we can obtain $\mathbb{E}\big[\big( \sum_{i=1}^{N-1} \vert F^{-1}(z_i;\theta_k) - F^{-1}(z_{i-1};\theta_k)\vert  \vert\xi_{k,i} \vert \big)^2 \big]^{\frac{1}{2}}\leq \mathbb{E}\big[\big(L_q \sum_{i=1}^{N-1} \vert z_i - z_{i+1}\vert\cdot \vert\xi_{k,i} \vert \big)^2 \big]^{\frac{1}{2}} = O(\sup_i \sqrt{\mathbb{E}[\xi^2_{k,i}]}\sum_{i=1}^{N-1}|z_{i}-z_{i-1}|) = O(\sup_i \sqrt{\mathbb{E}[\xi^2_{k,i}]})$, which shows that the bound is independent of $N$. To eliminate the dependency of $N$ in $\mathbb{E}\big[\big( \sum_{i=1}^{N-1} \vert q_{k,i} - q_{k,i+1}\vert\cdot \vert\xi_{k,i} \vert \big)^2 \big]^{\frac{1}{2}}$, we assume there exists a constant $C_{16}>L_q$ such that $q_{k,i} - q_{k,i-1}<C_{16}(z_i-z_{i-1})$. If this condition is not met, we can fix one endpoint of the sorted quantile vector $\{q_{k,(i)}\}$ and sequentially clip any interval $q_{k,(i)}-q_{k,(i-1)}$ that violates the Lipschitz condition of $F^{-1}(z;\theta)$, replacing $\{q_{k,i}\}$ with the adjusted vector $\{\tilde{q}_{k,(i)}\}$, where $\tilde{q}_{k,(i)}=q_{k,(0)}+\sum_{i'=1}^i  \max\{q_{k,(i)}-q_{k,(i-1)}, L_q(z_{i}-z_{i-1})\}$. Since $F^{-1}(z;\theta)$ is Lipschitz, the effect of the sorting and clipping operations becomes negligible for sufficiently large $k$ as $q_{k,i}$ converges to the real $z_i$-quantile. Therefore, we have $\mathbb{E}\big[\big( \sum_{i=1}^{N-1} \vert q_{k,i} - q_{k,i+1}\vert\cdot \vert\xi_{k,i} \vert \big)^2 \big]^{\frac{1}{2}} \leq \mathbb{E}\big[\big(L_q \sum_{i=1}^{N-1} \vert z_i - z_{i+1}\vert\cdot \vert\xi_{k,i} \vert \big)^2 \big]^{\frac{1}{2}} = O(\sup_i \sqrt{\mathbb{E}[\xi^2_{k,i}]})$. As a result, $C_{15}$ is no longer relies on $N$, and the rate $\mathbb{E}[\| \theta_k - \theta^+_N \|^2] = O(\gamma^{q}_k + (\gamma^{\theta}_k)^2 (\gamma^{q}_k)^{-2})$ holds for all $N$.
\endproof

\subsection{Proof of Theorem \ref{th:totalerror2}}
By definition and Assumptions \ref{aspt:dist_grad}, \ref{aspt:density}$^\ast$(a), we have 
\begin{align}
    \Vert \nabla_\theta &\mathcal{J}_N^q(\theta) - \nabla_\theta \mathcal{J}(\theta)\Vert\nonumber \\
    \leq&  \sum_{i=1}^N \bigg(\sup_{z\in[z_{i-1},z_i]}\big\Vert \tilde{w}'(z_i)\nabla_\theta F(q;\theta)\vert_{q=F^{-1}(z_i;\theta)} - \tilde{w}'(z)\nabla_\theta F(q;\theta)\vert_{q=F^{-1}(z ;\theta)}\big\Vert  \nonumber\\ &\quad\quad\ \cdot \vert  F^{-1}(z_i;\theta) -  F^{-1}(z_{i-1};\theta)\vert\bigg)+C_g\varepsilon_f^{-1} (\vert\tilde{w}(z_0)-\tilde{w}(0)\vert + \vert\tilde{w}(1)-\tilde{w}(z_N)\vert)\nonumber\\
    \leq&   \sup_i \sup_{z\in[z_{i-1},z_i]}\big\Vert \tilde{w}'(z_i)\nabla_\theta F(q;\theta)\vert_{q=F^{-1}(z_i;\theta)} - \tilde{w}'(z)\nabla_\theta F(q;\theta)\vert_{q=F^{-1}(z ;\theta)}\big\Vert \nonumber\\
    &\quad\quad\ \cdot \sup_{\theta\in\Theta}\vert  F^{-1}(z_N;\theta) -  F^{-1}(z_{0};\theta)\vert+C_g\varepsilon_f^{-1} (\vert\tilde{w}(z_0)-\tilde{w}(0)\vert + \vert\tilde{w}(1)-\tilde{w}(z_N)\vert), \label{eq:tmperrorJ2}
\end{align}
where the second term on the left-hand side arises from the same tail truncation as in $\nabla_\theta\mathcal{J}_N(\theta)$, and $\sup_{\theta\in\Theta}\vert  F^{-1}(z_N;\theta) -  F^{-1}(z_{0};\theta)\vert\leq \sup_{\theta\in\Theta}\vert  F^{-1}(1;\theta) -  F^{-1}(0;\theta)\vert<\infty$ by the compactness of $\Theta$ and the continuity of $F^{-1}(z;\theta)$.
Since $\tilde{w}'(z)$, $\nabla_\theta F(q;\theta)$, and $F^{-1}(z;\theta)$ are all continuous, the composition $\tilde{w}'(z)\nabla_\theta F(q;\theta)\vert_{q=F^{-1}(z ;\theta)}$ is continuous on the compact set $[0,1]\times\Theta$, and thus uniformly continuous. Therefore, by an argument similar to that in the proof of Theorem~\ref{th:totalerror}, we conclude that $\lim_{N\to\infty}\lim_{k\to\infty}\theta_k=\lim_{N\to\infty}\theta^+_N=\theta^*$, w.p.1.
\endproof

\subsection{Proof of Theorem~\ref{th:totalerrorrate2}}
Since $\nabla_\theta F(q;\theta)\vert_{q=F^{-1}(z;\theta)}$ is uniformly Lipschitz and $w'(\cdot)$ is Lipschitz continuous, it follows from (\ref{eq:tmperrorJ2}) that $$\sup_{\theta\in\Theta}\Vert \nabla_\theta \mathcal{J}_N^q(\theta) - \nabla_\theta \mathcal{J}(\theta)\Vert = O(\max\{\sup|z_i-z_{i-1}|, z_0\})=O(N^{-1}).$$
Therefore, similar to the proof of Theorem~\ref{th:totalerrorrate}, we have
$\Vert \theta^\ast-\theta^+_N \Vert  = O(N^{-1})$,
which implies the final results.
\endproof

\subsection{Proof of Theorem~\ref{th:main_convergence3}}
Due to the relative time-scale separation specified in Assumption~\ref{aspt:step_size}, the tracking of quantiles and their gradients is asymptotically independent of $\theta_k$ update. Following the arguments in the proofs of Theorems~\ref{th:main_convergence} and~\ref{th:main_convergence2}, the sequences $\{q_{k,i}\}$ and $\{D_{k,i}\}$ in Algorithm~\ref{alg:hybrid_drm} converge to the corresponding limits $\{F^{-1}(z_i;\bar\theta)\}$ and $\{d_i(\bar\theta)\}$ for some $\bar\theta \in \Theta$.
Finally, applying Theorem~\ref{th:single_convergence} to the last recursion, we conclude that the sequence $\{\theta_k\}$ converges almost surely to a stationary point of the projected ODE.
\endproof

\section{Experimental Settings}\label{appendix Experimental Settings}

\subsection{Experimental Details in Section \ref{Robust Portfolio Selection with DRM Criteria}}

In this scenario, we employ a mixed Gaussian distribution with $10$ components as the distributional prior across all instances. The feasible domain of trainable parameters is restricted to $[-2.5,2.5]^{30}$, and any violation is projected back into this region.
Except for Wang’s example, the numerical discretization of the DRM integral in all other instances utilizes $100$ equidistant grid points.
In Wang’s example, we adopt a non-uniform grid defined by $z_i = \sqrt{\frac{i}{250}}$ for $i=0,1,\dots,250$, which results in a denser discretization near one. This setting allocates more resolution to regions where the integration weights vary rapidly.
For higher numerical accuracy, we evaluate integrals at the midpoints of the grid, i.e., $\tilde{z}_i = \frac{z_i + z_{i-1}}{2}$, and apply linear extrapolation at both ends of the interval to further compensate for the dropped boundary intervals. The remaining hyperparameters for learning rate scheduling and kernel density estimation are summarized in Table~\ref{tab:para_portfolio}.

\begin{table}[h]
\caption{Hyperparameters in robust portfolio selection examples.}
\vspace{0.5em}
\label{tab:para_portfolio}
\centering
\renewcommand{\arraystretch}{0.9} 
\begin{adjustbox}{max width=\linewidth}
\begin{tabular}{ccccc}
\toprule
\multirow{2}{*}{DRM instance} & S-shape & Wang & CVaR & \multirow{2}{*}{Discontinuous} \\
 & ($\alpha=5.0$)& ($\alpha=-0.85$) & ($\alpha=0.7$) &  \\
\midrule
Burn-in offset $k_0$ & $1\times10^{3}$ & $1\times10^{3}$ & $5\times10^{2}$ & $5\times10^{2}$\\
Initial quantile gradient learning rate $\gamma_0^D$ & $6.25\times10^{-2}$ & $1\times10^{-1}$ & $2.5\times10^{-1}$ & $2.5\times10^{-1}$\\
Initial quantile learning rate $\gamma_0^q$ & $2.5\times10^{-1}$ & $1\times10^{0}$ & $2.5\times10^{-1}$ & $2.5\times10^{-1}$\\
Initial learning rate $\gamma_0^\theta$ & $6.25\times10^{-2}$ & $1\times10^{-2}$ & $6.25\times10^{-2}$ & $6.25\times10^{-2}$\\
Initial band width $h_0$ & $1\times10^{-3}$ & $1\times10^{-2}$ & $1\times10^{-2}$ & $1\times10^{-2}$\\
\bottomrule
\end{tabular}
\end{adjustbox}
\end{table}

\subsection{Experimental Details in Section \ref{Risk-sensitive Inventory Management via DRM-based Deep RL}}

In the multi-echelon supply chain system, the customers' demand is generated by $Q_t^0 = x_t + [(t+6)\text{ mod }15]$,
where $x_t$ is uniformly sampled from $\{0,1,\cdots,7\}$. Other exogenous parameters required by the inventory management simulation are shown in Table \ref{tab: exogenous parameters}.

\begin{table}[htbp]
\centering
\caption{Exogenous parameters in the multi-echelon supply chain system.}
\label{tab: exogenous parameters}
\vspace{0.5em}
\begin{tabular}{c>{\centering\arraybackslash}p{1.5cm}>{\centering\arraybackslash}p{1.5cm}cccc}
\toprule
Echelon & 1 & 2 & 3 & 4\\ 
\midrule
Lead time $L$ & 2 & 3 & 5 & / \\
Price $p$ & 2 & 1.5 & 1 & 0.5 \\
Holding cost $h$ & 0.2 & 0.15 & 0.1 & / \\
Penalty of lost sale $l$ & 0.125 & 0.1 & 0.075 & / \\
Initial inventory $I_0$ & 10 & 10 & 10 & $+\infty$ \\
\bottomrule
\end{tabular}
\end{table}

For deep RL, we adopt settings as consistent as possible with the baseline algorithm PPO if applicable, while all DPPO variants under different distortion functions share the same configuration.
The policy networks of both PPO and DPPO variants include a convolutional block composed of a temporal convolutional layer with a kernel size of 3 and 32 output channels, followed by another temporal convolutional layer with 64 output channels, and three parallel fully connected output layers. The critic network in PPO adopts the same convolutional block, but is followed by a single fully connected output layer. We employ the Tanh function as the activation function throughout. Other hyperparameters involved in algorithms are listed in Table \ref{tab:hyp inv}.

\begin{table}[htbp]
\centering
\caption{Hyperparameters in the inventory management example.}
\label{tab:hyp inv}
\vspace{0.5em}
\begin{tabular}{ccc}
\toprule
Hyperparameter & PPO & DPPO variants \\
\midrule
Initial learning rate & $2.5\times10^{-4}$ & $2.5\times10^{-4}$ \\
Burn-in offset & $2.5\times10^5$ & $2.5\times10^5$ \\ 
Update interval & $2.5\times10^2$ episodes & $2.5\times10^2$ episodes \\
Initial quantile gradient learning rate & / & $5\times10^0$ \\ 
Initial quantile learning rate & / & $1\times10^{-3}$ \\
Initial band width & / & $1\times10^0$ \\
\bottomrule
\end{tabular}
\end{table}

\newpage
\bibliography{mybib}

\end{document}